%% file: lbm_tpami_format.tex
\documentclass[twocolumn]{old.IEEEtran}

\usepackage{newtxtext,newtxmath}

\usepackage{graphicx}

\usepackage[english]{babel}
\usepackage{amsmath}

\usepackage{amssymb}
\usepackage{booktabs}
\usepackage{graphicx}
\usepackage{hyperref} 
\usepackage[capitalise]{cleveref}
\usepackage{multicol}
\usepackage{multirow}
\usepackage{caption}
\usepackage{subcaption}
\usepackage{comment}
\usepackage{xcolor}  
\usepackage{mdframed}
\usepackage{cite}
\usepackage{url}
\usepackage{etoolbox}
\usepackage[letterpaper,margin=1in]{geometry}

\makeatletter
\newcommand{\adaptivefigure}[1]{%
  \ifCLASSOPTIONtwocolumn
    \begin{figure*}[t]
      #1
    \end{figure*}
  \else
    \begin{figure}[t]
      #1
    \end{figure}
  \fi
}
\makeatother

\makeatletter
\newcommand{\adaptivetable}[1]{%
  \ifCLASSOPTIONtwocolumn
    \begin{table*}[]
      #1
    \end{table*}
  \else
    \begin{table}[]
      #1
    \end{table}
  \fi
}
\makeatother

\newcommand{\todo}[1]{}
\renewcommand{\todo}[1]{{\color{red} TODO: {#1}}}
\usepackage{xcolor}

\title{\bfseries \boldmath {\huge A Careful Examination of Large Behavior Models for Multitask Dexterous Manipulation}}

\author{
	TRI LBM Team\thanks{see Section~\ref{sec:authors} for full author list.}
}

\begin{document} 

\maketitle

\input{sections/01_abstract}

\input{sections/02_intro}

\input{sections/08_related_work}

\input{sections/07_results}

\input{sections/06_evaluation_protocol}

\input{sections/04_method}

\input{sections/05_experimental_setup}

\input{sections/09_conclusion_and_discussion}


\clearpage 

\renewcommand\refname{References}

{\scriptsize
\bibliographystyle{IEEEtran}
\bibliography{refs}
}


\newpage


\renewcommand{\thefigure}{S\arabic{figure}}
\renewcommand{\thetable}{S\arabic{table}}
\renewcommand{\theequation}{S\arabic{equation}}
\renewcommand{\thepage}{S\arabic{page}}
\setcounter{figure}{0}
\setcounter{table}{0}
\setcounter{equation}{0}
\setcounter{page}{1} 


\begin{center}
\section*{Supplementary Materials for\\ {A Careful Examination of Large Behavior Models for Multitask Dexterous Manipulation}}

TRI LBM Team

\end{center}

\input{sections/10_appendix}

\end{document}

%% file: sections/01_abstract.tex
\adaptivefigure{
\centering
\makebox[\textwidth][c]{%
  \includegraphics[width=1.0\textwidth]{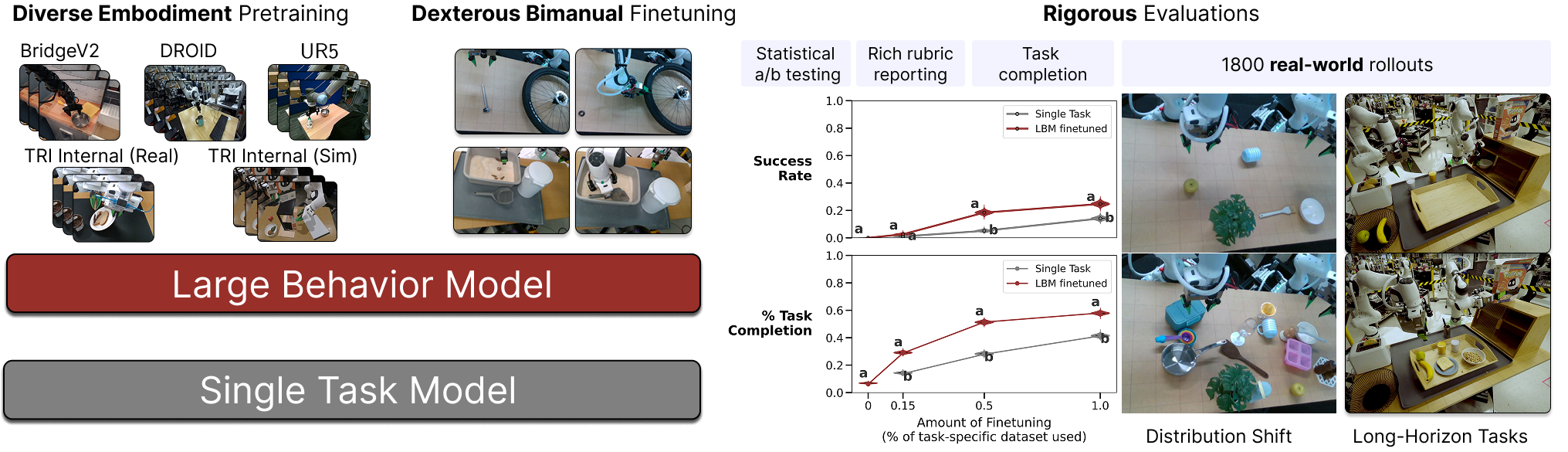}%
}

\caption{\small\textbf{Large Behavior Models (LBMs)} are visuomotor policies trained on a diverse corpus of simulation and real-world manipulation data. Through a carefully designed evaluation pipeline that leverages both simulation and real-world experiments, we show that finetuned LBMs are more robust to distribution shift and achieve better performance  than single-task baselines. Moreover, when finetuning on novel tasks, LBMs require a fraction of the data to achieve the same performance as baseline methods.
}
\label{fig:figure3}
}

\begin{abstract}
Robot manipulation has seen tremendous progress in recent years, with imitation learning policies enabling successful performance of dexterous and hard-to-model tasks. Concurrently, scaling data and model size has led to the development of capable language and vision foundation models, motivating large-scale efforts to create general-purpose robot foundation models. While these models have garnered significant enthusiasm and investment, meaningful evaluation of real-world performance remains a challenge, limiting both the pace of development and inhibiting a nuanced understanding of current capabilities.

In this paper, we rigorously evaluate multitask robot manipulation policies, referred to as Large Behavior Models (LBMs), by extending the Diffusion Policy paradigm across a corpus of simulated and real-world robot data. We propose and validate an  evaluation pipeline to rigorously analyze the capabilities of these models with statistical confidence. We compare against single-task baselines through blind, randomized trials in a controlled setting, using both simulation and real-world experiments. We find that multi-task pretraining makes the policies more successful and robust, and enables teaching complex new tasks more quickly,  using a fraction of the data when compared to single-task baselines. Moreover, performance predictably increases as pretraining scale and diversity grows. Project page: \href{https://toyotaresearchinstitute.github.io/lbm1/}{https://toyotaresearchinstitute.github.io/lbm1/}

\end{abstract}

%% file: sections/02_intro.tex
\section{Introduction}

Achieving flexible, generalist robots is a central ambition of robotics research. While modern robots are physically capable of performing a wide array of tasks in myriad settings, reliable autonomy has traditionally been limited to simple tasks or highly structured environments. Recently, visuomotor learning-based methods---trained to condition on robot sensor observations and produce low-level actions---have emerged as promising solutions to bridge this gap between hardware capabilities and autonomous performance. Behavior cloning, the most commonly used approach in this regime, eschews task-specific robot programming in favor of task-specific demonstrations, typically collected via teleoperation. Methods based on behavior cloning~\cite{chi2024diffusionpolicy,zhao2023learningfinegrainedbimanualmanipulation,zhao2024aloha} can produce complex, reactive, and contact-rich behaviors from hundreds to thousands of demonstrations, are well suited to handle traditionally challenging task attributes such as object deformability, transparency, reflectivity, and bimanual coordination, and offer the promise of producing general-purpose manipulation systems capable of performing arbitrary tasks. 

Despite these strengths, single-task behavior-cloned policies remain brittle, exhibiting limited generalization to task variations or environments outside their training distributions. To overcome this brittleness, the field is increasingly adopting \textit{Large Behavior Models} (LBMs)~\cite{octomodelteam2024octoopensourcegeneralistrobot, zhao2023learningfinegrainedbimanualmanipulation, black2024pi0visionlanguageactionflowmodel, nvidia2025gr00tn1openfoundation, geminiroboticsteam2025geminiroboticsbringingai, wang2024scaling, liu_rdt-1b_2025}--visuomotor foundation models trained on large-scale multitask datasets containing action-level demonstrations. Inspired by the success of large-scale generalist models in Computer Vision~\cite{kirillov2023segment, radford2021learningtransferablevisualmodels, zhai2023sigmoidlosslanguageimage, oquab2024dinov2learningrobustvisual} and Natural Language Processing~\cite{openai2024gpt4technicalreport, touvron2023llama1}, these models seek to improve reliability through broad training data support and more robust learned visual and sensory representations. Despite the surge in LBM research and development, significant uncertainty remains regarding the extent to which observed successes primarily stem from multitask pretraining.

To rigorously study the impact of multitask pretraining, we train multiple LBMs on approximately 1,700 hours of robot demonstrations comprised of over 500 internally collected high-diversity tasks as well as publicly available robot data. We comprehensively evaluate these models both in simulation and through 1,800 rigorously controlled real-world trials, including complex multi-step tasks that require tool use and precise manipulation. We design and employ an experimental protocol that incorporates blind A/B testing in the real world, large trial sizes with robust statistical analysis, qualitative and quantitative performance metrics, and carefully controlled initial conditions to ensure our conclusions hold with statistical significance.

Through these experiments, we find that:
\begin{enumerate}
\item LBM pretraining reduces the amount of task-specific data required, enabling finetuned specialist models to match single-task model performance with fewer demonstrations.
\item Given the same amount of task-specific data, finetuned specialists derived from pretrained LBMs outperform single-task models when aggregating over tasks.
\item Pretrained LBMs demonstrate increased robustness in scenarios diverging from their training conditions, amplifying the benefits described in points 1 and 2 in out-of-distribution evaluation settings.

\end{enumerate}

%% file: sections/08_related_work.tex
\section{Related Work}
\subsection{Robot Learning at Scale}

Robot learning is undergoing a paradigm shift towards creating generalist manipulation policies~\cite{octomodelteam2024octoopensourcegeneralistrobot, zhao2023learningfinegrainedbimanualmanipulation, black2024pi0visionlanguageactionflowmodel,nvidia2025gr00tn1openfoundation, geminiroboticsteam2025geminiroboticsbringingai,wang2024scaling,liu_rdt-1b_2025}, inspired by the scaling hypothesis successfully applied in Natural Language Processing~\cite{openai2024gpt4technicalreport, touvron2023llama1} and Computer Vision~\cite{kirillov2023segment, radford2021learningtransferablevisualmodels, zhai2023sigmoidlosslanguageimage, oquab2024dinov2learningrobustvisual}. This shift is driven by the creation of large-scale and diverse datasets~\cite{khazatsky2024droid, embodimentcollaboration2024openxembodimentroboticlearning, agibot-world} as well as high-capacity models, particularly transformer-based Vision-Language-Action (VLA) models~\cite{octomodelteam2024octoopensourcegeneralistrobot,kim2024openvlaopensourcevisionlanguageactionmodel, brohan2023rt2visionlanguageactionmodelstransfer, intelligence2025pi05visionlanguageactionmodelopenworld, yang2025magmafoundationmodelmultimodal, reed2022generalistagent, driess2023palmeembodiedmultimodallanguage, zawalski2025roboticcontrolembodiedchainofthought}, which have become central, integrating perception, language, and action within a unified framework trained largely via imitation learning. A key driver of VLA performance is transferring knowledge from large pretrained foundation models~\cite{bommasani2022opportunitiesrisksfoundationmodels} bringing  semantic understanding, improved reasoning, and strong visual representations, which when finetuned on robotics datasets, enable capabilities like zero-shot task execution. The choice of action representation---whether discretized into tokens~\cite{gemini_robotics_team_gemini_2025, brohan2023rt1roboticstransformerrealworld, kim_openvla_2024}, directly regressed as continuous commands~\cite{zhao2023learningfinegrainedbimanualmanipulation}, or generated through diffusion models~\cite{octomodelteam2024octoopensourcegeneralistrobot, black2024pi0visionlanguageactionflowmodel}---affects the policy’s ability to produce precise, multimodal, and real-time behaviors. Despite progress in training generalist policies, challenges such as catastrophic forgetting, data heterogeneity, scarcity of high-quality data, multimodal fusion, handling dexterity, and maintaining real-time inference speed remain open research problems. This work focuses on rigorously evaluating the effects of multi-task pretraining (as opposed to, for example, architectural novelty), and studies a fixed policy architecture (described in~\cref{subsec:policy_arch}) throughout.

\subsection{Datasets for Robot Learning}
Training generalist robot policies requires large-scale, diverse datasets, yet acquiring this data poses significant challenges. Unlike large-scale language and vision datasets~\cite{brown2020language,gao2020pile,dodge2021documenting,touvron2023llama1,radford2021learningtransferablevisualmodels,schuhmann2022laion,schuhmann2021laion,liu2023llava}, which are generally derived from internet sources, collecting robotics data in the real world is inherently slow and expensive. Robot data is most commonly collected via teleoperation, wherein human operators remotely control robots, yielding high-fidelity, embodiment-specific data. Large datasets such as RT-1~\cite{brohan2023rt1roboticstransformerrealworld}, Bridge~\cite{ebert2021bridgedataboostinggeneralization}, 
RH20T~\cite{fang2023rh20t}, DROID~\cite{khazatsky2024droid}, and AgiBot~\cite{agibotworldcontributors2025agibotworldcolosseolargescale}, among others, have been collected using this approach, often over the course of months or years with multiple robots. Pooled datasets like Open X-Embodiment~\cite{embodimentcollaboration2024openxembodimentroboticlearning} aggregate teleoperation and other types of robot data from numerous labs (over 1 million trajectories and 22 embodiments) with the hope of training policies that benefit from transfer across robot embodiments. 
Simulation~\cite{geng2025roboverse, mittal2023orbit, tao2024maniskill3gpuparallelizedrobotics} provides one promising alternative for cost-effectively generating robot manipulation data at scale~\cite{wang2023robogen, nasiriany2024robocasa, mandlekar2023mimicgen, james2020rlbench}; however, the differences between simulators and the real world present challenges. One method of overcoming these differences is to simultaneously train (``co-train'') on data from both domains~\cite{wei2025empiricalanalysissimandrealcotraining, maddukuri2025simandrealcotrainingsimplerecipe}. We use sim and real co-training in this work in order to more effectively \textit{evaluate} LBMs that were trained primarily on real-world data in simulation. 
Another approach to quickly and cost-effectively collect data is to circumvent the need for robots entirely through the use of specialized devices manually controlled by people~\cite{chi2024universalmanipulationinterfaceinthewild, Seo_2025, kareer2024egomimicscalingimitationlearning, fang2024airexolowcostexoskeletonslearning}. In this work, we train LBMs on a mixture of data (described in \cref{subsec:data}) sourced from openly available datasets as well as from our internal data collection efforts in the real world (using both teleoperated robots and specialized devices~\cite{chi2024universalmanipulationinterfaceinthewild}) and in simulation with the goal of better understanding the value of training on these large-scale, multi-task datasets.

\subsection{Evaluating Robotic Manipulation Policies}
Measuring the performance of LBMs, either for research or real-world deployment purposes, requires reproducible, reliable, and scalable evaluation methods and frameworks~\cite{kress2024robot}. Absence of standardized hardware makes consistent benchmarking a challenge. As a result, most benchmarks are simulation-based with notable examples including RLBench~\cite{james2020rlbench}, ManiSkill~\cite{mu2021maniskillgeneralizablemanipulationskill}, Meta-World~\cite{yu2021metaworldbenchmarkevaluationmultitask}, Robosuite~\cite{robosuite2020}, BEHAVIOR~\cite{srivastava2021behaviorbenchmarkeverydayhousehold}, and RoboTHOR~\cite{deitke2020robothoropensimulationtorealembodied}. 
Evaluation within these benchmarks typically relies on quantitative metrics such as success rate, task completion percentage, and completion time, and emphasizes generalization (for example, to unseen objects, tasks, or scenes) or sample efficiency.
While prior work in navigation highlights simulation-to-reality gaps caused by dynamics and visual discrepancies~\cite{Kadian_2020,deitke2020robothoropensimulationtorealembodied,zhang2019vrgogglesrobotsrealtosimdomain}, evaluation of manipulation policies poses additional challenges due to tighter robot and environment coupling and the sensitivity of task outcomes to subtle variations. 
SIMPLER~\cite{li24simpler}, a recent simulation framework, mitigates control and visual discrepancies between real and simulated environments through the use of system identification~\cite{memmel2024asidactiveexplorationidentification, pfaff2025_scalable_real2sim} and various image editing and matching techniques.
Alternative real-world evaluation approaches focus on establishing standardized object sets, tasks, datasets, and evaluation protocols~\cite{dasari2022rb2roboticmanipulationbenchmarking,morgan2019benchmarking,heo2023furniturebench,kimble2020benchmarking,khargonkar2024scenereplica,bekiroglu2019benchmarking,bottarel2020graspa,chatzilygeroudis2020benchmark}, remote access to shared robots~\cite{liu2021ocrtoc,bauer2022real,zhou2023train}, or improving evaluation efficiency~\cite{zhou2025autoevalautonomousevaluationgeneralist,snyder2025your}. Despite progress, challenges remain in evaluating generalist robotic manipulation policies across many diverse tasks, reliably benchmarking complex long-horizon interactions, and assessing robustness and safety critical aspects in dynamic environments.

%% file: sections/07_results.tex
\section{Results}
\label{sec:results}

We aim to achieve a nuanced understanding of LBM performance under a number of real-world conditions. Our main hypotheses are that due to  pretraining, 1) new tasks can be learned with less data; 2) policies achieve better performance; and 3) policies are more robust under distribution shift. For each individual task, we compare against a single-task policy trained from scratch. We use simulation and real-world (hardware) experiments to measure LBM performance on tasks for which demonstrations were seen during pretraining (``seen" in the following) as well as on novel tasks that were not used for pretraining (``unseen" in the following). We evaluate the policies under two conditions, nominal and distribution shift; we systematically create distribution shifts as described in Sec.~\ref{subsec:eval_tasks}. Task complexity varies from simple pick-and-place manipulation to long-horizon tasks requiring a high degree of dexterity, such as coring an apple, setting a breakfast tray, or installing a bike rotor. Each real-world task was evaluated with 50 rollouts per task per policy per condition. Simulation tasks were run 200 times per task per policy per condition; due to missing data, a few tasks were analyzed with fewer than 200 rollouts, see Section~\ref{subsec:missing_sim_rollouts} for details.

Section~\ref{subsec:eval_protocol_and_analysis} describes our evaluation protocol and our process for creating controlled and repeatable distribution shifts. Our metrics for measuring performance are success rate (SR) and task completion (TC). Success rate, while a useful and important signal that is standard across robot learning publications, does not tell the full story of policy performance~\cite{kress2024robot}. There is a notable difference between a policy that almost, but not quite, succeeds and one that does nothing. To capture and quantify this nuance, we create rubrics for real-world evaluation and predicates for simulation evaluation, as described in Section~\ref{subsec:rubrics_and_predicates}. Using these, we measure task completion, based on task-specific intermediate milestones. For real-world evaluation, task completion is evaluated by filling out rubrics manually; we  created a quality assurance (QA) process to measure the reliability of the rubrics. For simulation, task completion is computed automatically. 

Although we report absolute success rates, the most important results are the \emph{relative} success rates of the different methods. The absolute success rates are very task dependent and can easily be shifted up or down based on how difficult we make the task (e.g., by broadening the distribution of initial conditions) and/or by changing the number of task-specific demonstrations. We design our experiments to make tasks quite difficult---targeting policy success rates around ~50\%---so that the relative success rates are as informative as possible, but in practice end up with significantly higher or lower rates.

In the following, as described in Section~\ref{subsubsec:stats}, we use violin plots to convey the quantitative results with a horizontal line indicating the mean. For SR, the mean is the empirical success probability (successful runs/all runs) and the violin is the Bayesian posterior of individual success rates under a uniform Beta prior; it represents the uncertainty in the true success probability given the success rate and total number of runs. For TC, the mean is the mean of the task completion across all runs in either individual tasks or the aggregate of all tasks. For individual tasks, the violin represents the full data distribution of TC for that task. In plots where the tasks are aggregated, the violin represents the Bayesian posterior of the mean TC under a uniform 
Dirichlet prior. We emphasize that these distributions are evaluated for each individual policy checkpoint, and do not capture the randomness from the training process.

For each task we perform multi-task hypothesis checking to test whether the separation observed is statistically significant. We indicate statistical significance in the result plots using the Compact Letter Display (CLD) method, which assigns the same letter to policies that cannot be separated and different letters to indicate separation with statistical significance. 

In the following we discuss ``seen'' tasks (Section~\ref{subsec:LBMs_seen_tasks}), ``unseen'' tasks (Section~\ref{subsec:LBMs_unseen_tasks}), and the effects of the amount of data used to pretrain and finetune LBMs on the performance (Section~\ref{subsec:fractional_pretraining}).   

\subsection{LBM performance on ``seen'' tasks}
\label{subsec:LBMs_seen_tasks}

\adaptivefigure{
\centering
\hfill

\begin{subfigure}[t]{\linewidth}
\includegraphics[width=\linewidth]{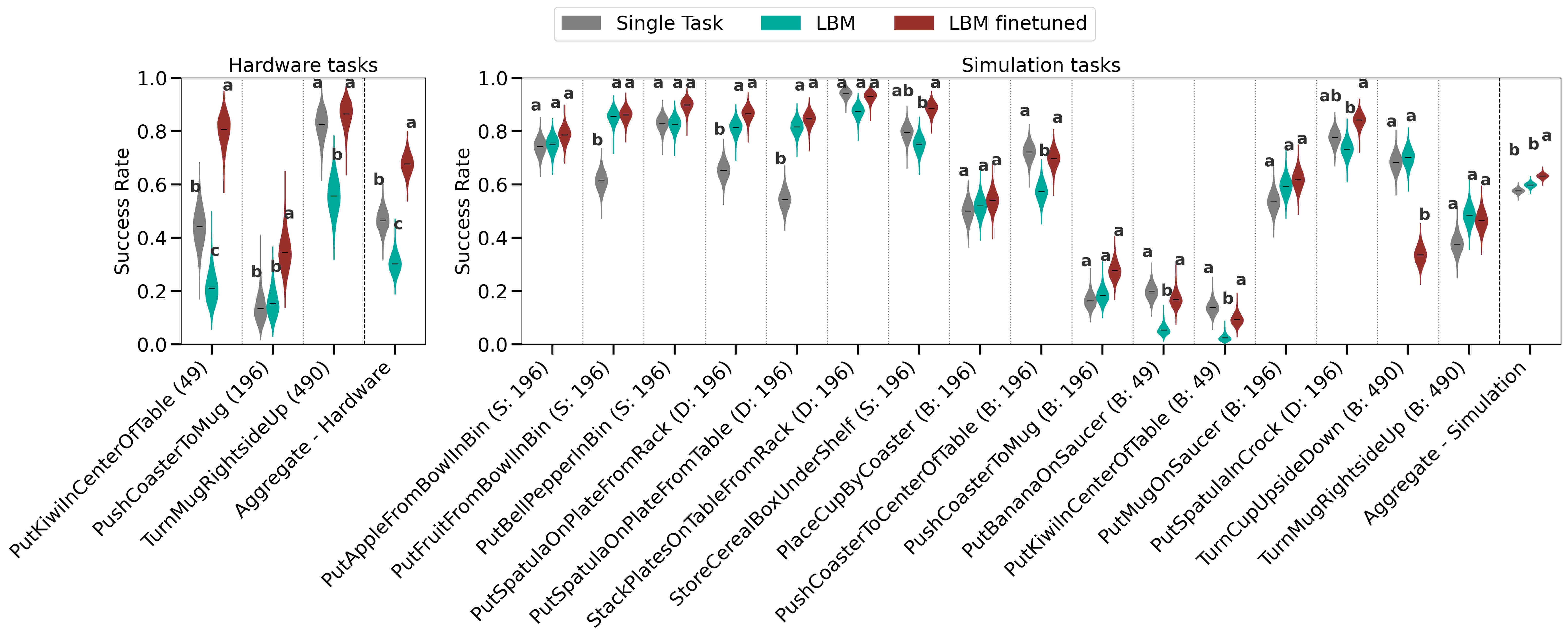}
    \caption{Nominal - no distribution shift}
    \label{fig:seen_tasks_sim_and_real_noDS}
  \end{subfigure}

\vspace{0.50em}

\begin{subfigure}[t]{\linewidth}
\includegraphics[width=\linewidth]{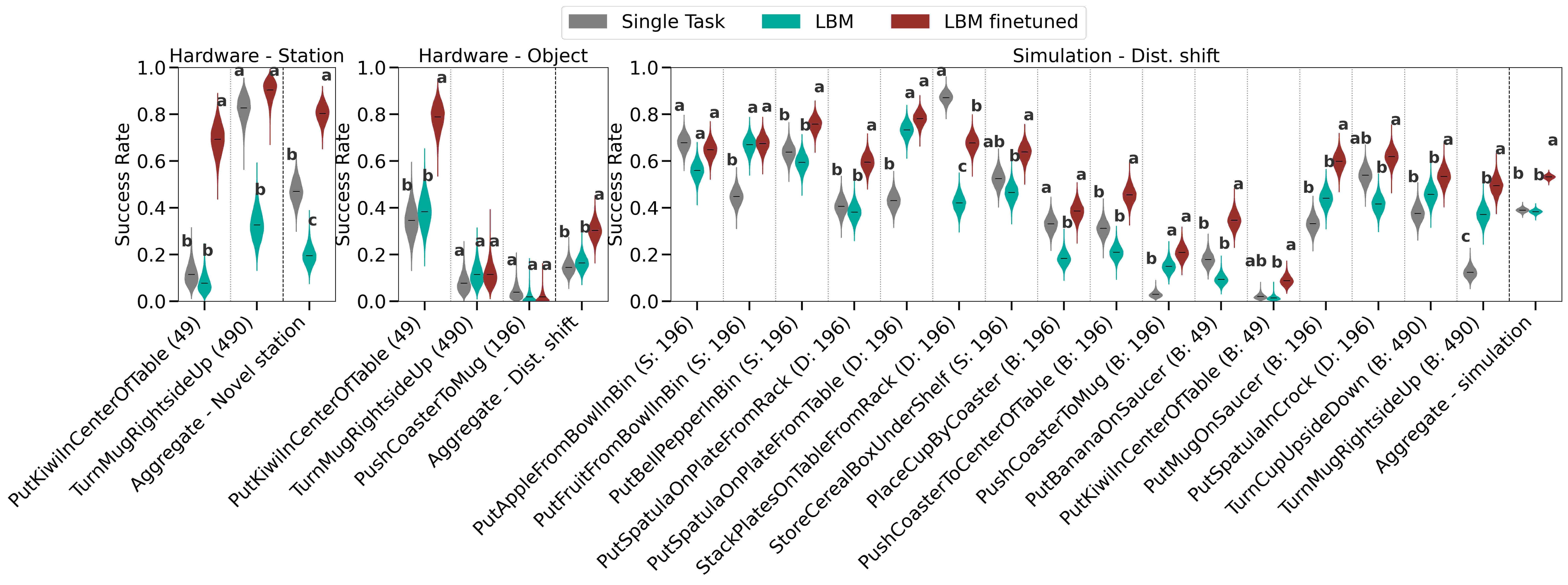}
    \caption{With distribution shift}
    \label{fig:seen_tasks_sim_and_real_DS}
  \end{subfigure}
\hfill
 \caption{\small\textbf{LBM performance on ``seen" tasks in real-world and in simulation without (a) and with (b) distribution shift.} We compare single-task with pretrained LBMs and with LBMs after finetuning. The x-axis labels show the task name, scenario name (for simulation tasks), and the number of demonstrations. Violin plots represent Bayesian posteriors of success rates under a uniform Beta prior and the observed success/failure data; policies labeled with different letters are statistically distinguishable.
 }
  \label{fig:seen_tasks_sim_and_real}
}

We first analyze LBM performance on a subset of tasks included in the pretraining dataset (i.e., ``seen" tasks), with results summarized in Figure~\ref{fig:seen_tasks_sim_and_real}. We evaluate LBMs that are only pretrained (Figure~\ref{fig:seen_tasks_sim_and_real}, teal) as well as pretrained LBMs with additional finetuning on individual tasks (Figure~\ref{fig:seen_tasks_sim_and_real}, maroon). Since these are relatively simple, short-horizon tasks, we do not present task completion results. The top row represents nominal conditions, and the bottom row represents experiments conducted under distribution shift. Details regarding how we systematically created the distribution shift are in Section~\ref{subsubsec:sim_tasks_and_data} for simulation and in Section~\ref{subsubsec:real_tasks_and_data} for the real world. 

For ``seen" tasks, we expect the pretrained LBM's success rate to be greater than zero since the tasks were in the training data, and we expect the finetuned LBMs to perform better than the single-task baseline, given the information that is encoded as part of the pretraining. In this set of experiments, we find that:

\noindent\textbf{Finetuned LBMs perform better on ``seen" tasks than the single-task baselines}: From Figure~\ref{fig:seen_tasks_sim_and_real}, we see that when aggregating over tasks, the finetuned LBM performs better than the single-task baseline under nominal and distribution-shift conditions both in simulation and in the real world; the finetuned LBM is statistically distinguishable from single-task in all the aggregate plots. 

When looking at individual tasks, the finetuned LBM is statistically the same or better than the single-task policy (i.e., it is labeled with ``a") in 3/3 real-world tasks and 15/16 sim tasks, both for the nominal conditions and the distribution shift. Interestingly, the one sim task in which single-task statistically outperforms the finetuned LBM is different between the experimental conditions, as further discussed below. Overall, the finetuned LBM is statistically better than single-task policies under nominal conditions in 2/3 real-world tasks and 3/16 simulation tasks; under distribution shift, finetuned LBM statistically outperforms single-task policies in 2/5 real-world tasks and 10/16 simulation tasks.

\noindent\textbf{Finetuned LBMs are more robust to distribution shift on ``seen" tasks than the single-task baseline}: As expected, and as seen when comparing the two rows of Figure~\ref{fig:seen_tasks_sim_and_real}, when we introduce distribution shift the overall task performance, here defined as success rate, drops on most tasks. However, we also observe that the finetuned LBMs go from statistically outperforming single-task policies on 3/16 policies in simulation under nominal conditions to 10/16 under distribution shift. Furthermore, for the aggregate simulation plots, the separation between finetuned LBM and single-task widens under distribution shift. These results suggest that finetuned LBM created policies that are more robust to distribution shift than policies trained from scratch; this result also holds for the ``unseen" tasks, as described in Sec.~\ref{subsec:LBMs_unseen_tasks}.

\noindent\textbf{LBMs without finetuning have nonzero success rate on ``seen" tasks and exhibit similar performance to the single-task baselines:} As expected, the pretrained LBM (without any task specific finetuning) has a success rate that is greater than zero on all tasks under nominal conditions. When aggregating over tasks, pretrained LBM is statistically indistinguishable from single-task in simulation (both conditions) and in real-world with object-centric distribution shift. On nominal real-world and station distribution shift, pretrained LBM performs worse than the single-task baseline. For individual tasks, in all real-world tasks, pretrained LBM is either statistically indistinguishable (5/8) or worse than (3/8) the single-task baseline, across both conditions. In simulation, under nominal conditions, pretrained LBM is statistically better than single-task in 3/16 tasks, and is statistically worse than single-task in 4/16. When considering distribution shift, the situation is similar, with pretrained LBM statistically better than single-task in 4/16 tasks and worse in 2/16 tasks. We make two observations related to the pretrained LBM; first, it performs worse than single-task mainly in the real-world tasks. 
Second, as discussed in Section~\ref{subsec:normalization}, we discovered an error in the pretraining process (data normalization) after the evaluations were complete. This error might also affect the pretrained LBM's performance (see Section~\ref{sec:supp_stage5_bug_comparison}).

We further investigate specific task performances that we found surprising.

\noindent\textbf{There are cases where the finetuned LBM performs statistically worse than the single-task baseline:} 
In each condition, there is one simulation task for which finetuned LBM statistically underperforms the single-task baseline; the under-performing task depends on whether evaluation was conducted under nominal conditions or distribution shift.

Under nominal conditions, 
\textit{TurnCupUpsideDown} finetuned LBM performs substantially worse than both pretrained LBM and the single-task baseline, with a success rate of $0.335$. Here, in almost half (89/200) of the rollouts for finetuned LBM on this task, the robot did not move away from its initial pose before the simulation times out, as shown in Fig.~\ref{fig:supp_qual_analysis_seen_single_skill_pause_histogram}(e-left). When introducing distribution shift, we did not observe this behavior (Fig~\ref{fig:supp_qual_analysis_seen_single_skill_pause_histogram}(f-left)). Under distribution shift, \textit{StackPlatesOnTableFromRack} performs worse than the single-task baseline. Inspecting the policy behavior did not reveal a systematic qualitative failure mode as in the previous case but confirmed the poor policy performance in general. 



\noindent\textbf{Performance on the \textit{Breakfast} scenario (simulation) tasks:} Our simulation tasks are grouped into a handful of ``scenarios", which start with the same environment assets and contain thematically similar tasks (see Section~\ref{subsec:simulation} for more details). We see that tasks belonging to the \textit{Breakfast} scenario (denoted by \textit{B:} in the X-axis caption of Fig~\ref{fig:seen_tasks_sim_and_real}, right) have higher variance in task success across tasks, compared to other scenarios, and that three tasks have substantially lower performance than the rest. We attribute this to two factors: first, for \textit{PutBananaOnSaucer} and \textit{PutKiwiInCenterOfTable} we have only 49 demonstrations; as expected and as discussed in the following sections, the less demonstration data available, the worse the performance. Second, for \textit{PushCoasterToMug}, we observe that the task is more difficult than the rest because the robot needs to push obstacles out of the way, the goal (mug) can be anywhere on the table, and all the demonstrations use a pushing strategy on the side of coaster, which is sensitive to variance in the height of the end effector. We provide additional details in Section~\ref{subsec:supplemental_breakfast_sceario}.

\noindent\textbf{A few tasks were more successful in the real world than in simulation:} For the ``seen" tasks, all the tasks that were evaluated in the real world were also evaluated in simulation. Given the lack of uncertainty in simulation, we would expect under nominal conditions that the success rate of these tasks would be higher in simulation. This is not the case for two tasks: \textit{PutKiwiInCenterOfTable} (\textit{Kiwi} for short) and \textit{TurnMugRightsideUp} (\textit{Mug} for short); for these tasks, across all policies, the real-world success rate is higher than simulation. One hypothesis for \textit{Mug} is that the simulation timeout (the maximum time a simulation rollout is run for) is too short for the policy to eventually succeed. In contrast to the real-world evaluation, where the operator decided when to terminate the rollout, in simulation the rollout will be stopped even if the task is about to be completed. 
Computing success rates on real-world rollouts by truncating to the simulation timeout yields a similarly low success rate.
For \textit{Kiwi} we observe different behavior between simulation and hardware and will continue to explore this discrepancy.    
We present further analysis in Section~\ref{subsec:supp_qual_analysis_sim_and_real}. Simulation and real-world evaluations under distribution shifts are not directly comparable due to the differences in how they are introduced respectively.

\subsection{LBM performance on ``unseen" tasks}
\label{subsec:LBMs_unseen_tasks}

\begin{figure}[htbp]
    \centering
    \includegraphics[width=\linewidth]{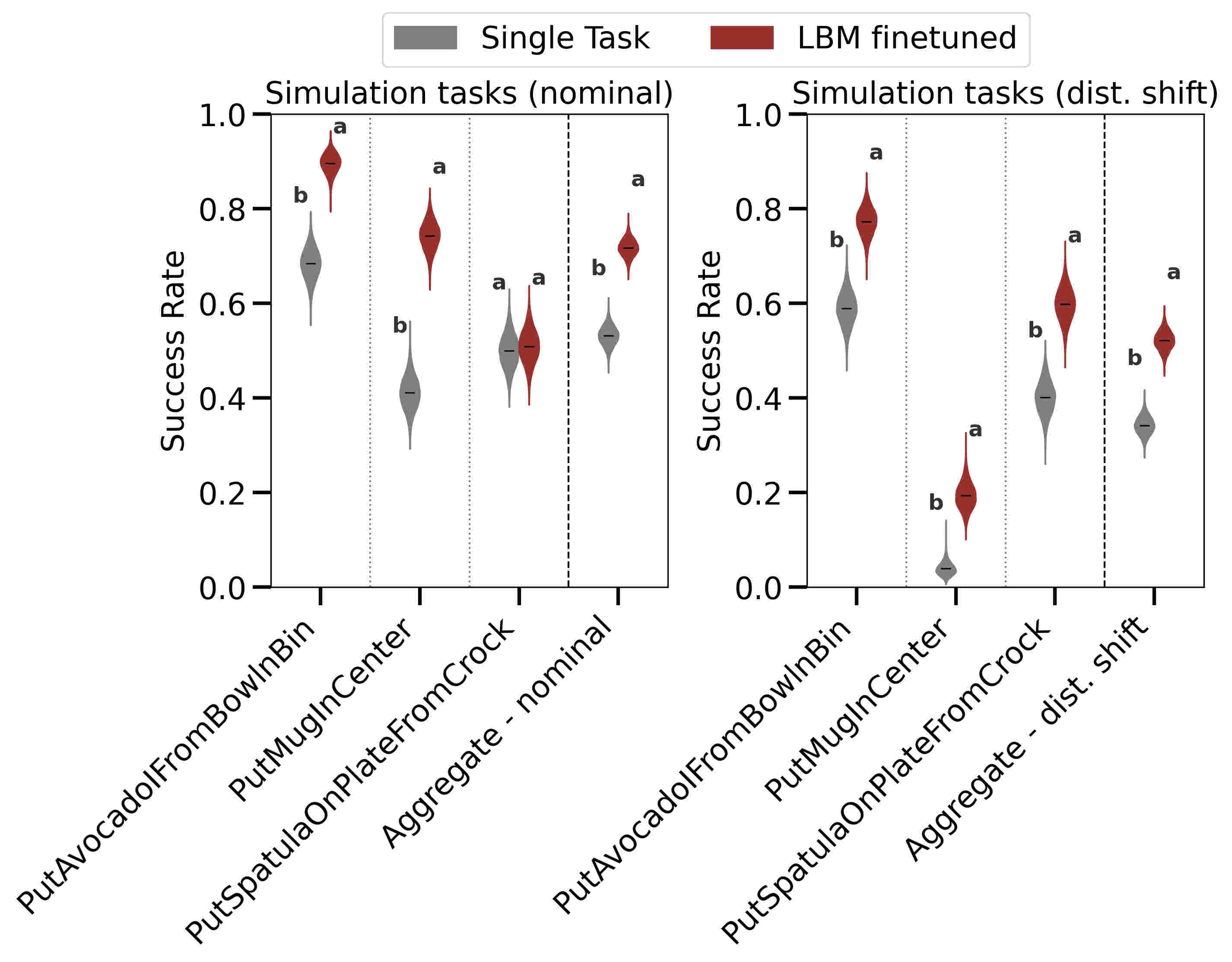}
    
    \caption{\small\textbf{LBM performance on ``unseen" simulation tasks from scenarios that are part of the simulation training set. Left:} evaluation done under nominal conditions. \textbf{Right:} evaluation done under distribution shift. Violin plots represent Bayesian posteriors of success rates under a uniform Beta prior and the observed success/failure data; policies labeled with different letters are statistically distinguishable.}
    \label{fig:unseen_easy_sim_tasks}
\end{figure}

\begin{figure}[htbp]
    \centering
    \includegraphics[width=0.8\linewidth]{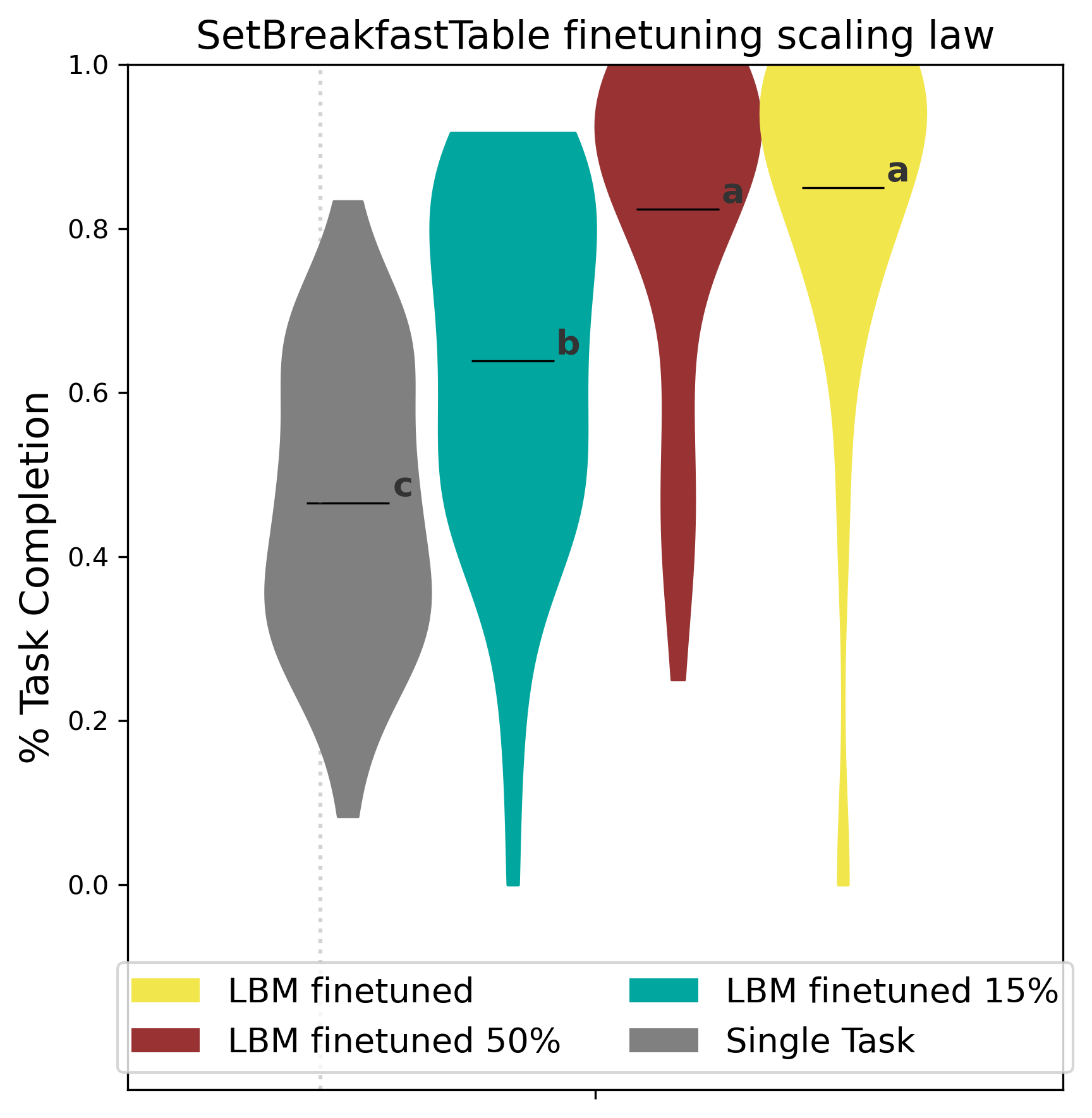}
    
    \caption{\small\textbf{Performance of LBMs finetuned on a fraction of \textit{SetBreakfastTable} data compared with the single-task baseline.} We highlight that the LBM finetuned with 15\% of data is already outperforming the baseline trained with all the data. }
    \label{fig:fractional_ft_breakfast}
\end{figure}

We explore LBM performance on tasks that do not appear in the pretraining dataset, i.e., unseen during training. It is straightforward to make ``unseen" tasks in simulation. First, we generate additional tasks in the same scenarios as the ``seen" tasks and that are similar in complexity, Figure~\ref{fig:unseen_easy_sim_tasks}. For the real world, generating tasks that are meaningfully distinct from our multi-year data collection effort is more challenging. To address this, we designed several long-horizon, multistep, dexterous tasks that were outside of the scope of  previous task collections. For parity, we also added a new simulation scenario, \textit{Kitchen}, with similarly long-horizon, multistep, dexterous tasks. These complex tasks are designed to test the limits of what our policies are capable of executing. Our results for the complex tasks are summarized in Fig~\ref{fig:unseen_tasks_sim_and_real} for nominal conditions and in Fig~\ref{fig:unseen_tasks_sim_and_real_ds} for evaluation with distribution shift (sim only). In both of these figures, we present success rate results on the top row and task completion results on the bottom row. Furthermore, in addition to the SR and TC results for policies created (trained or finetuned) with all the task data, we analyze the aggregate performance of the policies when finetuned (LBM) or trained (single-task baseline) with subsets of the data (plots on the right). We provide a similar analysis for one real-world task in Figure~\ref{fig:fractional_ft_breakfast}. 

For our ``unseen" tasks, especially the complex tasks, we do not expect the pretrained LBM to succeed; we therefore only compare the finetuned LBM and the single-task baseline. Furthermore, for the complex tasks, we expect low success rates and more intuition gained from the task completion plots. 

\input{figures_latex/composite_unseen}

\noindent\textbf{Finetuned LBMs perform better on ``unseen" tasks than the single-task baseline, both in nominal conditions and under distribution shift:} When aggregated across tasks, the finetuned LBM is statistically better than the single-task baseline in the real world and in simulation (both nominal and distribution shift), and across metrics (success rate and task completion). 

For the simpler simulation tasks, finetuned LBM is statistically better than single-task on 2/3 tasks in nominal conditions and all the tasks under distribution shift. For the complex tasks, while as expected the success rate is low, and even lower once considering distribution shift, some tasks still show statistical separation, and in all of those the finetuned LBM is more successful (2/5 tasks for real-world, 3/5 tasks in nominal simulation, and 1/5 tasks in simulation with distribution shift).

When considering task completion, the conclusion that finetuned LBMs outperform single-task baselines becomes clearer; finetuned LBM is statistically better than the single-task baseline in 4/5 real-world tasks, and in 4/5 simulation tasks both in nominal conditions and under distribution shift. Visually inspecting the data distribution for task completion indicates that the finetuned LBM is able to achieve more steps of the task compared to the single-task baseline. When considering real-world tasks, we see that there is a task, \textit{SetBreakfastTable}, that the single-task policy never completes,\footnote{Due to the Bayesian analysis prior assumptions, the success rate graph appears to have a mean that is greater than 0; the empirical success rate for the single-task baseline of \textit{SetBreakfastTable} is in fact 0.} while the finetuned LBM succeeds and achieves higher task completion; conversely, there are two tasks, \textit{BikeRotorInstall} and \textit{CutAppleInSlices}, where in all rollouts the finetuned LBM completed part of the task, while the single-task baseline sometimes completely failed.  

\noindent\textbf{Finetuned LBMs require less task-specific data to achieve similar performance as the single-task baseline:} In Figures~\ref{fig:unseen_tasks_sim_and_real} and~\ref{fig:unseen_tasks_sim_and_real_ds}, the rightmost column shows the finetuned LBM and single-task baseline performances when aggregating across all five simulation tasks; each data point corresponds to finetuning/training with a different fraction (by demonstration) of the available task-specific data. For task completion, across both conditions (nominal and distribution shift), for all fractions of data, finetuned LBM is statistically better than the single-task baseline. For success rate, since the overall success rate is low, especially under distribution shift, the finetuned LBM is statistically better starting at 50\% of the data. In aggregate, and interpolating, we see that to achieve similar performance in simulation, when finetuning an LBM we require less than 30\% of the data needed for training from scratch. 

We performed a similar experiment on the \textit{SetBreakfastTable} real-world task, as shown in Figure~\ref{fig:fractional_ft_breakfast} where the violin plot represents the full data distribution. LBM finetuned with only 15\% of the data, statistically outperforms the single-task baseline (trained on all the data), further supporting our simulation findings.

\input{figures_latex/composite_unseen_ds}

\subsection{Pretraining scaling laws}
\label{subsec:fractional_pretraining}

\begin{figure}[htbp]
    \centering
    \includegraphics[width=0.5\textwidth]{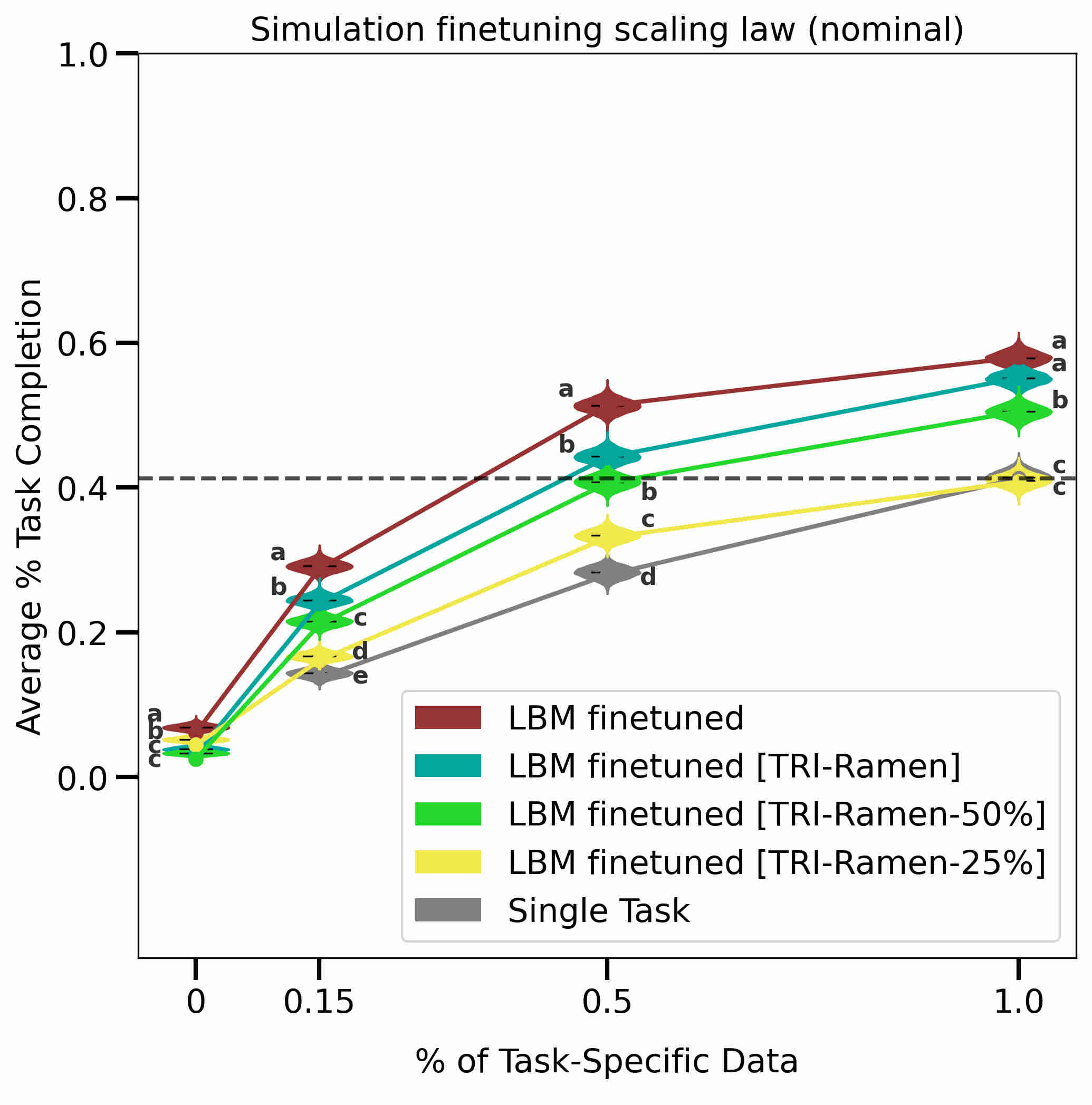}
    
    \caption{\small\textbf{LBM performance when pretraining with varying subsets of the full pretraining dataset. We evaluate under nominal conditions on five ``unseen" tasks in simulation and report the mean of the task completion metric across all tasks.} Violin plots represent Bayesian posteriors of the task completion mean under a uniform Dirichlet prior and the observed task completion scores; policies labeled with different letters are statistically distinguishable. Each data point corresponds to 200 rollouts per task, for a total of 1000 rollouts.  
    }
    \label{fig:unseen_tasks_fractional}
\end{figure}

Using simulation, we explore the scaling laws of LBM pretraining--the effects of pretraining dataset size on the performance of LBMs, measured through task completion on ``unseen" tasks. Due to the complexity of the tasks, the success rates are low (as seen in Figure~\ref{fig:unseen_tasks_sim_and_real}) and do not provide a statistically distinguishable conclusion, therefore we present only task completion results here.

We evaluate on the same five ``unseen" simulation tasks as in Sec.~\ref{subsec:LBMs_unseen_tasks} and perform experiments under nominal conditions, where we compare LBMs that were pretrained with different fractions of the data (Section~\ref{subsec:data}) used to train the LBM of the previous subsections. We create four pretraining datasets for this experiment: 1) the full dataset (\textbf{OXE-Ramen} and \textbf{TRI-Ramen}), referred to as \textbf{LBM finetuned}, 2) \textbf{TRI-Ramen} only, referred to as \textbf{LBM finetuned [TRI-Ramen]}, 3)  50\% of all tasks present in the \textbf{TRI-Ramen} dataset, referred to as \textbf{LBM finetuned [TRI-Ramen-50\%]}, and 4) 25\% of all tasks present in the \textbf{TRI-Ramen} dataset, referred to as \textbf{LBM finetuned [TRI-Ramen-25\%]}. The results are shown in Figure~\ref{fig:unseen_tasks_fractional}; each line represents pretraining with a different fraction of data then finetuned, or the single-task baseline. For single-task, the graph starts from 15\% of the finetuning data because without data (0\%) we do not have a single-task policy.  Note that the performance of the single-task and finetuned LBM are consistent with the previously reported results in Sec.~\ref{subsec:LBMs_unseen_tasks} and specifically in Fig.~\ref{fig:unseen_tasks_sim_and_real}. Similarly to Figures~\ref{fig:unseen_tasks_sim_and_real} and ~\ref{fig:unseen_tasks_sim_and_real_ds}, each point on the X-axis indicates the percentage of total available demonstration data used to finetune the LBM. 

We note consistent separation with statistical significance between LBM finetuned and the TRI-Ramen version when using 0\%, 15\% and 50\% finetuning data. Interestingly, when using 15\% finetuning data, all five models are separable with statistical significance, with performance steadily increasing as we add more pretraining tasks. The same trend holds when using 50\% and 100\% pretraining data, and all models achieve the best performance when pretraining and finetuning with all available data. 

\textbf{Pretraining and finetuning data tradeoff:} This experiment suggests that there is a tradeoff between pretraining and task-specific finetuning. If there is limited task-specific data for finetuning,  more tasks/data in the pretraining dataset corresponds to  better finetuned LBM performance, even if the data is diverse (here the OXE dataset). Conversely, if there is a lot of task-specific data, LBMs pretrained with less data might suffice. 

%% file: figures_latex/composite_unseen.tex
\adaptivefigure{

    \centering
    \includegraphics[width=\textwidth]{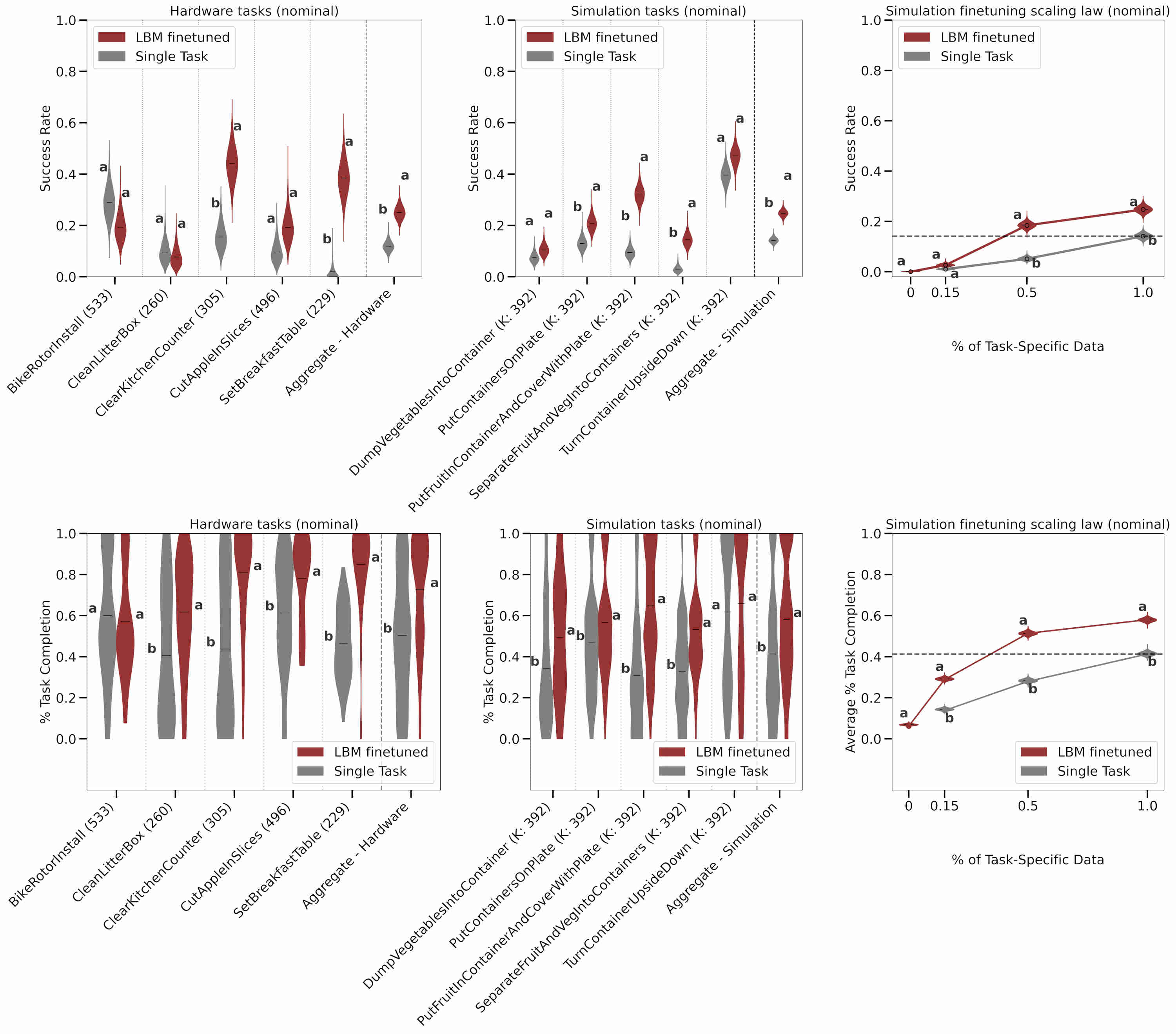}


  \caption{\small\textbf{LBM performance on ``unseen" tasks in real-world and in simulation evaluated under nominal conditions. We compare the single-task baseline with LBMs after finetuning. The top row shows success rate results while the bottom row shows task completion results.}The x-axis labels show the task name, scenario name (for simulation tasks), and the number of demonstrations. Violin plots for SR represent Bayesian posteriors of success rates under a uniform Beta prior and the observed success/failure
data. For TC, violin plots of individual tasks (left) represent the entire data distribution; we use statistical hypothesis tests over the mean TC for the CLD letters shown for these plots, and we provide the Bayesian posteriors used for the tests in Fig.~\ref{fig:individual_task_progress_no_ds_violin}. The violin plots for TC as a function of percentage of data (right), the plots represent the Bayesian posterior of
the mean TC under a uniform Dirichlet prior. 
Policies labeled with different letters are statistically distinguishable. 
}
  \label{fig:unseen_tasks_sim_and_real}
}


%% file: figures_latex/composite_unseen_ds.tex
\adaptivefigure{

    \centering
    \includegraphics[width=0.8\textwidth]{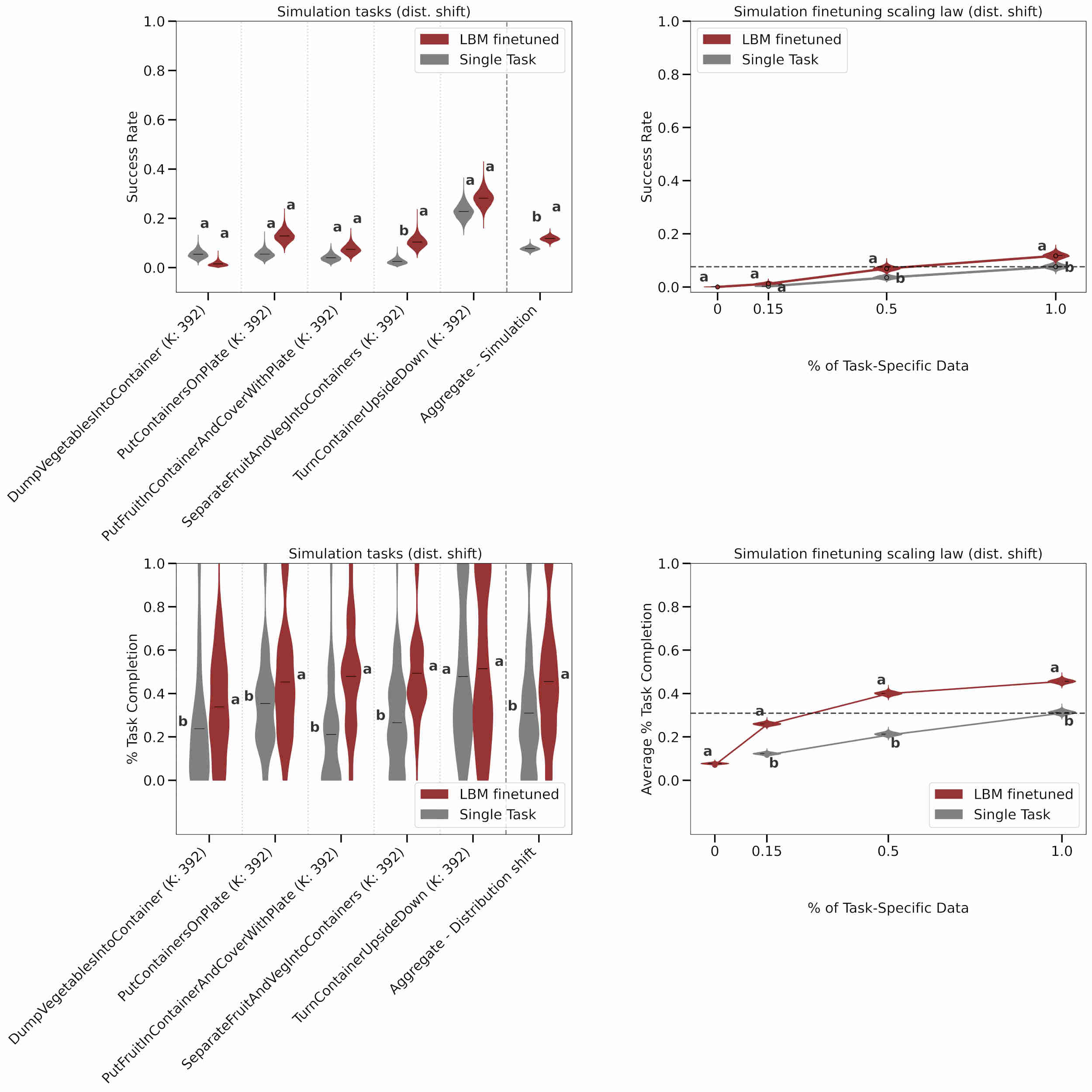}

  \caption{\small\textbf{LBM performance on \textit{unseen} tasks in simulation evaluated under distribution shift. We compare the single-task baseline, with LBMs after finetuning. The top row shows success rate results while the bottom row shows task completion results.} The x-axis labels show the task name, scenario name (for simulation tasks), and the number of demonstrations. Violin plots for SR represent Bayesian posteriors of success rates under a uniform Beta prior and the observed success/failure
data. For TC, violin plots of individual tasks (left) represent the entire data distribution; we use statistical hypothesis tests over the mean TC for the CLD letters shown for these plots, and we provide the Bayesian posteriors used for the tests in Fig.~\ref{fig:individual_task_progress_with_ds_violin}. The violin plots for TC as a function of percentage of data (right), the plots represent the Bayesian posterior of
the mean TC under a uniform Dirichlet prior. 
Policies labeled with different letters are statistically distinguishable. 
  }
  \label{fig:unseen_tasks_sim_and_real_ds}
}

%% file: sections/06_evaluation_protocol.tex
\section{Materials and Methods}
In this section we describe our evaluation protocol, the architecture we use to train the models, details regarding the experiments, the tasks, and the data we use to train and evaluate LBMs.

\subsection{Evaluation Protocol and Analysis}
\label{subsec:eval_protocol_and_analysis}
One of the main contributions of this paper is our focus on rigorous robot policy evaluation that goes beyond what is typically done in the robot learning community. 
We designed an evaluation protocol that  aims to ensure repeatable yet diverse initial conditions, and fairness across policy candidates. We evaluate our hypotheses in both simulation and the real world using statistical tools; this section describes the protocol we used, and the statistical methods we employ to test our hypotheses.

\subsubsection{Policy comparison protocol}
\label{subsec:eval_randomized_conditions}
We focus on fair comparison across the different policies (single-task models vs LBMs), that is, we make every effort to minimize bias and subject the different policies to similar testing conditions, including environmental and initial conditions. 
To do so, in all our evaluations, the evaluators did not know which policy was being tested (blind testing), policy ordering was randomized to maximize fairness for each policy, and the initial conditions were consistent across the policies, within human error for hardware~\cite{kress2024robot}.


In simulation, we use the same simulation parameters and initial conditions (set via random seed) for all policies being compared. Initial conditions are sampled from the same distribution from which associated training data was generated, unless we are explicitly testing out-of-distribution generalization, but these initial conditions are new samples from the training distribution---we do not reuse exact initial conditions between training and evaluation in simulation.

To ensure fairness when evaluating policies on hardware, we split evaluation into test bundles with the size of each bundle equal to the number of policies being compared. Each bundle corresponds to a single initial condition with a randomized policy ordering. 
After evaluating each policy, the initial conditions are reset until the bundle is completed and a new initial condition is evaluated. Real-world evaluation is blind--we ensured that the evaluator had no knowledge of which policy was being evaluated during each run. As for ensuring consistent test conditions, running policies in bundles with randomized order mitigates the effects on policy performance from environment changes (e.g., lighting). For initial conditions, we created a workflow where the robot evaluator was given an image overlay of the desired visual scene (see Fig.~\ref{fig:initial_conditions}); by matching the robot's environment to the overlay, we mitigate the uncertainty in the initial conditions. The initial conditions come from simulation if the task and conditions are modeled in simulation, or from real pictures of initial robot scenes in the case of real-world tasks. We provide additional details for setting up initial conditions in simulation in Section~\ref{subsec:sim_sample_ic} and for real-world in Section~\ref{subsec:supp_hardware_ic_sampels}.

\adaptivefigure{
\centering
\includegraphics[width=\textwidth]{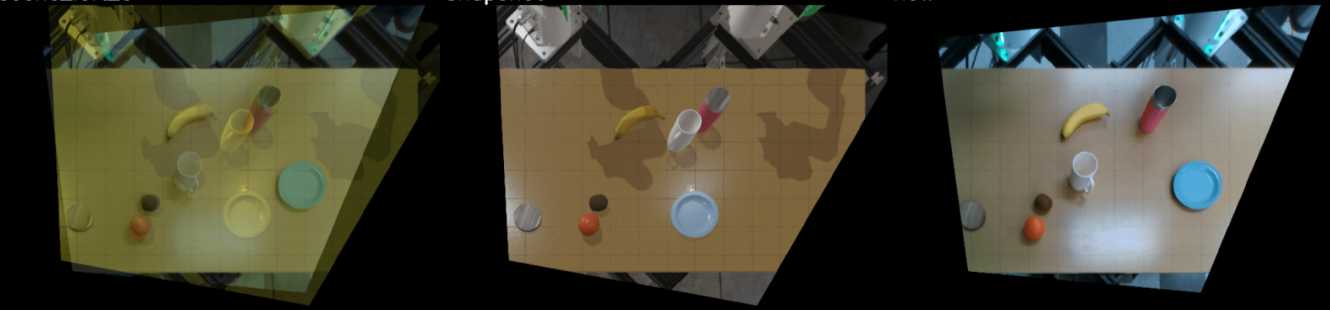}
\caption{\small\textbf{Initial conditions overlay.} \textbf{Left:} overlay of current camera image and desired initial condition for the experiment. \textbf{Middle:} desired initial condition, created using simulation or from pictures of initial robot scenes; the operator will try to recreate this by manipulating the scene elements present. \textbf{Right:} current camera view.}
\label{fig:initial_conditions}
}

\subsubsection{Rubrics and predicates}
\label{subsec:rubrics_and_predicates}
Rubrics are a set of questions that the robot evaluator fills out as they are observing the robot behavior. The questions address task progress (e.g., ``robot grasped the apple") and failures (e.g., ``robot dropped apple''). The questions are binary yes/no questions that can then be statistically analyzed to enrich our understanding of policy performance. The rubrics include a set of milestone questions per task, where each milestone is a step necessary towards a successful rollout. We calculated a Task Completion rate for the real-world rollouts by assigning a +1 credit to a successfully completed milestone and 0 otherwise, then the credits are summed and divided by the number of milestones. 

Policy evaluation in simulation is automated through predicates that are defined over the simulation state (e.g., a predicate that is True when the apple center is closer than a small threshold to the bin center). A success in simulation corresponds to a set of predicates being True. These predicates also allow us to analyze partial success and failure modes of the policy. 

For more details and example rubrics and predicates, refer to Sections~\ref{subsec:supp_rubrics_hardware} and~\ref{subsec:supp_predicates_simulation}.

\subsubsection{Rubric QA}
\label{subsec:rubric_qa}
Considering that the rubric results are provided by reviewers who are prone to human error
, we estimated the discrepancy percentage through an additional validation on a smaller subset of the reported rubric answers. To this end, we conducted a quality assurance (QA) round on $\sim27\%$ of nearly 2700 
real-world evaluation rollouts\footnote{Not all real-world rollouts are included in the results, since some came from earlier iterations. All rollouts that were QAed were from evaluation tasks.} to estimate the discrepancy in the answers. The subset of people doing rubric QA was separate from the robot evaluators. The QA success rate discrepancy was 2.31\% and the overall rubric question discrepancy was 6.25\%. 

We performed one round of corrections, based on the discrepancy between the recorded success and the success  calculated based on the rubrics. This involved updating five rollouts. We note that the results reported in Section~\ref{sec:results} were calculated after the corrections were applied. 

\subsubsection{Statistical Analysis: Performance Characterization of Individual Policies}
\label{subsubsec:stats}


Throughout the paper, we report statistical uncertainty of empirical performance of individual policies. Our  analysis is based on binary success/failure and on task completion  for more challenging, longer-horizon tasks.
\paragraph{Binary Success/Failure Criteria}
From the statistical perspective, computing the success rate of a policy is equivalent to estimating the Bernoulli parameter $p$ of the underlying Bernoulli distribution generating the success/failure labels. This assumes that each evaluation trial is independent and identically distributed (i.i.d.). In simulation, this assumption is satisfied by randomizing the initial conditions with seeds. For real-world experiments, we take a set of measures discussed in \cref{subsec:eval_randomized_conditions} to mitigate the effect of time-varying randomness and unwanted bias as much as possible.

In some cases, we report statistical results that are aggregated over multiple tasks. 
Strictly speaking, this violates the i.i.d. assumption as the number of trials from each task is fixed \textit{a priori} instead of a random draw from a uniform distribution over the 3 tasks. Nevertheless, the effect of this violation is negligible in practice 
and is recommended practice in prior work on policy evaluation~\cite{agarwal2021deep}.
To visualize the uncertainty over the unknown parameter $p$, we perform Bayesian analysis and compute the posterior after observing success/failure data. Specifically, we use a uniform prior over the Bernoulli parameter as suggested by~\cite{kress2024robot} and plot the Bayesian posterior in the form of a violin plot. See \cref{fig:seen_tasks_sim_and_real} for an example.
We use violin plots rather than standard Confidence Intervals (CIs) for two reasons. First, violin plots depict the entire distribution of the parameter $p$ instead of a single interval, thus presenting richer information on statistical uncertainty.
Second, CIs can be confusing or misleading when it comes to policy comparison; one may wrongly conclude that two results are not separated with statistical significance if their corresponding CIs overlap~\cite{greenland2016statistical}. In fact, CIs can overlap while more powerful hypothesis tests may still statistically separate them. In \cref{sec:comparison}, we discuss how we can incorporate such hypothesis tests in our analysis to accompany the individual Bayesian analysis.
\paragraph{Task Completion Criteria} For the task completion criteria based on our rubrics or predicates, the mean of the corresponding categorical distribution may not represent the entire distribution, unlike the Bernoulli case (i.e., variance and other higher-order moments are not uniquely determined by the mean). Therefore, we present the violin plots of the raw data instead of the Bayesian analysis of the mean so they illustrate the full distribution better; see \cref{fig:unseen_tasks_sim_and_real} (bottom row, left and middle) for an example. We will resort to the mean of the distribution only to a) examine the effect of fractional fine-tuning (\cref{fig:unseen_tasks_sim_and_real}, bottom row, right) and b) compare the performance of two or more policies. For a), we use a uniform Dirichlet prior, similar to the binary success/failure setting. For b), we discuss the details of policy comparison below.

\subsubsection{Statistical Analysis: Performance Comparison of Multiple Policies}
\label{sec:comparison}
We perform hypothesis tests to compare multiple policies; we run $k (k - 1) / 2$ pairwise tests where $k$ is the number of policy models to be compared at once. In each test, the confidence level is adjusted for multiplicity via Bonferroni correction unless otherwise noted, so that a global 95\% confidence level is maintained across all the $k (k - 1) / 2$ comparisons. 
We use the Compact Letter Display (CLD) algorithm~\cite{piepho2004algorithm} to summarize the results. With CLD, each policy is labeled with one or more letters such as ``a”, ``ab” or ``bc”. Two policies that do not share the same letter are separated with 95\% confidence. 


For each pairwise test, we use an appropriate statistical method depending on the type of the data. For the more common binary success/failure criteria, we use a sequential hypothesis testing framework as suggested in~\cite{snyder2025your}, which sequentially compares paired outcomes of each evaluation trial until either a decision is reached or all the trials are consumed. Specifically, we adopt the test originally proposed in~\cite{lai1988nearly}, which yields reasonable statistical power in the small sample size regime while maintining a strict Type-I error control for binary data. (That is, the chance of falsely concluding that the two policies are statistically separated is upper-bounded by 5\%). For the task progress, we cannot apply~\cite{lai1988nearly} as it does not readily extend to categorical data. Instead, we use the Welch's t-test~\cite{welch1947generalization} to compare the means of two distributions. Strictly speaking, the discrete nature of the data violates the underlying assumption of the t-test that the data is normally distributed. Thus, the Type-I error is not controlled. However, we ensure that the sample size is sufficiently large (i.e., about 50 or more) so that the degree of violation is small owing to the central limit theorem. 

Nominally, the aforementioned policy comparison procedure is performed per task. When multiple tasks are considered and plotted at once, we do not further adjust the confidence level of individual pairwise tests, since doing so would yield individual confidence levels that are too stringent. Therefore, we note that the Type-1 error is not globally controlled across tasks when they are presented in one plot. 
Nevertheless, we do aggregate the results over tasks when we compare the overall multi-task performance of different policies. 
This yields an exchangeable Bernoulli sequence, approximating an i.i.d. sequence by marginalizing over skills. Exchangeability is weaker than i.i.d., so the underlying assumption of the pairwise test is slightly violated. However, we empirically verified that this difference does not affect the statistical validity of results.

%% file: sections/04_method.tex
\subsection{Large Behavior Models}
\label{sec:lbm_arch}

\adaptivefigure{
    \centering
    \includegraphics[ width=\textwidth]{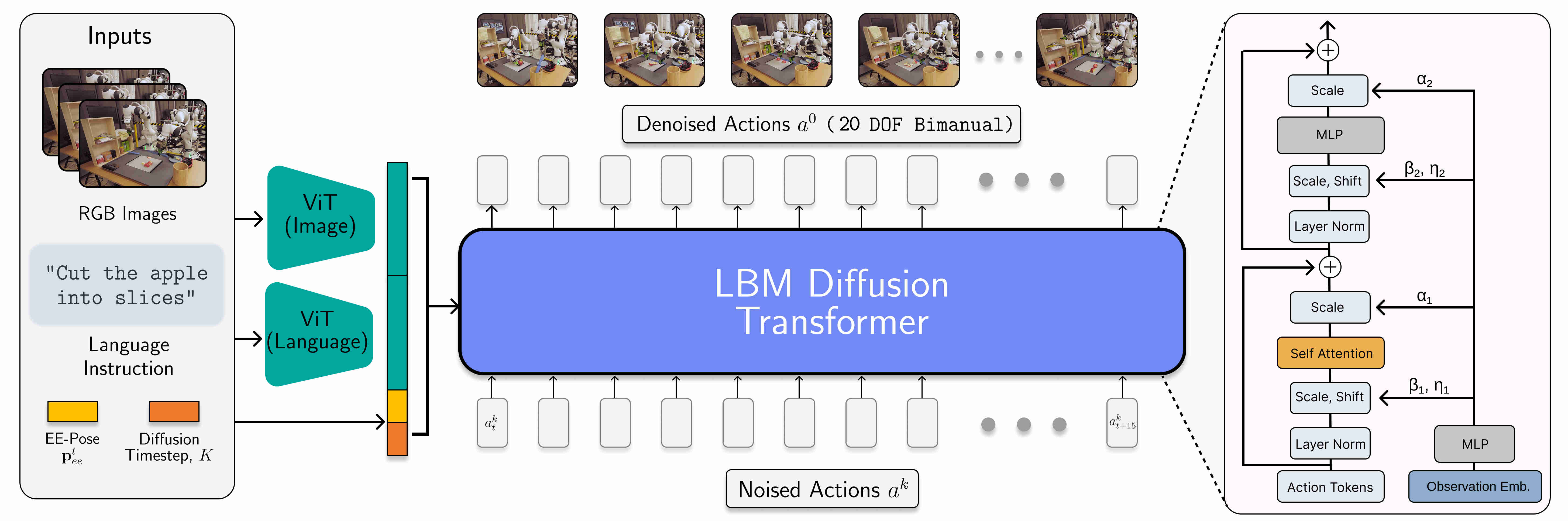}
    \caption{\small\textbf{LBM Architecture:} we use a Diffusion Transformer~\cite{peebles2023scalablediffusionmodelstransformers} conditioned on language, vision and proprioception and which outputs 20-dimensional actions over 16 timesteps (see Section~\ref{subsec:policy_arch} for details). During deployment, we run the policy at 10 Hz and the robot executes the predictions over the first 8 future timesteps before replanning actions.}
    \label{fig:lbm_architecture}
}

In this section we describe the LBM generative model, training objective, architecture, pretraining and finetuning recipes, and deployment details; see Figure~\ref{fig:lbm_architecture} for an overview of our architecture.

\subsubsection{Diffusion for Visuomotor Control}
We implement generative policies for visuomotor control by employing Denoising Diffusion Implicit Models (DDIM)~\cite{song_denoising_2022}. We choose this class of generative model because it has been shown to be effective at learning visuomotor manipulation policies from human demonstrations~\cite{chi2024diffusionpolicy}. DDIMs transform a simple prior distribution, typically Gaussian noise, into a complex, structured (action) distribution conditioned on input data -- in our case, visual, proprioceptive, and language observations. This transformation uses a deterministic sampling process derived from denoising diffusion probabilistic models (DDPM)~\cite{ho_denoising_2020}. Given $K \geq 1$ denoising steps, we start with a noise sample $A_t^K \sim \mathcal{N}(0, I)$ at time $t$ and use DDIM to denoise it into a continuous action $A_t^0$ in $K$ iterative steps. In order to predict actions conditioned on observation inputs, we modify the original DDIM update as follows:
\begin{equation}
A_{t}^{k-1} = \alpha \left( A_{t}^{k} - \gamma \cdot \epsilon_\theta(O_t, A_{t}^{k}, k)\right)
\end{equation}
\noindent where $A_{t}^k$ is a set of noisy actions at the $k$-th denoising step, $O_{t}$ are the observations, and $k$ is the diffusion timestep. Parameters $\alpha$ and $\gamma$ are determined by a noise schedule which varies with the diffusion timestep $k$, and $\epsilon_{\theta}$ is the noise-prediction neural network with weights $\theta$. 
To train $\epsilon_{\theta}$, we sample an action \(A_t^0\) and a random step \(k\), and add a step-dependent Gaussian noise \(\varepsilon_k\) to form a noisy action \(A_t^k=A_t^0+\varepsilon_k\). The network is then trained to predict \(\varepsilon_k\) from \(A_t^k\), enabling it to denoise across the full range of diffusion levels. Mathematically, we optimize the following DDPM loss with respect to $\theta$:
\begin{equation}
\mathcal{L}(\theta) = \| \epsilon_k - \epsilon_\theta \left( O_t, A_t^k, k \right) \|^2_2.
\end{equation}

\subsubsection{Policy Architecture}
\label{subsec:policy_arch}
We parametrize the noise prediction network, $\epsilon_{\theta}$, as a Diffusion Transformer (DiT) \cite{peebles2023scalablediffusionmodelstransformers}, which conditions on features extracted from the observations and the diffusion timestep in predicting actions (see Section~\ref{subsec:obs_act_spaces} for the observation and action spaces). 
To extract features from the image observations we use the CLS token output from a pretrained CLIP Vision Transformer (ViT) backbone~\cite{radford2021learningtransferablevisualmodels}. Language features are similarly computed from the task description using a CLIP text encoder, with a projection layer on top of the pooled End of Sequence token. The language and visual features are concatenated with the proprioception for each observation timestep as well as with the diffusion timestep, $k$, which is encoded with a sinusoidal positional embedding~\cite{ho_denoising_2020} followed by a two-layer MLP. During training, we finetune through the visual feature extractor, which is shared across all camera inputs. We keep the language-feature extractor frozen, but train a projection layer on top of the language features.

The DiT conditions on two timesteps of concatenated observation features, which together have a size of $6,732$, and the encoded diffusion timestep via an adaptive layer norm (adaLN) MLP \cite{peebles2023scalablediffusionmodelstransformers}. This model consists of eight DiT blocks with an embedding size of 768. The network predicts 16 timesteps of 20-dimensional actions, for a total output size of $A_t$ = 320. All experiments use the architecture described above. In the case of finetuning on single-task data or evaluating from-scratch policies we use the same architecture and only use the language prompts for the task of choice. 

\subsubsection{Training and Deployment}
Our training recipe follows a common pattern for foundation models where we first pretrain policies on the full data mixture and then finetune on narrower data subsets~\cite{touvron2023llama1,liu2023llava,openai2024gpt4technicalreport,kim_openvla_2024,black_0_nodate}. 
Hyperparameters for both stages are summarized in tables \ref{tab:opt_hparams} and \ref{tab:imgaug_arch_hparams}.

We pretrain on the full dataset mixture described in Section~\ref{subsec:data}. During training, we first resize images to 256x342, then randomly crop and apply color jitter, which yields 224x224 images. We train for 48k steps with a global batch size of 2560 with a constant learning rate of 3e-4. The vision encoder uses a learning rate one tenth that of the rest of the model. 

We finetune the pretrained policy on demonstrations from individual tasks. 
We found that the optimal checkpoint generally occurred earlier in training for simulated tasks than for real tasks. As a result, we finetune for 30k steps for real tasks and for 10k steps for simulated tasks with a global batch size of 320 and a reduced learning rate of 2e-5. We leave co-training, alternate learning rate schedules, and strategies for selecting optimal pretraining and finetuning checkpoints to future work. During finetuning, we use the same image augmentation hyperparameters as during pretraining. 

While we compute the loss on 16 action steps during training, during deployment we only execute eight timesteps before recomputing actions~\cite{chi2024diffusionpolicy}. As in training, images are resized to 256x342, but then center-cropped to 224x224 rather than randomly cropped. The policy loop executes at a rate of 10 Hz.

\adaptivetable{
\centering
\begin{tabular}{|l|c|c|c|c|c|}
\hline
\textbf{Hyperparameters} & \textbf{Pretraining} & \textbf{Finetuning (Real)} & \textbf{Finetuning (Sim)} & \textbf{Single-Task (Real)} & \textbf{Single-Task (Sim)}  \\
\hline
Learning Rate & 3e-4 & 2e-5 & 2e-5 & 2e-5 & 2e-5 \\
Global Batch Size & 2560 & 320 & 320 & 320 & 320 \\
Steps & 48000 & 30000 & 10000 & 100000 & 25000 \\
LR Schedule & Constant & Constant & Constant & Constant & Constant \\
\hline
\end{tabular}
\caption{\small Hyperparameters used during pretraining, finetuning, and with single-task models. }
\label{tab:opt_hparams}
}

\begin{table}[h!]
\centering
\begin{tabular}{|l|c|}
\hline
\textbf{Hyperparameters} & \textbf{Values} \\
\hline
\multicolumn{2}{|c|}{\textbf{Image Augmentation}} \\
\hline
Resize (HxW) & 256x342 \\
Random Crop (HxW) & 224x224 \\
Brightness Range & [0.8, 1.2] \\
Contrast Range & [0.6, 1.4] \\
Saturation Range & [0.8, 1.2] \\
Hue Range & [-0.05, 0.05] \\
\hline
\multicolumn{2}{|c|}{\textbf{Model Architecture}} \\
\hline
DiT Layers & 8 \\
DiT Embedding Dimension & 768 \\
CLIP Image Encoder & ViT-B/16 \\
\hline
\end{tabular}
\caption{\small Image Augmentation and Model Architecture Hyperparameters}
\label{tab:imgaug_arch_hparams}
\end{table}

%% file: sections/05_experimental_setup.tex
\subsection{Experimental Details}

\adaptivefigure{
\centering
\includegraphics[trim=20 290 20 10, clip, width=0.8\textwidth]{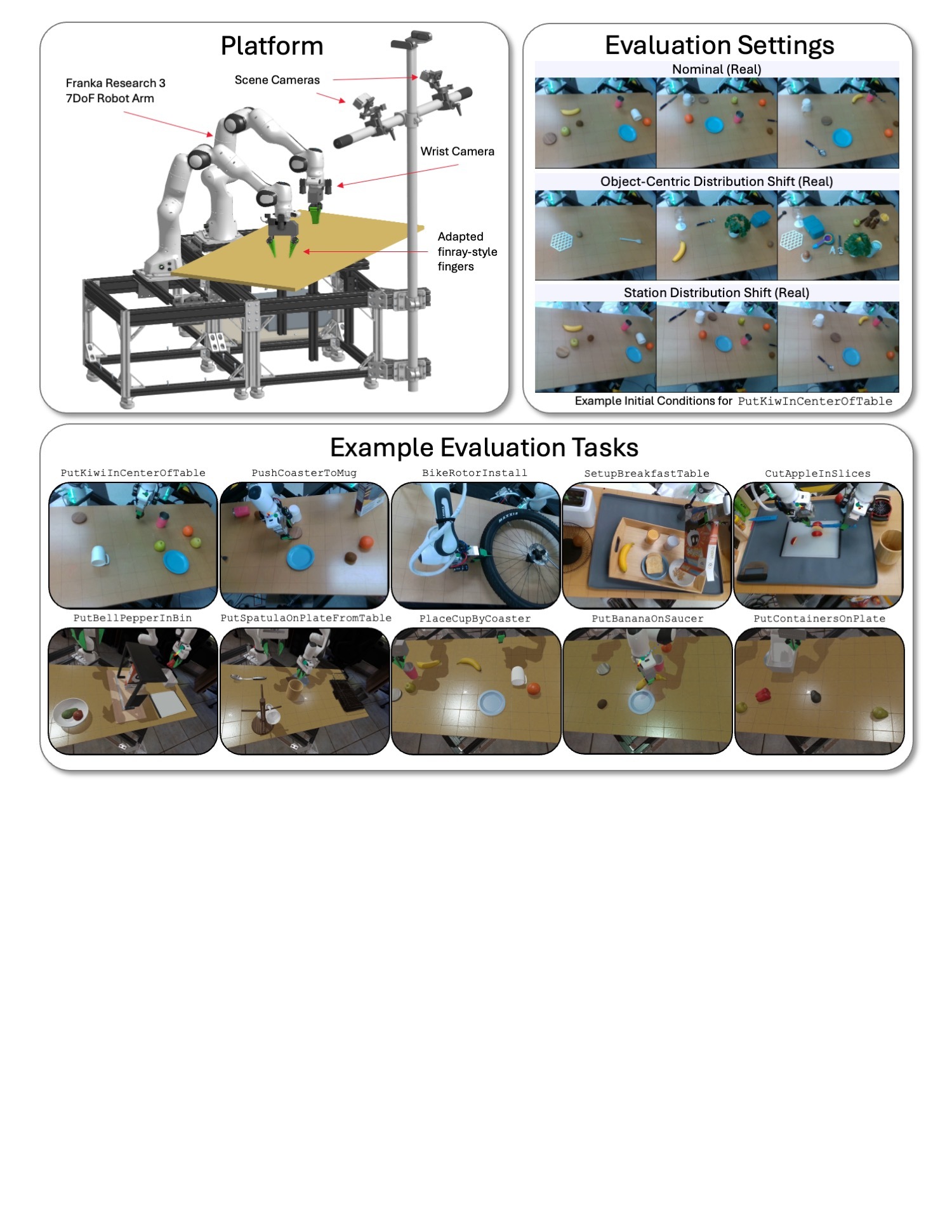}
\caption{\small\textbf{Experiment Overview}. To study the performance of LBMs, we conduct extensive experiments in both simulation and the real world on a bimanual table-top setting (top left), reporting results under various types of distribution shifts such as object-centric and station (top right). Our evaluation suite includes a diverse array of tasks, including short tasks such as \textit{PutKiwiInCenterOfTable} and long horizon multi-step dexterous tasks such as \textit{CuptAppleIntoSlices} (bottom). The diversity of our evaluation suite allows us to study LBMs under a wide range of conditions. Each column in the top right figure shows the same indexed initial condition under three types of real-world settings. For a full description of the different evaluation settings including simulation (not pictured), refer to Sec.~\ref{subsec:eval_tasks}. The platform varies slightly from the image shown in this figure for different scenarios; see Sec.~\ref{subsec:platform} for an overview. Representative images of example tasks are are not all from the same relative scene camera.
}
\label{fig:experiment_overview}
} 

\subsubsection{Platform}
\label{subsec:platform}
We focus here on tabletop bimanual manipulation using two Franka FR3 robot arms with parallel gripper and TRI's finray-style fingers \cite{DBLP:journals/firai/CrooksVOMR16}. See Figure~\ref{fig:experiment_overview}, top-left, for an overview of the physical platform. We had one major hardware upgrade that changed the gripper model and wrist camera configurations; see Sec.~\ref{subsec:supplemental_station_details} for more details. We collected training data on both platforms, and all real-world evaluations presented here are done on the new platform. We used a total of nine robot stations, see details in Table~\ref{tab:table_of_stations}. We run the Franka robots using our custom joint impedance controller, with a differential inverse kinematics controller on top to translate end-effector relative SE(3) commands \cite{chi2024universalmanipulationinterfaceinthewild} from the human teleoperator or the LBM policy. The differential inverse kinematics controller also handles collision avoidance. For the new platform, we use the WSG50-110 gripper on each arm. The workspace contain two FRAMOS D415e scene cameras and each wrist has two FLIR Blackfly S BFS-PGE-23S3C-CS cameras. All policies are run at 10 Hz. 

\subsubsection{Simulation}
\label{subsec:simulation}

Our primary usage for simulation is for evaluation. One of the main bottlenecks for iterating on policy design is evaluation both in terms of throughput and level of control. Relying on real-world testing alone for enough samples to support any meaningful statistical analysis is prohibitively expensive and time consuming. To address this, we have been developing our simulation benchmark, \texttt{lbm\_eval}, which is built on top of
Drake~\cite{drake}. \texttt{lbm\_eval} uses curated
assets and human-authored scenarios and tasks. Each simulation is
deterministic given a random seed (however, the GPU-based policies do not have
the same guarantee). 

Our simulation policy evaluations are run against four scenarios, all of which differ in their affordance for task complexity, object types, and object count (see Figures ~\ref{fig:PlaceFruitFromBowlIntoBin_ICs_nominal_vs_dist_shift}, ~\ref{fig:SpatulaInCrock_ICs_nominal_vs_dist_shift} ~\ref{fig:PushCoasterToCenterOfTable_ICs_nominal_vs_dist_shift} and ~\ref{fig:scenario4_ICs_nominal_vs_dist_shift}). All scenarios contain a workspace modeled after a real-world robot station (see station details in Table~\ref{tab:table_of_stations}). All four scenarios are inspired by kitchen settings and focus on food preparation and organization. 
All four scenarios contain task-relevant manipulands (objects that are manipulated as part of the task) in addition to varying numbers of distractors. The first two scenarios are modeled after the old hardware platform, and the later two are modeled after the new hardware platform.

\textbf{The \textit{DryingRack (D)} scenario} contains plates, mugs, and spatulas, in addition to a mug holder, utensil crock, and a dish drying rack. This scenario has 13 tasks. \textbf{The \textit{Shelf (S)} scenario} contains various types of fruits and vegetables, a shelf, fruit bowl, bin, cereal box, and a cutting board. There are 12 tasks in this scenario. \textbf{The \textit{Breakfast (B)} scenario} contains various types of fruits, a mug, plate, coaster, and a cup. This scenario is designed specifically for testing language conditioning since the initial conditions for different tasks visually appear the same. There are 18 tasks in this scenario. \textbf{The \textit{Kitchen (K)} scenario} contains bins, various types of fruits and vegetables, and a plate. Scenario \textit{K} has 5 tasks, and is the only scenario that is entirely absent in the pretraining dataset, as described in Section~\ref{subsec:data}. All tasks in this scenario are long horizon, and some require nonprehensile manipulation or 
understanding of semantic attributes (e.g., vegetable vs fruit, large vs small). Scenarios \textit{B} and \textit{K} are somewhat simpler in terms of scene complexity as compared to the first two scenarios. A fifth \textbf{scenario (0)} is used for debugging policies early on during training and consists of just one task where the robot has to move a box to the center of the table. This scenario is not used as part of evaluation, but its demonstration data is included in the pretraining dataset.

\subsection{Pretraining data}
\label{subsec:data}

Our pretraining mixture, called \textbf{Ramen}, consists of a large-scale dataset of robot demonstrations totaling $\sim$1695 hours of demonstration, including high-quality data collected at TRI ($\sim$545 hours; \textbf{TRI-Ramen}) combined with curated external robot data ($\sim$1150 hours; \textbf{OXE-Ramen}). 

\textbf{TRI-Ramen} data consists of a total of 545 hours of real data over 532 tasks for a total of 64,262 demonstrations. This is made up of: \textbf{TRI-Ramen-Real} - 468 hours, 362 tasks and 46063 demonstrations collected across 9 hardware stations; \textbf{TRI-Ramen-Sim} - 45 hours, 41 tasks and 7348 demonstrations collected across 2 simulation stations; and \textbf{TRI-Ramen-UMI} (32 hours, 129 tasks, 10851 demonstrations) collected with the Universal Manipulation Interface \cite{chi2024universalmanipulationinterfaceinthewild} using 7 pairs of handheld devices in ``in-the-wild" environments.

\textbf{TRI-Ramen-Sim} tasks used in the pretraining set exclude all five tasks from scenario \textit{K}, and one task each from scenarios \textit{D}, \textit{S} and \textit{B};  results of evaluating these ``unseen" tasks can be found in Section~\ref{subsec:LBMs_unseen_tasks}.

For each of the 40 simulation tasks from scenarios \textit{D}, \textit{S} and \textit{B} that are part of the pretaining set, we collect corresponding real-world demonstrations of the same task in the similar environments (subject to hardware differences across fleet). Scenario \textit{B} tasks are collected with matched initial conditions as in simulation. Scenarios \textit{D} and \textit{S} are collected with manually created initial conditions. These are part of \textbf{TRI-Ramen-Real} (i.e., they are also part of the pretraining set). During real-world data collection, we distributed each task's demonstrations evenly across 2 to 4 robot stations, such that each station produces approximately 100 demonstrations. For each task, one of the stations used to collect real data is the robot station the simulated environment is modeled after. 

\textbf{OXE-Ramen} data is a subset of the OpenX-Embodiment datasets. 
The subset was chosen based on a set of heuristics such as object and environment diversity and total number of episodes. The observations and actions were mapped to conform to the \textbf{TRI-Ramen} data format. 
Mappings include standardization of frames of reference and units for end-effector poses and gripper widths, and resizing and cropping images.

During training, we batch balance the \textbf{Ramen} datasets by a set of empirically found weights to ensure each batch contains samples of all the datasets. See Table~\ref{tab:supp_batch_balance_weights} for datasets used in pretraining and associated weights.

\subsubsection{Observation and Action spaces}
\label{subsec:obs_act_spaces}
The observation space includes i) end-effector poses w.r.t. the station base frame (table center), ii) end-effector poses w.r.t. the other end-effector 
, iii) continuous gripper width, iv) 6 RGB images (missing cameras are zero-padded), and v) one natural-language instruction. The action space includes i) end-effector poses w.r.t. the station's base frame, and gripper widths. Orientation is represented as a 6D vector that corresponds to the top 2 rows of the rotation matrix. For observations and actions, we use a similar relative trajectory representation as in \cite{chi2024universalmanipulationinterfaceinthewild}. Additionally, we use a history of observations (\(n_{obs}=2\)) and an action prediction horizon (\(n_{horizon}=15\)) as in \cite{chi2024diffusionpolicy}.  
Unimanual data from \textbf{OXE-Ramen} was converted into bimanual by zero padding the missing arm and randomly swapping the arm's side. 
Each episode in \textbf{TRI-Ramen} contains a list of language instructions that are randomly sampled during inference. Such a list includes one instruction written by a human and five instructions generated by prompting a LLM (ChatGPT) to give alternative versions of the human-generated instruction. Missing language annotations in \textbf{OXE-Ramen} were filled with a generic text: ``do something useful". 

\subsubsection{Data Normalization}
\label{subsec:normalization}

For data in \textbf{Ramen}, normalization is done on a per-feature dimension (e.g., end-effector pose) and per-timestep (i.e., observation history, and action prediction horizon) basis.
Values are normalized to fall within a fixed range of \([-1.5,\,1.5]\), by being scaled by the the 2nd and 98th percentiles, $x^{0.02}$ and $x^{0.98}$, and being clipped beyond the range $[-1.5,\, 1.5]$. For all data samples $x_i \in \mathcal{D}$, we compute the corresponding normalized value $y_i$: 
\[
y_i = \min\left(\max\big(-1.5,\, 2\frac{x_i - x^{0.02}}{x^{0.98} - x^{0.02}} - 1\big), \, 1.5 \right)
\]
This shifts and scales the percentile range of 2 to 98 to lie from $-1$ to $1$ while retaining some outliers, but keeps most of the resolution in the high-density center of the data distribution. Since we represent actions relative to the current time's observations, actions further into the future have a wider spread than the immediate next actions. Computing normalization parameters for each timestep independently better preserves resolution for near-future actions, which are the more important parts to predict accurately. We avoid this normalization procedure for the 6D rotation in the poses to avoid corrupting the rotation matrix. Note that 6D rotation lies in the [-1, 1] range.

The normalization parameters ($x^{0.02}$ and $x^{0.98}$) were calculated independently per data source for \textbf{OXE-Ramen} and \textbf{TRI-Ramen-UMI}. \textbf{TRI-Ramen-Real} and \textbf{TRI-Ramen-Sim} are used together to compute \texttt{lbm\_robot} normalization parameters. At training time, individual datagrams are normalized separately based on their data source. At test time, the \texttt{lbm\_robot} normalizer is used to de-normalize actions for our robots.  Due to an error in the code, some datagrams were normalized incorrectly (with normalization parameters belonging to a different data source) within each batch during pretraining of the models presented in Section~\ref{sec:results}. 
Due to the cost of performing extensive real-world evaluation, we did not repeat the experiments in Section~\ref{sec:results}, but rather performed a smaller scale experiment on ``seen" tasks in simulation to assess the impact of the incorrect normalization. 
Fig. ~\ref{fig:stage5_bug_fix_comparison} 
 shows that the difference is small under nominal conditions, and the LBM with the correct normalization parameters performs better under distribution shift. %

\subsubsection{Dataset filtering}
\label{subsec:pretraining_dataset_filtering}
The TRI-Ramen dataset consists of a number of low-motion frames at the start of certain demonstrations. This is due either to operator error or to the teleoperation UI being loaded more slowly than the start of the demonstration logging. We implemented a simple filtering operation, defining a motion threshold to capture when the gripper moved either more than 5~cm in translation or 15~deg in rotation with respect to its starting pose. We then remove all data from the start of the demonstration until the motion threshold is met. 

In simulation, we analyze the effects of filtering out this data and find that, when trained with unfiltered 
data, the single-task and LBM policies exhibit difficulty to initiate motion at the beginning of each rollout. The severity of this symptom is policy, task and evaluation conditions (nominal vs distribution shift) dependent. Filtering the low motion data improved single-task performance in simulation; however, it led to a surprising decrease in performance for pretrained LBM, where we observed that it would commit to some uncommanded task more often than before. 
We therefore made the design choice to pretrain with unfiltered data, but finetune LBMs and train single-task policies using the filtered dataset in simulation tasks. We performed an analysis of training exclusively with filtered data, and our findings are shown in Section~\ref{sec:supp_pauses}. Due to the cost of real-world evaluation, we did not study this phenomenon on real world tasks, using the unfiltered version of task-specific finetuning data.

\subsection{Evaluation Tasks}
\label{subsec:eval_tasks}
There are two types of tasks for evaluation, ``seen" and ``unseen", depending on their presence in the \textbf{TRI-Ramen} dataset during pretraining. For simulation evaluation, we selected 16 ``seen" tasks randomly from \textbf{TRI-Ramen-Sim} and 8 ``unseen" tasks, all part of \texttt{lbm\_eval}. 
For real-world evaluation, we selected 3 ``seen" tasks and 5 ``unseen" tasks, where the 3 ``seen" tasks are part of the simulation ``seen" tasks, i.e., they have matching simulation and real data demonstrations, as described in Section~\ref{subsec:data}. The task names and number of demonstrations collected are listed in Tables~\ref{tab:suppl_sim_eval_data_amount} and~\ref{tab:suppl_real_eval_data_amount}, while example evaluation tasks shown in Figure~\ref{fig:experiment_overview}, bottom.

We measure performance under several distinct conditions, including nominal in-distribution conditions, as well as conditions exhibiting distribution shift. For each task, we test 200 initial conditions in simulation and 50 in real world. See Appendix ~\ref{subsec:sim_sample_ic} and ~\ref{subsec:supp_hardware_ic_sampels} for sample initial conditions in simulation and in the real world. 

\subsubsection{Simulation}
\label{subsubsec:sim_tasks_and_data}
We select a small subset of our simulation tasks to use for evaluation, covering four scenarios. All tasks in scenarios \textit{K} as well as many in \textit{D} and \textit{S} are designed specifically to be visually ambiguous given their initial conditions, requiring policies to rely on language conditioning to determine what actions to execute. We consider two conditions for simulation experiments: \textbf{Nominal (Sim)} and \textbf{Distribution Shift (Sim)}.

\textbf{Nominal (Sim)}: Under this setting, object poses as well as quantities are drawn from predefined distribution with rejection sampling to obey constraints such as no interpenetration. Additionally, in-distribution scene lighting parameters for a single directional light source are also drawn from predefined distributions. Lighting parameters include color (over the HSV color wheel), intensity, and direction. In addition to the single parameterized light source, a fixed environment map provides additional ambient lighting to the scene.

\textbf{Distribution Shift (Sim)}:
Most of \texttt{lbm\_eval}'s distribution shift is implemented to test policy robustness against appearance changes. For lighting, we define a secondary directional light source with intensity and direction parameters drawn from shifted distributions. Additionally, the environment map is drawn from a discrete set of twenty choices unseen during training. Scene-camera extrinsics and intrinsics are also randomized. Finally, alternate textures and colors are sampled for objects as well as the table top. The level of distribution shift for colors and textures varies, depending on the specific scenario. In addition to pure appearance variations, we also introduced random distractor objects in scenario \textit{B}. More details are presented in Section~\ref{subsec:sim_sample_ic} and Table ~\ref{tab:sim_distribution_types}.

\subsubsection{Real}
\label{subsubsec:real_tasks_and_data}
We test model performance on eight real-world tasks. Three are short-horizon tasks for which both real and simulation data was seen at pretraining. The other five tasks are unseen long-horizon, multistep tasks that require sequencing diverse types of manipulation. For example, the \textit{CutAppleInSlices} task (Figure ~\ref{fig:long_horizon_CutAppleIntoSlices_filmstrips}) requires the robot to use an apple corer to core an apple, retrieve a knife from a crock, unsheath the knife to slice the apple into halves, slice the halves into slices, and finally wipe the knife with a cloth before re-sheathing it and placing it back into the crock. We further design three distinct evaluation conditions\footnote{To determine axes of distribution shift, we informally evaluated earlier experimental models under a wide variety of distribution shifts. We selected modes that were operationally easy to implement, avoiding modes for which a large number of initial conditions would fail and therefore provide a less informative signal.}: \textbf{Nominal (Real)}, \textbf{Station Distribution Shift (Real)}, and \textbf{Object-Centric Distribution Shift (Real)}; see Figure.~\ref{fig:experiment_overview} for an example.

\textbf{Nominal (Real)}: For tasks that have simulation counterparts, the initial conditions of manipulands and distractors are matched to initial conditions generated in simulation via the overlay described in Sec.\ref{subsec:eval_randomized_conditions}. For tasks without simulation counterparts, the original demonstrator of the task created a new set of initial conditions, using the original objects. The tasks are then tested on an robot station with training data coverage. 

\textbf{Station Distribution Shift (Real)}: To test cross-station transfer, we evaluate policies on stations not in the finetuning dataset for the task. For this setting, we use the same initial conditions for objects and distractors as in \textbf{Nominal (Real)}.

\textbf{Object-Centric Distribution Shift (Real)}: We implement two types of object-centric distribution shift in real experiments: manipulands and distractors. Each key manipuland for a given task is tested on at least five novel instantiations (see Figures~\ref{fig:experiment_overview} and~\ref{fig:novel_object_overview}). Evaluation scenes feature various levels of distractor clutter, where distractors were sampled from a set of both seen and novel objects. To the best of our ability, we sourced novel manipulands and distractors that we believe to not be present in any of the pretraining data. In order to create such initial conditions, we start by populating an empty scene with task-specific manipulands using their initial conditions sampled in simulation, then gradually adding different levels of clutter.

%% file: sections/09_conclusion_and_discussion.tex
\section{Discussion and Conclusion}

Large Behavior Models move dexterous manipulation away from task-specific engineering and into a scalable and data-driven paradigm similar to recent progress in language and vision. To rigorously quantify the capabilities of current LBMs, we train a series of models on roughly 1,700 hours of heterogeneous demonstration data and analyze their performance on 1,800 blind A/B-style real-world rollouts and over 47,000 simulation rollouts. 

We find that finetuning LBMs into task-specific specialists consistently outperforms from-scratch training with a given amount of finetuning data or allow achieving from-scratch-equivalent performance with 3-5x less data required. These differences are also amplified under deployment distribution shift---when test-time conditions differ from those encountered during training. This finding is critical because distribution shift is virtually inescapable in real-world use cases and is often omitted from empirical robotics work, masking important information about real-world utility.

We also find that finetuned performance smoothly improves with increasing pretraining data. At the data scales we examined, we find no evidence of performance discontinuities or sharp inflection points.

Interestingly, we encountered mixed results with non-finetuned LBMs. Encouragingly, we found that a single network is able to learn many tasks simultaneously, but we don't observe consistent outperformance of from-scratch single-task training without finetuning. We expect this is partially due to language-steering brittleness of our models with their small language encoders. We've seen promising early signs that larger VLA prototypes overcome some of this difficulty, but more work is required to rigorously examine this effect in higher-language-capacity models.

Our findings largely support the recent surge in popularity of LBM-style robot foundation models, adding to evidence that large-scale pretraining on diverse robot data is a viable path towards more capable robots. However, we also find evidence for caution in the field. Many of the effects we observe were only measurable with larger-than-standard sample sizes and careful statistical testing that is non-standard for empirical robotics. Due to the size of current effects and the magnitude of experimental noise, there is significant risk that many robotics papers are measuring statistical noise due to insufficient statistical power. Additionally, we find that decisions like data normalization have a large effect on downstream performance, often dominating architectural or algorithmic changes; when comparing methods it is critical that these design choices are studied in isolation to avoid conflating the source of performance changes.

\subsection{Limitations} \label{sec:limitations}
One important limitation of our analysis is that we do not explicitly account
for the stochasticity across training when we compare two policy architectures
(this would be very expensive to do with statistical significance).
Specifically, given a fixed dataset, fixed (stochastic) evaluation benchmark,
and a policy architecture, we have \[ p(\text{\small success} | \text{\small dataset}) = \int\!\! 
p_{\text{eval}}(\text{\small success} | w) \, p_{\text{train}}(w | \text{\small dataset}) \, dw,\] where
$w$ are the parameters (weights) of the policy. Our confidence intervals are
computed for the first term, but do not account for the second. 

We made a decision to run 50 real-world rollouts per task per policy per condition, and to further reduce the measurement uncertainty with hardware displays for reproducible initial conditions. The reproducible initial conditions did mean that each rollout took more time; we intend to continue to optimize the evaluation protocol to improve throughput.
Additionally, despite these experimental protocols designed to minimize environment variability and human error, we expect that both initial condition and scoring mistakes are non-zero, likely adding to the noise of our measurements. As a result, our real-world results potentially miss small-magnitude effects due to signal-to-noise limitations.

We also study LBMs with modestly-sized language encoders pretrained via CLIP. While we expect many of our findings will generalize to larger VLAs, we expect that some aspects like language steerability will differ in that setting.

%% file: sections/10_appendix.tex

In the following, we provide additional details that complement the information in the paper. Specifically, we provide the author list as well as roles (Section~\ref{sec:authors});  describe the experimental platforms, both for hardware and simulation (Section~\ref{sec:supp_platform_details}); provide the simulation (Section~\ref{sec:sim_evaluation_details}) and the real-world (Section~\ref{sec:hardware_eval_details}) evaluation details; add additional analysis and insights complementing the results in Section~\ref{sec:results}  (Sections~\ref{subsec:supplemental_breakfast_sceario} and~\ref{sec:supp_bayesian}); and discuss data quality and details (Sections~\ref{sec:supp_pauses} and~\ref{sec:supp_oxe_dataset_details})

\input{sections/10_appendix_authors}

\input{sections/11_appendix_platform_details}

\input{sections/appendix_simulation_eval_details}

\input{sections/12_appendix_hardware_eval_details}

\input{sections/13_appendix_additional_results}

\input{sections/21_appendix_violins}

\input{sections/18_appendix_pauses}

\input{sections/14_appendix_dataset_details}

\input{sections/23_appendix_stage5_bugfix}

%% file: sections/10_appendix_authors.tex
\section{Authors and contributions}
\label{sec:authors}

The authors are sorted alphabetically within each group. Unless otherwise indicated, all authors are affiliated with Toyota Research Institute; email: \texttt{firstname.lastname@tri.global}.

\noindent\textbf{First authors} are primary contributors and made substantial contributions to the work (policy architecture and training, evaluation, simulation): Jose Barreiros, Andrew Beaulieu, Aditya Bhat, Rick Cory, Eric Cousineau, Hongkai Dai, Ching-Hsin Fang, Kunimatsu Hashimoto, Muhammad Zubair Irshad, Masha Itkina, Naveen Kuppuswamy, Kuan-Hui Lee, Katherine Liu, Dale McConachie, Ian McMahon, Haruki Nishimura, Calder Phillips-Grafflin, Charles Richter, Paarth Shah, Krishnan Srinivasan, Blake Wulfe, Chen Xu, Mengchao Zhang.

\noindent\textbf{Second authors} assisted with the work (developing infrastructure, data collection, paper edits and feedback): Alex Alspach, Maya Angeles, Kushal Arora, Vitor Campagnolo Guizilini, Alejandro Castro, Dian Chen, Ting-Sheng Chu, Sam Creasey, Sean Curtis, Richard Denitto, Emma Dixon, Eric Dusel, Matthew Ferreira, Aimee Goncalves, Grant Gould, Damrong Guoy, Swati Gupta, Xuchen Han, Kyle Hatch, Brendan Hathaway, Allison Henry, Hillel Hochsztein, Phoebe Horgan, Shun Iwase, Donovon Jackson, Siddharth Karamcheti, Sedrick Keh, Joseph Masterjohn, Jean Mercat, Patrick Miller, Paul Mitiguy, Tony Nguyen, Jeremy Nimmer, Yuki Noguchi, Reko Ong, Aykut Onol, Owen Pfannenstiehl, Richard Poyner, Leticia Priebe Mendes Rocha, Gordon Richardson, Christopher Rodriguez, Derick Seale, Michael Sherman, Mariah Smith-Jones, David Tago, Pavel Tokmakov, Matthew Tran, Basile Van Hoorick, Igor Vasiljevic, Sergey Zakharov, Mark Zolotas.

\noindent\textbf{Last authors} led the project and are responsible for strategic decisions (method, benchmark, paper writing and presentation): Rares Ambrus, Kerri Fetzer-Borelli, Ben Burchfiel, Hadas Kress-Gazit\footnote{Cornell University, Ithaca, NY 14850, USA.
Email: \texttt{hadaskg@cornell.edu}}, Siyuan Feng, Stacie Ford, Russ Tedrake.

%% file: sections/11_appendix_platform_details.tex
\section{Platform Details}
\label{sec:supp_platform_details}

Table~\ref{tab:table_of_stations} provides a summary of the hardware and simulated robots used for data collection and evaluation. 

\subsection{Robot Hardware}
\label{subsec:supplemental_station_details}

All stations contain a tabletop marked with grid lines (useful for correlating object placement on the real robot stations), two Franka Research 3 arms, and two parallel jaw grippers with custom compliant fingers \cite{DBLP:journals/firai/CrooksVOMR16}. During data collection, we performed one major hardware upgrade where we switched the Franka Hand to a Schunk WSG50-110 gripper; replaced the single FRAMOS D435 wrist camera with dual FLIR wrist cameras per arm; and replaced the fingers with shorter ones. We will refer to the platform with the Franka Hand and single D435 wrist camera as the old platform, which has been retired and was not used for any reported real-world evaluation results.
Each physical robot has meaningfully different scene camera and robot configurations, both are calibrated weekly during operations. Some controller parameters (such as Cartesian velocity bounds) differ slightly between robots, but these do not impact normal operations.

\subsection{Simulated Robot}
\begin{figure}
\centering
\begin{subfigure}[t]{.45\linewidth}
\includegraphics[width=\linewidth]{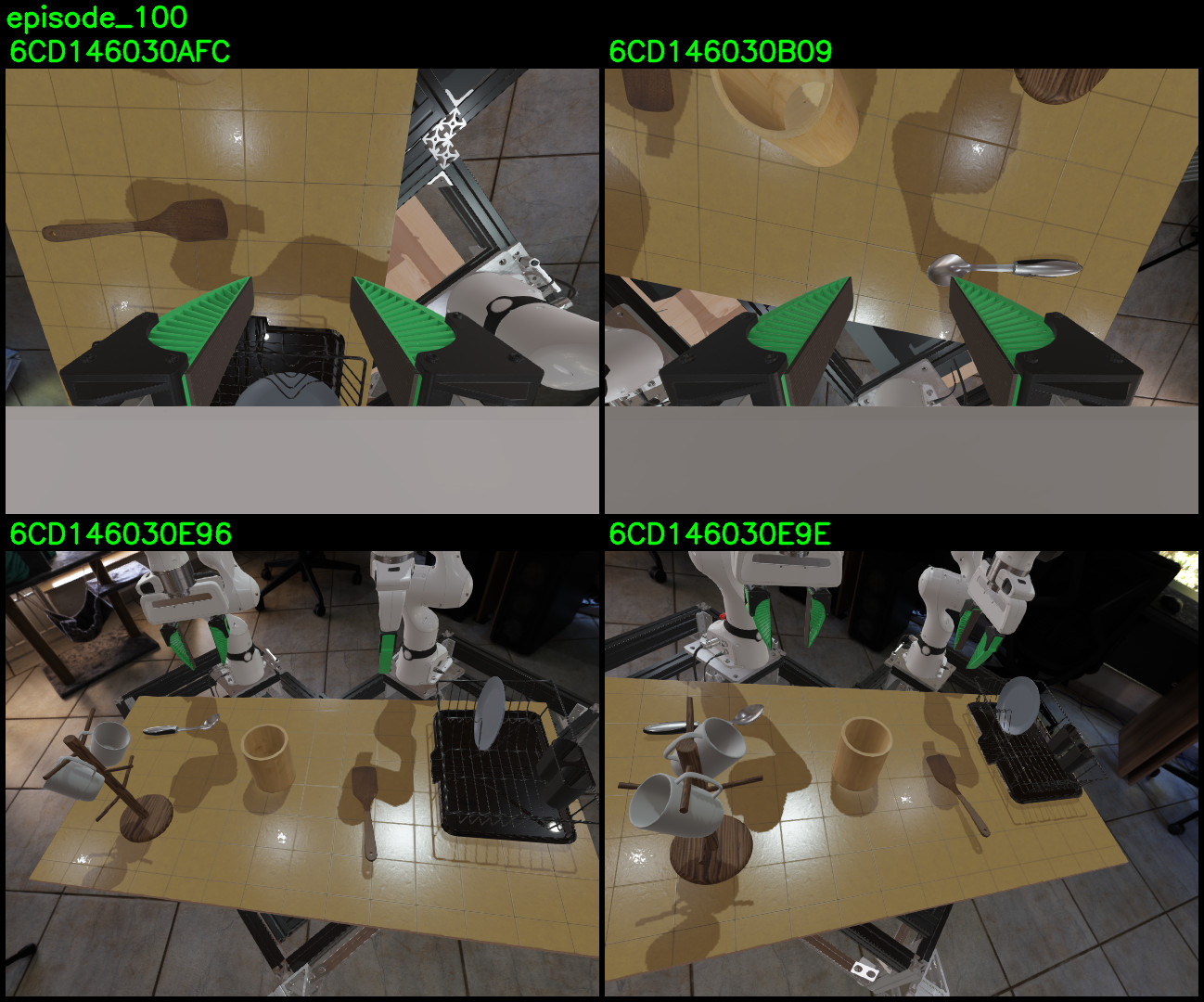}
\subcaption{Rendered camera images on simulated \textbf{riverway}.}
\end{subfigure}
\begin{subfigure}[t]{.45\linewidth}
\includegraphics[width=\linewidth]{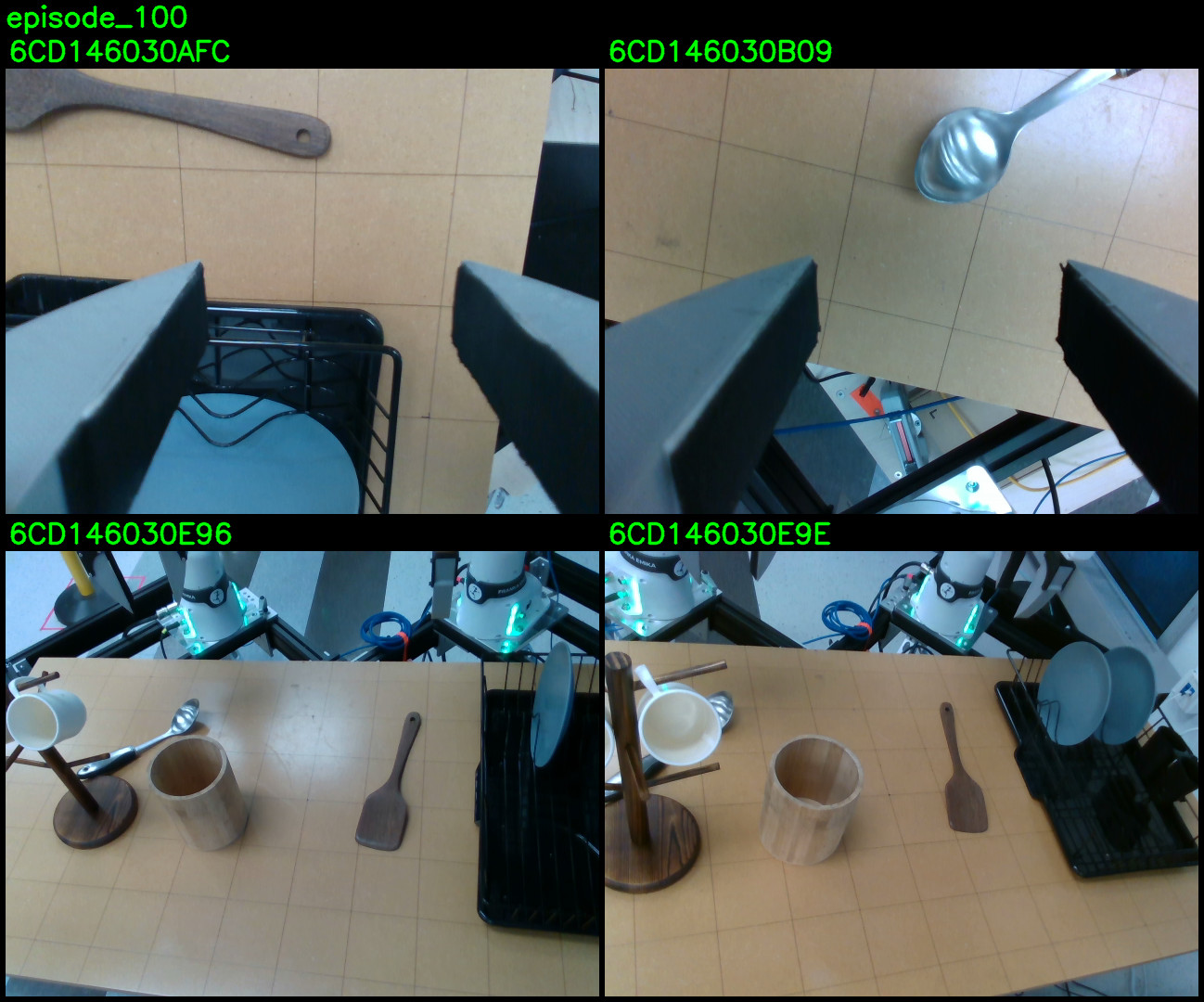}
\subcaption{Camera images on hardware \textbf{riverway}.}
\end{subfigure}
\begin{subfigure}[t]{.45\linewidth}
\includegraphics[width=\linewidth]{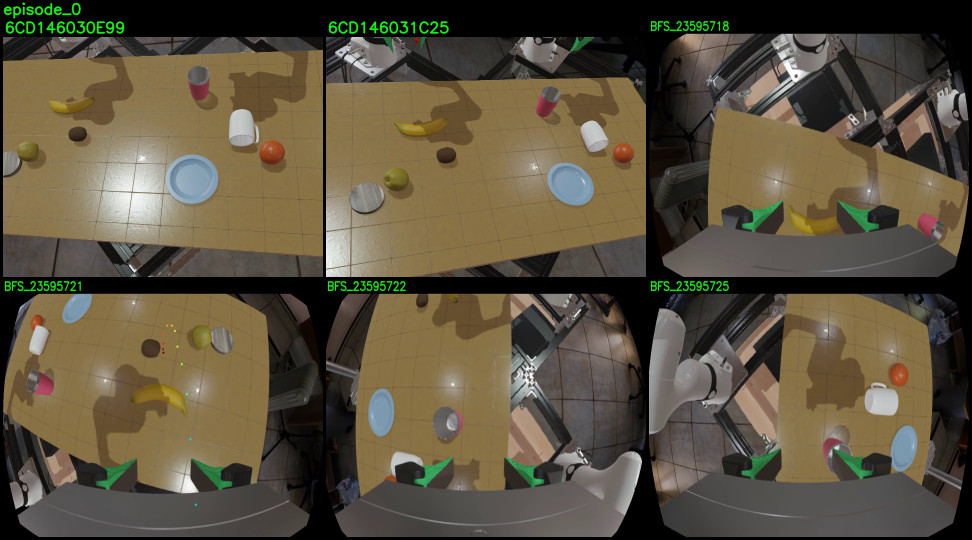}
\subcaption{Rendered camera images on simulation \textbf{cabot} station (modeled after \textbf{salem}).}
\end{subfigure}
\begin{subfigure}[t]{.45\linewidth}
\includegraphics[width=\linewidth]{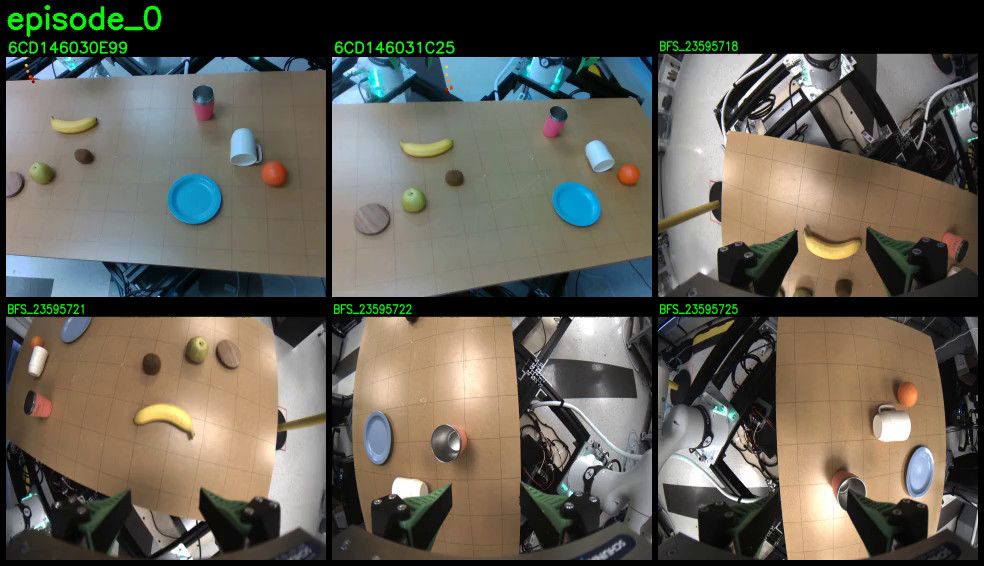}
\subcaption{Camera images on hardware \textbf{salem} station.}
\end{subfigure}
\caption{\small \textbf{Camera images from simulated and hardware platforms.} \textbf{riverway} is the old platform with 2 wrist cameras and the Franka Hand. \textbf{salem} and its counterpart in simulation, \textbf{cabot}, belongs to the new platform with 4 wrist cameras and a Schunk gripper. }
\label{fig:images_from_sim_and_real}
\end{figure}

We model the simulated robot station based on hardware stations, as can be seen in Fig~\ref{fig:images_from_sim_and_real}. In general, we did not try to achieve exact sim and real matching. We did a best effort to match the hardware platforms behaviorally (e.g., controllers). When simulating the wide angle lens on the new platform, we first render using a pin-hole model with the calibrated and rectified intrinsics, then apply fisheye distortion. On simulated \textbf{riverway}, we intentionally made both the simulated scene and wrist cameras wider angle than on hardware for better observability.

\adaptivetable{

\centering
\begin{tabular}{|l|llll|}
\hline
Station Name             & HW / Sim  & Platform type & Data collection  & Evaluation \\
\hline
\textbf{cabot}           & sim       & new         & Yes               & Yes       \\
\textbf{riverway} (sim)  & sim       & old         & Yes               & Yes       \\
\hline
\textbf{wood\_island}    & HW        & new         & Yes               & Yes       \\
\textbf{hersey}          & HW        & new         & Yes               & Yes       \\
\textbf{maverick}        & HW        & new         & Yes               & Yes       \\       
\textbf{ruggles}         & HW        & new         & Yes               & Yes       \\
\textbf{salem}           & HW        & new         & Yes               & Yes       \\
\textbf{davis}           & HW        & new         & Yes               & Yes       \\
\textbf{milton}          & HW        & new         & Yes               & Yes       \\
\textbf{wollaston}       & HW        & old         & Yes               & No        \\
\textbf{riverway} (HW)   & HW        & old         & Yes               & No        \\
\hline
\end{tabular}
\caption{\small Summary of hardware and simulation robots. Data collection and Evaluation columns indicate usage for this paper. Hardware \textbf{riverway} and \textbf{wollaston} were decommissioned and not used for evaluations.}
\label{tab:table_of_stations}
}

%% file: sections/appendix_simulation_eval_details.tex
\section{Simulation evaluation details}
\label{sec:sim_evaluation_details}
We present additional detailed information regarding the simulation evaluation as well as the predicates used to measure task completion in this section. Both platforms mentioned in Sec. ~\ref{subsec:supplemental_station_details} are used in the simulated scenarios. In particular, scenario \textit{D} and \textit{S} are modeled after the old hardware station \textbf{riverway}. These two scenarios are still actively used for simulation evaluation. Scenario \textit{B} and \textit{K} are modeled after a new station \textbf{salem}. 

\subsection{Data used for training or finetuning during simulation evaluation}
The amount of training data for each evaluation task is summarized in Table~\ref{tab:suppl_sim_eval_data_amount}. Note that we only used 16 ``seen" and 5 ``unseen" tasks for simulation evaluation out of a total of 44 tasks, as described in Section~\ref{subsec:data}.

\adaptivetable{
\centering
\resizebox{0.7\textwidth}{!}{%
\begin{tabular}{lll|rrrrrr}
\toprule
\textbf{Task name} & \textbf{Seen in pretraining} & \textbf{Scenario} & \textbf{riverway} & \textbf{cabot} & \textbf{Total} \\
\midrule
PutSpatulaOnPlateFromDryingRack & Seen & \textit{DryingRack (D)} & 196 & 0 & 196 \\
PutSpatulaOnPlateFromTable & Seen & \textit{DryingRack (D)} & 196 & 0 & 196 \\
StackPlatesOnTableFromDryingRack & Seen & \textit{DryingRack (D)} & 196 & 0 & 196 \\
PutSpatulaInUtensilCrock & Seen & \textit{DryingRack (D)} & 196 & 0 & 196 \\
PutSpatulaOnPlateFromUtensilCrock & Unseen & \textit{DryingRack (D)} & 196 & 0 & 196 \\
PlaceAppleFromBowlIntoBin & Seen & \textit{Shelf (S)} & 196 & 0 & 196 \\
PlaceFruitFromBowlIntoBin & Seen & \textit{Shelf (S)} & 196 & 0 & 196 \\
PutBellPepperInBin & Seen & \textit{Shelf (S)} & 196 & 0 & 196 \\
StoreCerealBoxUnderShelf & Seen & \textit{Shelf (S)} & 196 & 0 & 196 \\
PlaceAvocadoFromBowlIntoBin & Unseen & \textit{Shelf (S)} & 196 & 0 & 196 \\
PlaceCupByCoaster & Seen & \textit{Breakfast (B)} & 0 & 196 & 196 \\
PushCoasterToCenterOfTable & Seen & \textit{Breakfast (B)} & 0 & 196 & 196 \\
PutBananaOnSaucer & Seen & \textit{Breakfast (B)} & 0 & 49 & 49 \\
PutMugOnSaucer & Seen & \textit{Breakfast (B)} & 0 & 196 & 196 \\
TurnCupUpsideDown & Seen & \textit{Breakfast (B)} & 0 & 490 & 490 \\
PutMugInCenterOfTable & Unseen & \textit{Breakfast (B)} & 0 & 294 & 294 \\
TurnMugRightsideUp & Seen & \textit{Breakfast (B)} & 0 & 490 & 490 \\
PushCoasterToMug & Seen & \textit{Breakfast (B)} & 0 & 196 & 196 \\
PutKiwiInCenterOfTable & Seen & \textit{Breakfast (B)} & 0 & 49 & 49 \\
TurnLargeContainerUpsideDown & Unseen & \textit{Kitchen (K)} & 0 & 392 & 392 \\
PutContainersOnPlate & Unseen & \textit{Kitchen (K)} & 0 & 392 & 392 \\
DumpVegetablesFromSmallToLargeContainer & Unseen & \textit{Kitchen (K)} & 0 & 392 & 392 \\
PutFruitInLargeContainerAndCoverWithPlate & Unseen & \textit{Kitchen (K)} & 0 & 392 & 392 \\
SeparateFruitsVegetablesIntoContainers & Unseen & \textit{Kitchen (K)} & 0 & 392 & 392 \\
\bottomrule
\end{tabular}}
\caption{\small Number of demonstrations for the simulation evaluation tasks per station.}
\label{tab:suppl_sim_eval_data_amount}
}

\subsection{Simulation initial conditions}
\label{subsec:sim_sample_ic}
Our simulation allows instantiating scenes by drawing from a predefined distribution. For manipulands, for example, we can randomize 
the number of manipulands, their shape 
texture and color, and pose relative to any arbitrary frames. The initial condition for any task is implicitly defined as a distribution. 
At the start of each simulation, a sample initial condition is drawn deterministically based on the simulation seed; when comparing different policies during evaluation, we use the same seed.

\begin{figure}[t!]
    \centering
    \begin{subfigure}[t]{0.495\textwidth}
        \centering
        \includegraphics[width=1.0\linewidth]{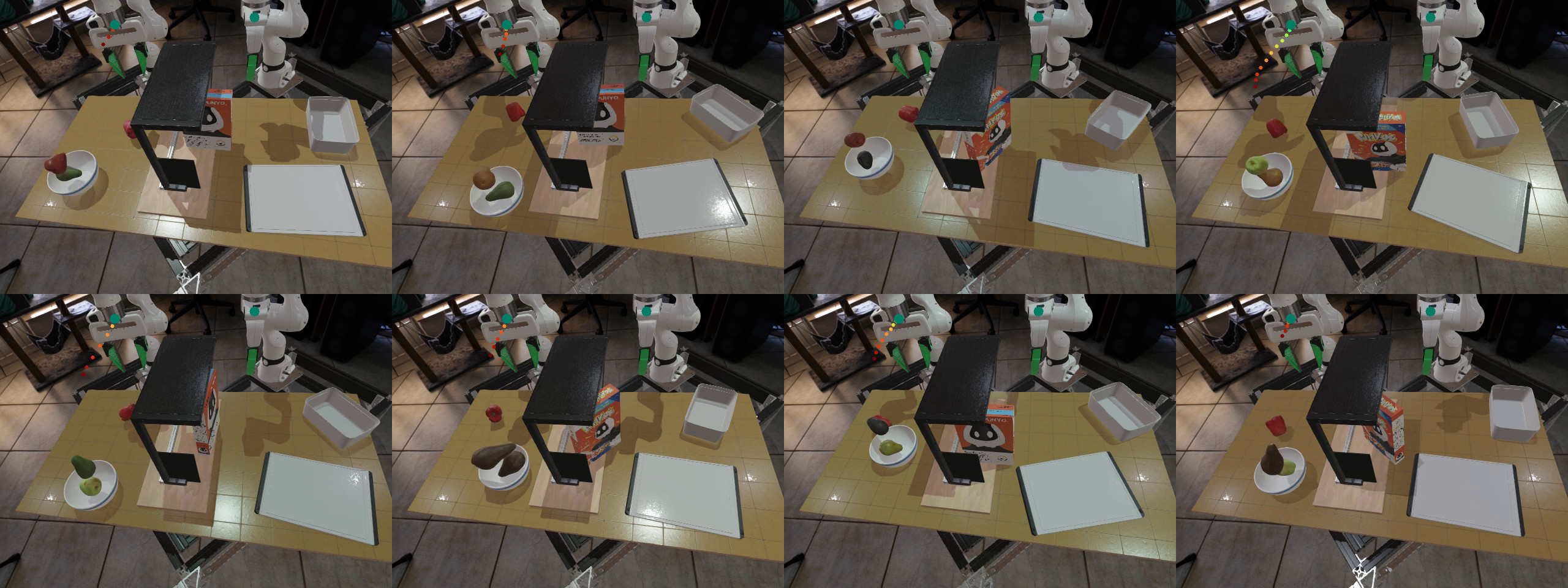}
        \caption{Nominal (Sim)}
    \end{subfigure}%
    \hfill
    \begin{subfigure}[t]{0.495\textwidth}
        \centering
        \includegraphics[width=1.0\linewidth]{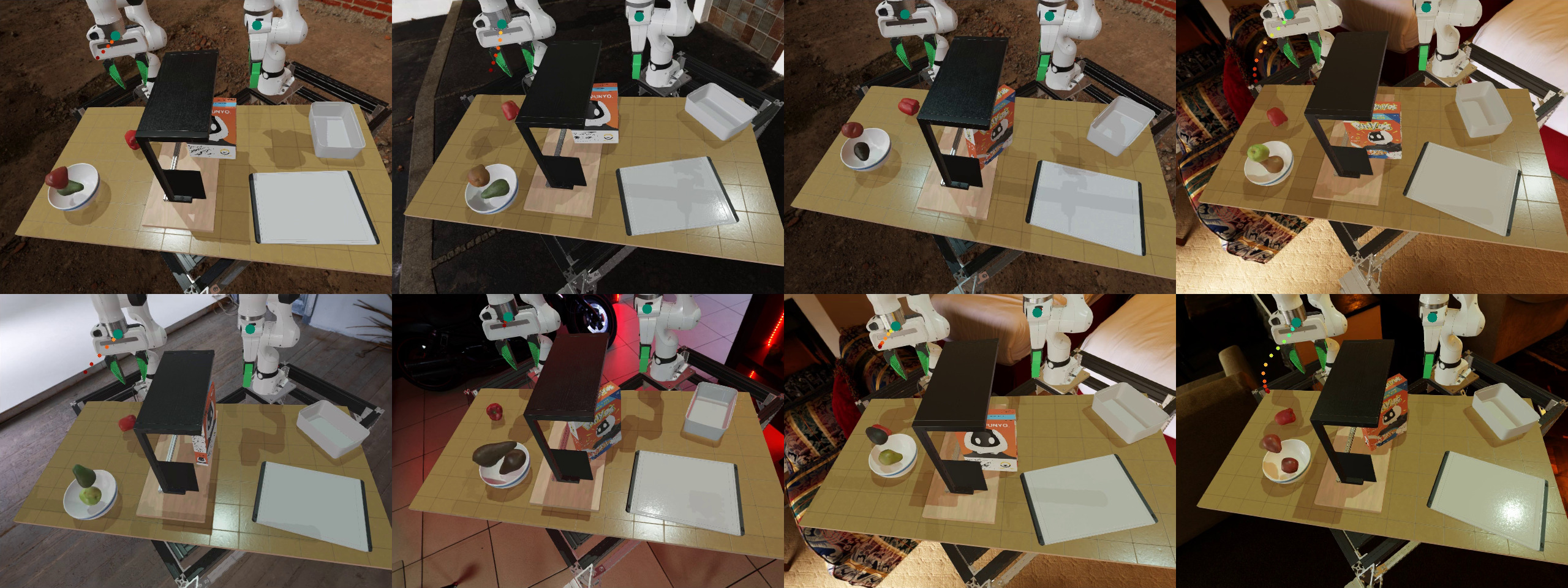}
        \caption{Distribution Shift (Sim)}
    \end{subfigure}
    
    \caption{\small\textbf{Sample initial conditions} of the \textit{PlaceFruitFromBowlIntoBin} task from scenario \textit{S} on \textbf{riverway}.}
    \label{fig:PlaceFruitFromBowlIntoBin_ICs_nominal_vs_dist_shift}
\end{figure}

\begin{figure}[t!]
    \centering
    \begin{subfigure}[t]{0.495\textwidth}
        \centering
        \includegraphics[width=1.0\linewidth]{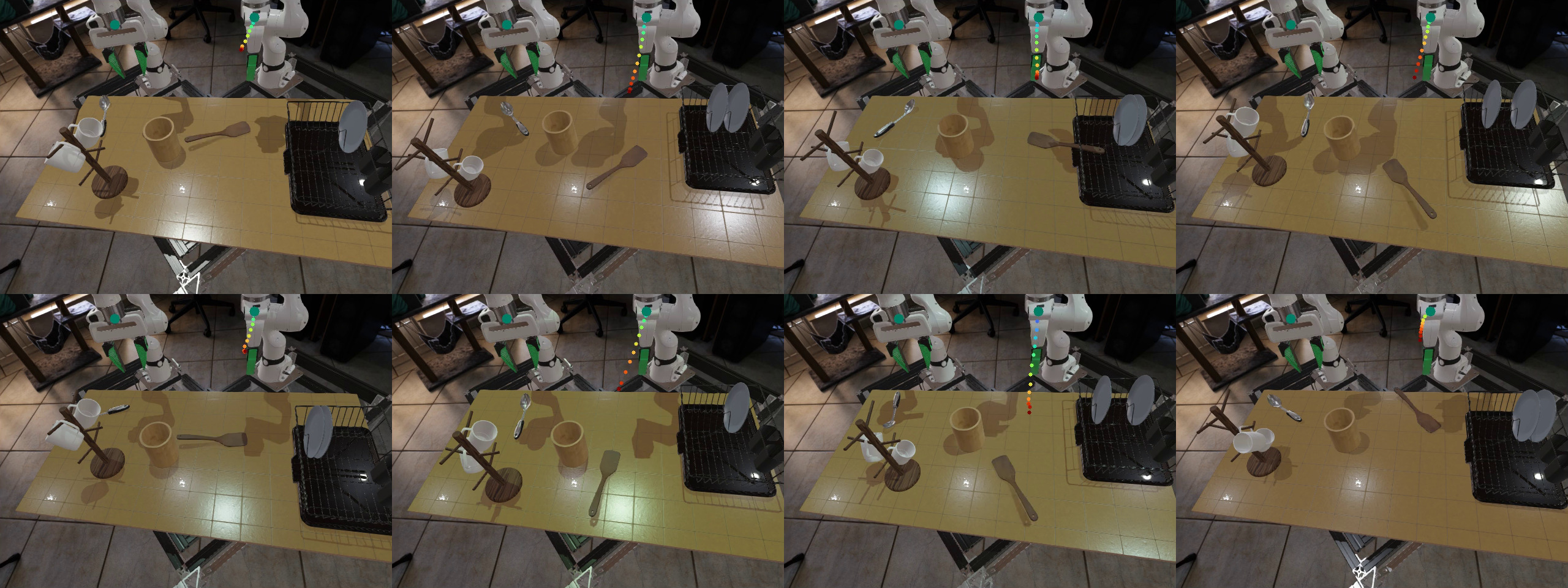}
        \caption{Nominal (Sim)}
    \end{subfigure}%
    \hfill
    \begin{subfigure}[t]{0.495\textwidth}
        \centering
        \includegraphics[width=1.0\linewidth]{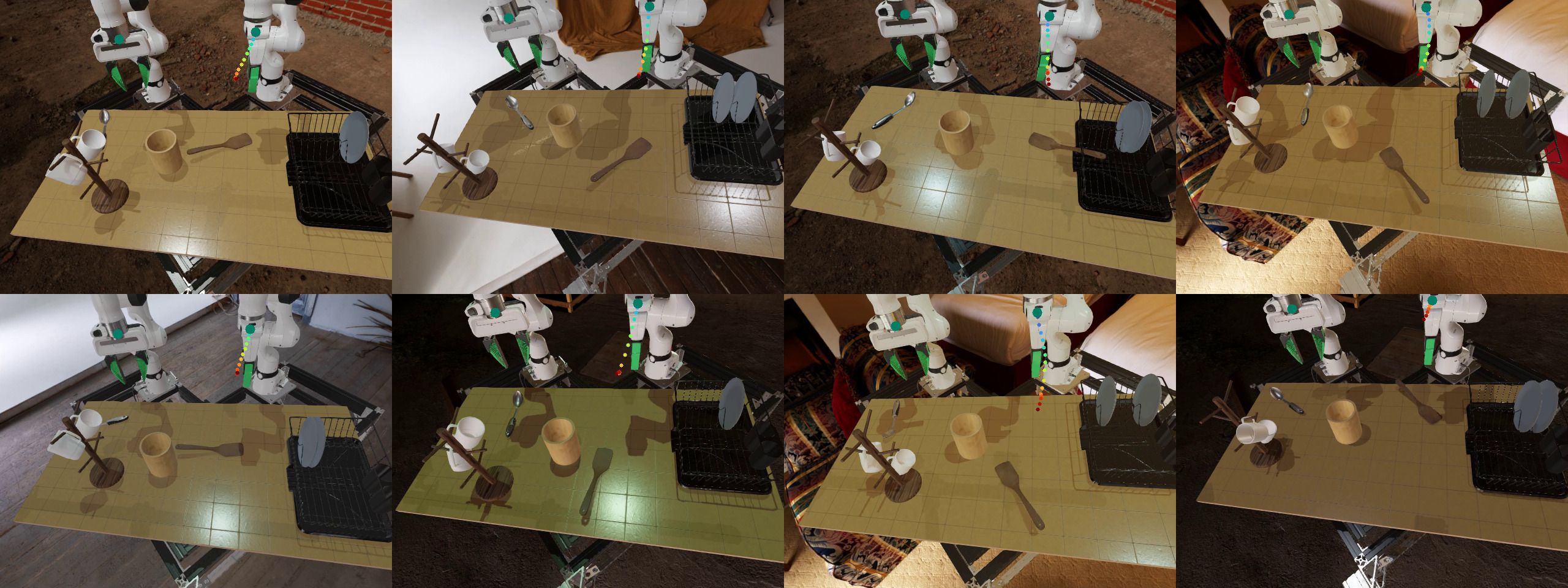}
        \caption{Distribution Shift (Sim)}
    \end{subfigure}
    
    \caption{\small\textbf{Sample initial conditions} of the \textit{PutSpatulaInUtensilCrock} task from scenario \textit{D} on \textbf{riverway}.}
    \label{fig:SpatulaInCrock_ICs_nominal_vs_dist_shift}
\end{figure}

\begin{figure}[t!]
    \centering
    \begin{subfigure}[t]{0.495\textwidth}
        \centering
        \includegraphics[width=1.0\linewidth]{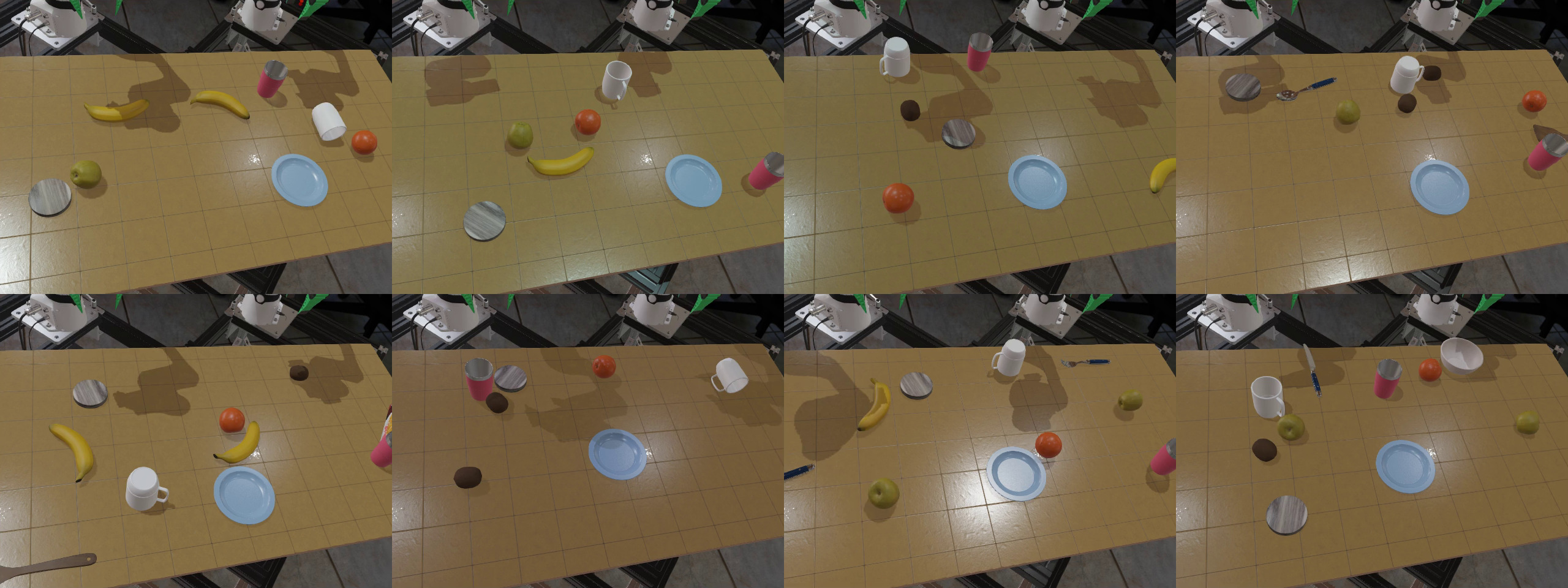}
        \caption{Nominal (Sim)}
    \end{subfigure}%
    \hfill
    \begin{subfigure}[t]{0.495\textwidth}
        \centering
        \includegraphics[width=1.0\linewidth]{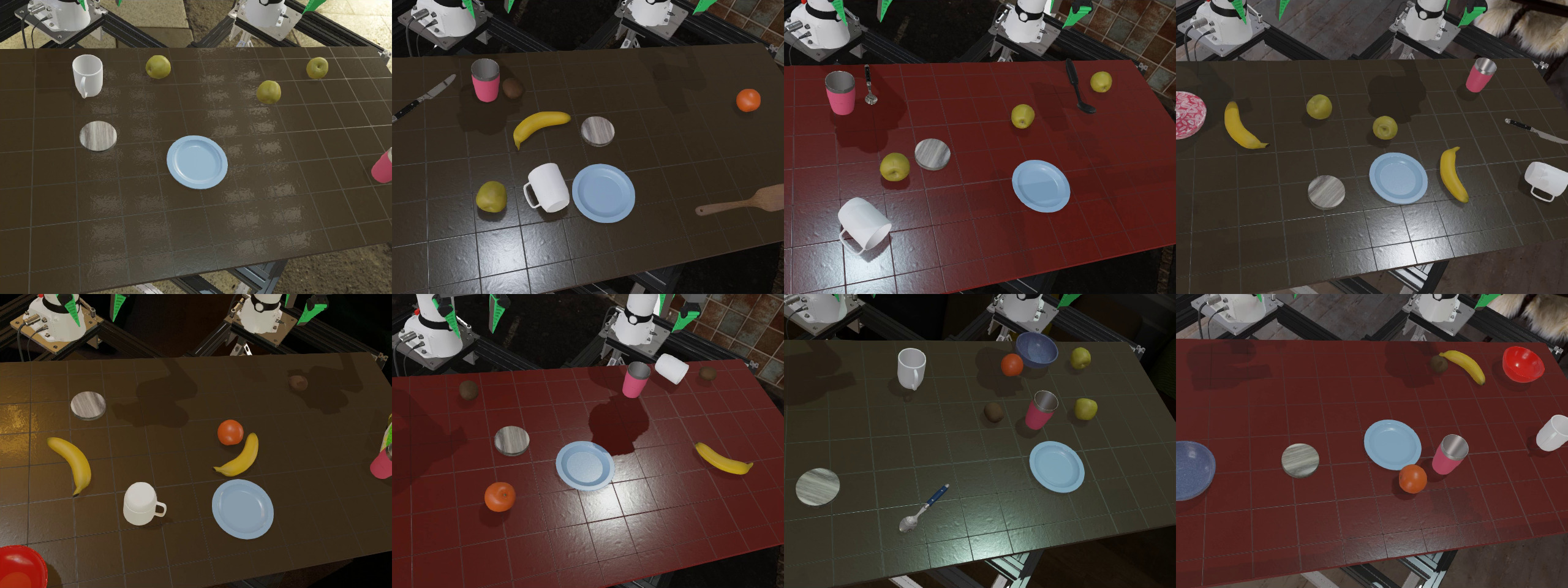}
        \caption{Distribution Shift (Sim)}
    \end{subfigure}
    
    \caption{\small\textbf{Sample initial conditions} of the \textit{PushCoasterToCenterOfTable} task from scenario \textit{B} on \textbf{cabot}.}
    \label{fig:PushCoasterToCenterOfTable_ICs_nominal_vs_dist_shift}
\end{figure}

\begin{figure}[t!]
    \centering
    \begin{subfigure}[t]{0.495\textwidth}
        \centering
        \includegraphics[width=1.0\linewidth]{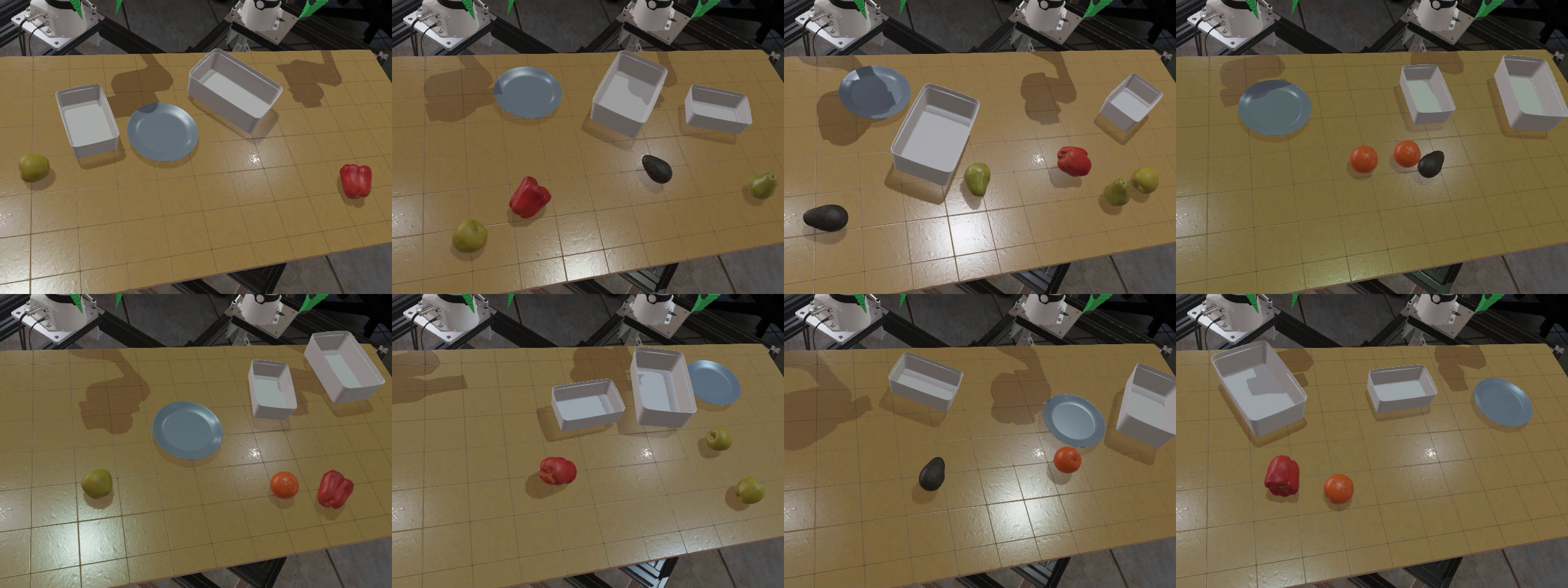}
        \caption{Nominal (Sim)}
    \end{subfigure}%
    \hfill
    \begin{subfigure}[t]{0.495\textwidth}
        \centering
        \includegraphics[width=1.0\linewidth]{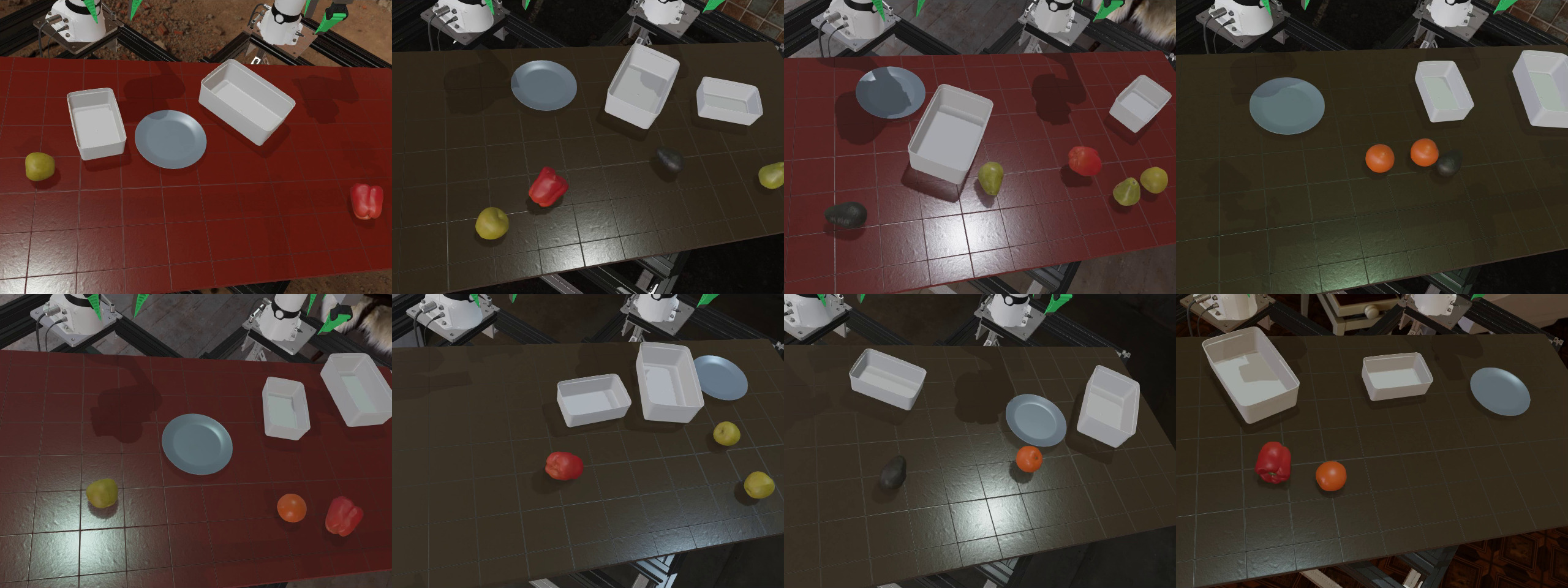}
        \caption{Distribution Shift (Sim)}
    \end{subfigure}
    
    \caption{\small\textbf{Sample initial conditions} of any task in scenario \textit{K} on \textbf{cabot}. All five tasks (\textit{SeparateFruitsVegetablesIntoContainers}, \textit{PutContainersOnPlate}, \textit{DumpVegetablesFromSmallToLargeContainer}, \textit{TurnLargeContainerUpsideDown} and \textit{PutFruitInLargeContainerAndCoverWithPlate}) in this scenario have exactly the same distribution for initial conditions. }
    \label{fig:scenario4_ICs_nominal_vs_dist_shift}
\end{figure}

We designed the scenarios to be visually ambiguous, meaning policies cannot infer the task solely based on visual appearances. 
We present sample initial conditions for nominal and distribution shift for a representative task in each scenario 
in Figures ~\ref{fig:PlaceFruitFromBowlIntoBin_ICs_nominal_vs_dist_shift}, ~\ref{fig:SpatulaInCrock_ICs_nominal_vs_dist_shift} ~\ref{fig:PushCoasterToCenterOfTable_ICs_nominal_vs_dist_shift} and ~\ref{fig:scenario4_ICs_nominal_vs_dist_shift}. 

The types of distribution shift in simulation used for each scenario are summarized in Table ~\ref{tab:sim_distribution_types}. 

\adaptivetable{
\centering
\begin{tabular}{|l|cccc|}
\hline
\textbf{Distribution Shift Type}  & \textbf{Scenario \textit{S}} & \textbf{Scenario \textit{D}} & \textbf{Scenario \textit{B}} & \textbf{Scenario \textit{K}} \\
\hline
\texttt{distractor\_textures}     &              & \checkmark   & \checkmark   &              \\
\texttt{environment\_map}         & \checkmark   & \checkmark   & \checkmark   & \checkmark   \\
\texttt{lighting}                 & \checkmark   & \checkmark   & \checkmark   & \checkmark   \\
\texttt{table\_top\_texture}      &              &              & \checkmark   & \checkmark   \\
\texttt{scene\_extrinsics}        & \checkmark   & \checkmark   & \checkmark   & \checkmark   \\
\texttt{scene\_intrinsics}        & \checkmark   & \checkmark   & \checkmark   & \checkmark   \\
\texttt{wrist\_intrinsics}        & \checkmark   & \checkmark   & \checkmark   & \checkmark   \\
\hline
\end{tabular}

\caption{\small\textbf{Types of distribution shift in the simulation results reported in Figures~\ref{fig:seen_tasks_sim_and_real} and ~\ref{fig:unseen_tasks_sim_and_real_ds}}. 
\texttt{distractor\_texture} randomizes textures for distractors; \texttt{environment\_map} samples one of 20 different environment maps; \texttt{lighting} adds one more randomized directional light source; 
\texttt{manipuland\_texture} applies randomized texture on manipulands; \texttt{table\_top\_texture} applies randomize table-top texture; \texttt{scene(wrist)\_extrinsics(intrinsics)} applies randomized deltas on top of nominal camera parameters. }

\label{tab:sim_distribution_types}
}

\subsection{Predicates}
\label{subsec:supp_predicates_simulation}
Success criteria as well as task-completion percentage in simulation are based on task-specific predicates. 
Each predicate is a function from instantaneous simulation state to a Boolean value, and is logged densely in the raw rollout log. These logged predicate trajectories can be used to perform more granular analysis or answer questions that require stateful information (e.g., did the robot achieve goal A first and then B) offline. Note that the predicates for each task are implemented as part of the task authoring, where the task designer has a nominal strategy in mind. The predicates are thus heavily influenced by the nominal strategy.We analyze task completion only for the five tasks that belong to the \textit{Kitchen (K)} scenario, as discussed in Section~\ref{subsec:LBMs_unseen_tasks}. The task-specific predicates for these tasks are shown below:

\vspace{1em}
{
\noindent
\scriptsize
\begin{tabular}{p{0.95\linewidth}}
\textbf{TurnLargeContainerUpsideDown} \\
\hline
Robot's right hand fingers touched the larger container  \\
Robot's left hand fingers touched the larger container  \\
Robot turned the large container over 45 degrees  \\
Robot turned the large container turn 90 degrees  \\
Robot turned the large container turn 135 degrees  \\
Robot flipped the large container upside down \\ 
\end{tabular}
}
\vspace{1em}

{
\noindent
\scriptsize
\begin{tabular}{p{0.95\linewidth}}
\textbf{PutContainersOnPlate} \\
\hline
Robot lift the larger container  \\
The large container on top of the plate  \\
Robot lift the smaller container  \\
The smaller container inside the larger container  \\
\end{tabular}
}
\vspace{1em}

{
\noindent
\scriptsize
\begin{tabular}{p{0.95\linewidth}}
\textbf{PutFruitInLargeContainerAndCoverWithPlate} \\
\hline
Robot picked up any fruit  \\
Any fruit in the large container  \\
Robot picked up the plate  \\
All fruit in the large container and the plate on top of the large container \\ 
\end{tabular}
}
\vspace{1em}

{
\noindent
\scriptsize
\begin{tabular}{p{0.95\linewidth}}
\textbf{SeparateFruitsVegetablesIntoContainers} \\
\hline
Robot picked up any fruit  \\
Any fruit in the small container  \\
Robot picked up any vegetable  \\
Any vegetable in the large container  \\
All fruit in the smaller container and all vegetable in the large container \\
\end{tabular}
}
\vspace{1em}

{
\noindent
\scriptsize
\begin{tabular}{p{0.95\linewidth}}
\textbf{DumpVegetablesFromSmallToLargeContainer} \\
\hline
Robot picked up any vegetable  \\
Robot put any vegetable in small container  \\
Robot lifted the small container  \\
Robot turned the small container over 45 degrees  \\
Robot turned the small container over 90 degrees  \\
Robot moved the small container over large container  \\
Any vegetable inside the large container  \\
All vegetable inside the large container while robot held the smaller container \\ 
\end{tabular}
}
\vspace{1em}

\subsection{Missing simulation rollouts}
\label{subsec:missing_sim_rollouts}
We run every simulation task 200 times per policy and per condition; since the evaluation is run in a distributed manner on the cloud, rollout data can be missing due to various reasons (e.g., sync rollout data to the cloud storage failed). Table~\ref{tab:num_sim_missing_rollouts} below lists all the tasks and conditions for which we use less than 200 rollouts in the results (Section~\ref{sec:results}); all other tasks/policy/conditions have 200 rollouts. In terms of overall impact: out of 238 task/policy/condition combinations, 39 have missing rollouts, for a total of 104 rollouts representing 0.2\% of the overall simulation evaluation data. 

\begin{table}[h!]
\tiny
\centering
\begin{tabular}{|l|c|c|c|}
\hline
\textbf{Task} & \textbf{Policy} & \textbf{Condition} & \textbf{Rollouts}\\
\hline
\multicolumn{4}{|c|}{Nominal conditions} \\
\hline
PutBellPepperInBin & Single task & - & 199 \\
PutBellPepperInBin & LBM & Finetuned & 185 \\
PutSpatulaOnPlateFromRack & LBM & Pretrained & 199 \\
StackPlatesOnTableFromRack & LBM & Pretrained & 198 \\
StoreCerealBoxUnderShelf & Single task & - & 199 \\
PushCoasterToMug & Single task & - & 199 \\
PushCoasterToMug & LBM & Pretrained & 199 \\
SeparateFruitsVegIntoContainers & Single task  & Trained with 15\% data & 199 \\
PutFruitInLargeContainerAndCoverWithPlate & Single task  & Trained with 50\% data & 199 \\

\hline
\multicolumn{4}{|c|}{Distribution shift} \\
\hline
PutAppleFromBowlInBin & LBM & Pretrained & 198 \\
PutFruitFromBowlInBin & Single task & - & 199 \\
PutFruitFromBowlInBin & LBM & Pretrained & 198 \\
PutFruitFromBowlInBin & LBM & Finetuned & 198 \\
PutBellPepperInBin & Single task & - & 197 \\
PutBellPepperInBin & LBM & Pretrained & 193 \\
PutSpatulaOnPlateFromRack & Single task & - & 195 \\
PutSpatulaOnPlateFromRack & LBM & Pretrained & 197 \\
PutSpatulaOnPlateFromRack & LBM & Finetuned & 198 \\
StoreCerealBoxUnderShelf & Single task & - & 198 \\
PushCoasterToCenterOfTable & LBM & Pretrained & 198 \\
PushCoasterToMug & LBM & Pretrained & 198 \\
PushCoasterToMug & LBM & Finetuned & 199 \\
PutBananaOnSaucer & Single task & - & 199 \\
PutKiwiInCenterOfTable & Single task & - & 191 \\
TurnCupUpsideDown & LBM & Pretrained & 199 \\
PutContainersOnPlate & Single task & - & 199 \\
PutContainersOnPlate & Single task & Trained with 15\% data & 199 \\
PutContainersOnPlate & LBM & Pretrained & 199 \\

PutFruitInLargeContainerAndCoverWithPlate & Single task & - & 198 \\
PutFruitInLargeContainerAndCoverWithPlate & Single task & Trained with 15\% data & 199 \\
SeparateFruitsVegIntoContainers & Single task & - & 199 \\
SeparateFruitsVegIntoContainers & LBM & Pretrained & 199 \\
TurnLargeContainerUpsideDown & LBM & Finetuned with 50\% data & 199 \\
BimanualPlaceAvocadoFromBowlIntoBin & Single task & - & 198 \\
\hline
\multicolumn{4}{|c|}{Fractional pretraining} \\
\hline
DumpVegetablesIntoContainer & LBM & TRI-Ramen-25 + FT@100 & 199 \\
DumpVegetablesIntoContainer & LBM & TRI-Ramen-50 + FT@15 & 199 \\
PutContainersOnPlate & LBM & TRI-Ramen + FT@100 & 198 \\
PutFruitInLargeContainerAndCoverWithPlate & LBM & TRI-Ramen-25 + FT@15 & 198 \\
SeparateFruitsVegIntoContainers & LBM & TRI-Ramen-50 + FT@50 & 199 \\
\hline

\hline
\end{tabular}
\caption{\small Simulation tasks presented in Section~\ref{sec:results} that have less than 200 rollouts listed by  policy and condition. LBM FT=Finetuned LBM, LBM=Pretrained LBM, ST=Single task baseline, DS=Distribution shift. 
}
\label{tab:num_sim_missing_rollouts}
\end{table}

%% file: sections/12_appendix_hardware_eval_details.tex
\section{Real-world evaluation details}
\label{sec:hardware_eval_details}
In this section, we provide more details about the real-world tasks themselves, as well as information regarding our real-world evaluation process. 

\subsection{Real-world evaluation process}

\textbf{Initializing a rollout:} For physically setting up the scene to match desired initial condition, we overlay current live camera feed on top of a snapshot of the desired initial condition as illustrated in Fig. ~\ref{fig:initial_conditions}. In particular, we use homographic projection to canonicalize images from different stations to facilitate using the same initial conditions among the hardware fleet. This method works well for objects close to the plane (e.g., table surface) used to compute the homographic projection matrix, but has severe artifacts for tall objects. This can confuse operators during manual alignment. Maintaining the same visibility across the robot fleet is also challenging, and can make certain scene configurations not observable across different stations. Finally, the effort it takes to set up a scene is highly correlated with the task and number of objects involved. We have found the following situations to be particularly challenging operationally: 1) many objects in the scene, 2) non-rigid / deformable objects, 3) messy materials (e.g., food, liquid, fine particles), and 4) irreversible actions (e.g., cutting objects in half, mixing ingredients). A few typical failure modes observed in our experiments when setting up the initial conditions are: 1) incorrect object placement; 2) similar but incorrect objects, typically distractor objects; 3) missed objects, typically in scenarios with many objects or due to poor visibility.

\textbf{Ending a rollout:} After starting a rollout, the operator needs to closely monitor progress and terminate according to certain criteria. Due to the effort involved in setting up each real-world rollout, our operators are instructed to only terminate each rollout if the robot 1) succeeded at the task, 2) exhibited dangerous behavior, or 3) made no progress for a while or was stuck in a repetitive loop. These criteria are more reliant on human judgment and forgiving than those used in simulation, which uses an unconditional fixed timeout and boolean checks for success criteria. It provides a robot ample opportunities to make mistakes and recover, and even allows the robot to perform some other tasks and then perform the commanded task. This in fact is part of the reason for the performance differences for the ``seen" tasks in simulation and real-world (see Section ~\ref{subsec:supp_qual_analysis_sim_and_real} for more details). 


The real-world evaluation process is highly repetitive and laborious, with ample room for human errors. However, our QA analysis indicates an overall low discrepancy rate when success flags and rubric questions were compared to a secondary set (see QA results in Section~\ref{subsec:rubric_qa}).

\subsection{Real-world evaluation tasks}
\label{subsec:supplemental_hardware_number_of_episodes}
\begin{figure}
\begin{subfigure}[b]{\linewidth}
\includegraphics[width=\linewidth]{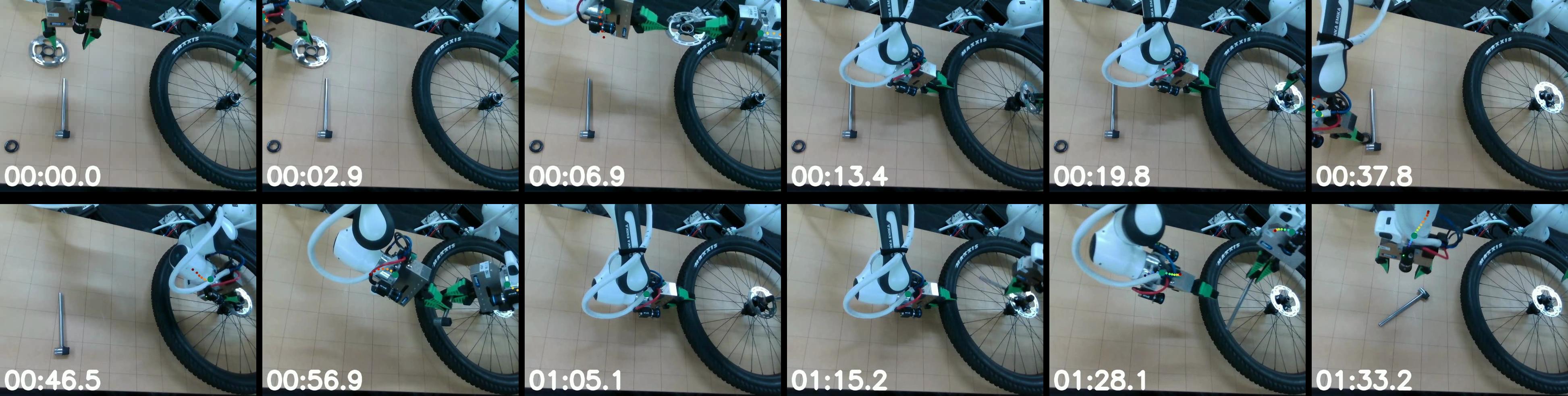}
\caption{\footnotesize Frames from a successful rollout for the \textit{BikeRotorInstall} task on \textbf{ruggles}. This task requires a high degree of precision, as well as bimanual coordination: while holding the wheel down with one arm, the robot has to place a rotor and a lockring onto the wheel as well as manipulate a tool with which to tighten the lockring for at least one revolution of the tool around the center axle of the wheel.}
\label{fig:long_horizon_BikeRotorInstall_filmstrips}
\end{subfigure}
\begin{subfigure}[b]{\linewidth}
\vspace{0.2in}
\includegraphics[width=\linewidth]{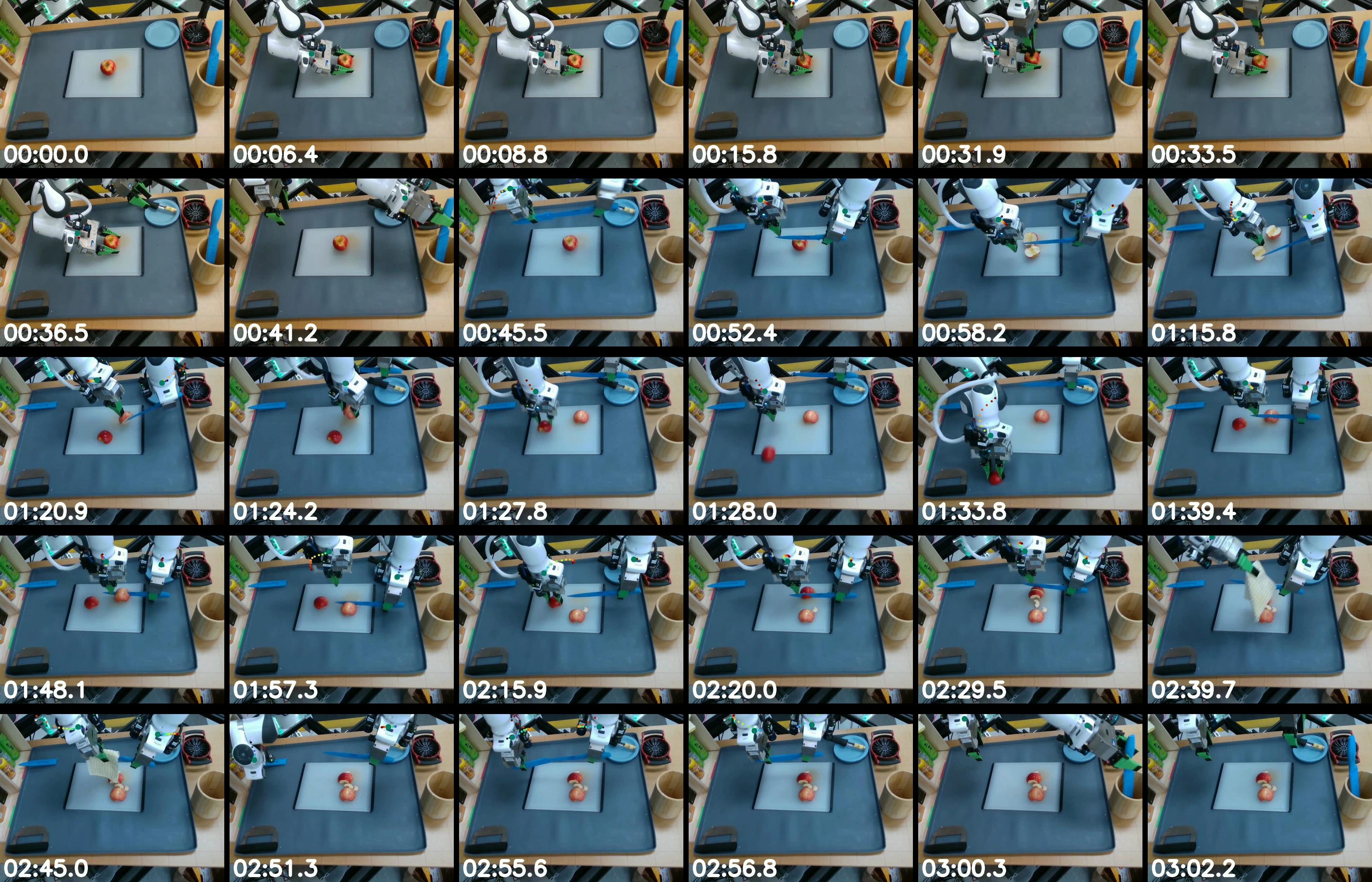}
\caption{\footnotesize Frames from a successful rollout for the \textit{CutAppleInSlices} task on \textbf{wood\_island}. This is the longest horizon task in our benchmark, and it can take up to 3 minutes to complete. The robot has to first use a tool to remove the core of the apple, after which, using a knife, it has to first slice the apple in half, position each half in the workspace and slice at least one half in at least 3 slices. Finally, the robot has to wipe the knife with a cloth, put it in the sheath, and place is back into the utensil crock.}
\label{fig:long_horizon_CutAppleIntoSlices_filmstrips}
\end{subfigure}
\begin{subfigure}[b]{\linewidth}
\vspace{0.2in}
\includegraphics[width=\linewidth]{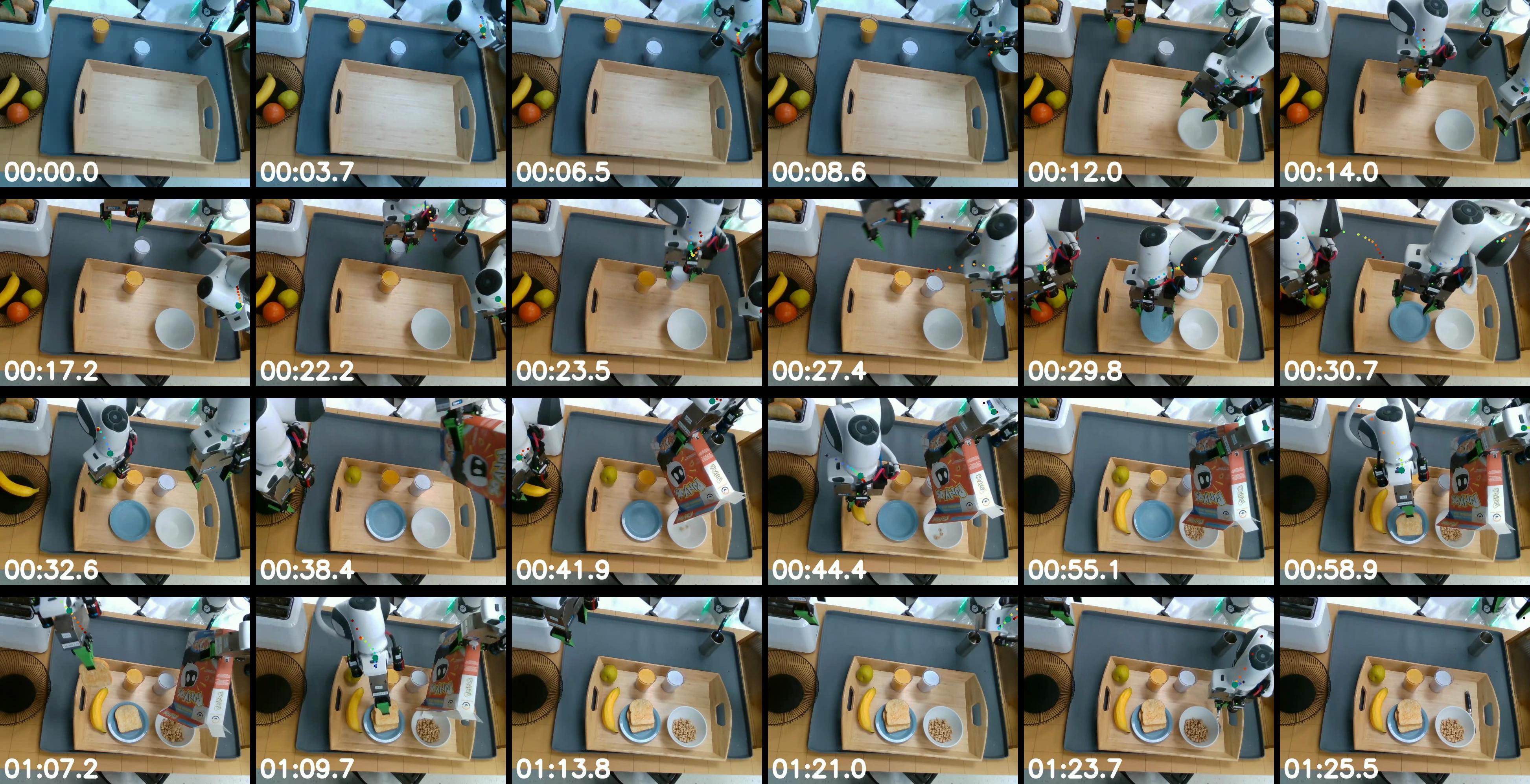}
\caption{\footnotesize Frames from a successful rollout for the \textit{SetBreakfastTable} task on \textbf{hersey}. This task involves sliding open a cabinet door and manipulating objects in the right order: a bowl, a milk cup, a juice cup, a plate, an apple, a banana, a toast, cereal and a spoon. Each object has to be placed on the tray in the right configuration (i.e., the cups should be upright, the cereal has to be poured into the bowl).}
\label{fig:long_horizon_SetBreakfastTable_filmstrips}
\end{subfigure}
\label{fig:long_horizon_filmstrips}
\caption{\small\textbf{Sample frames from successful long-horizon task rollout videos on hardware.} All of the above are from LBM policies finetuned on the respective tasks. 
}
\end{figure}

We designed five long-horizon complex tasks to specifically test the efficacy of finetuning LBMs on downstream complex tasks (see quantitative results in Section~\ref{subsec:LBMs_unseen_tasks}). Here, we provide example film strips for three of the most interesting tasks: \textit{BikeRotorInstall} in Fig.~\ref{fig:long_horizon_BikeRotorInstall_filmstrips}, \textit{CutAppleInSlices} in Fig.~\ref{fig:long_horizon_CutAppleIntoSlices_filmstrips} and \textit{SetBreakfastTable} in Fig.~\ref{fig:long_horizon_SetBreakfastTable_filmstrips}. 

We present the number of task-specific demonstrations used to pretrain or finetune our policies, together with the experimental condition (nominal, station distribution shift or object-centric distribution shift) in Table~\ref{tab:suppl_real_eval_data_amount}.

\adaptivetable{
\centering
\resizebox{\textwidth}{!}{%
\begin{tabular}{ll|rrr|rrrrrrrrrr}
\toprule
\textbf{Experiment} & \textbf{Task} & \textbf{Nominal} & \textbf{Diff. Station} & \textbf{DS} & \textbf{wood\_island} & \textbf{hersey} & \textbf{maverick} & \textbf{ruggles} & \textbf{salem} & \textbf{davis} & \textbf{milton} & \textbf{Total} \\
\midrule
Seen & PutKiwiInCenterOfTable & X & X & X & 0 & 0 & 0 & 0 & 49 & 0 & 0 & 49 \\
Seen & TurnMugRightsideUp & X & X & X & 0 & 121 & 0 & 0 & 122 & 122 & 125 & 490 \\
Seen & PushCoasterToMug & X & - & X & 98 & 0 & 0 & 0 & 98 & 0 & 0 & 196 \\
Unseen & ClearKitchenCounter & X & X & - & 0 & 0 & 0 & 0 & 0 & 305 & 0 & 305 \\
Unseen & BikeRotorInstall & X & - & - & 0 & 0 & 0 & 533 & 0 & 0 & 0 & 533 \\
Unseen & CutAppleInSlices & X & - & - & 496 & 0 & 0 & 0 & 0 & 0 & 0 & 496 \\
Unseen & CleanLitterBox & X & - & - & 0 & 0 & 0 & 0 & 0 & 0 & 260 & 260 \\
Unseen & SetUpBreakfastTable & X & - & - & 0 & 229 & 0 & 0 & 0 & 0 & 0 & 229 \\

\bottomrule
\end{tabular}}
\caption{\small Number of demonstrations for real-world evaluation tasks per robot station. The nominal, diff. station (station distribution shift), and DS (object distribution shift) columns summarize the types of experiments we performed for these tasks during the real-world evaluation. }
\label{tab:suppl_real_eval_data_amount}
}

\subsection{Sample real-world initial conditions}
\label{subsec:supp_hardware_ic_sampels}

\begin{figure}
\centering
\begin{subfigure}[t]{.22\linewidth}
\includegraphics[width=\linewidth]{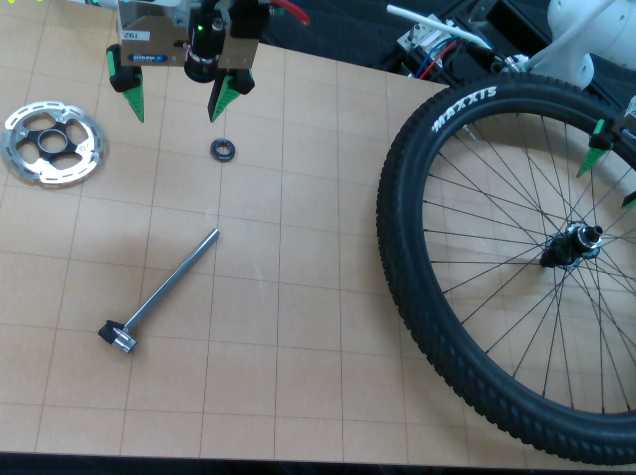}
\caption{BikeRotorInstall}
\end{subfigure}
\begin{subfigure}[t]{.22\linewidth}
\includegraphics[width=\linewidth]{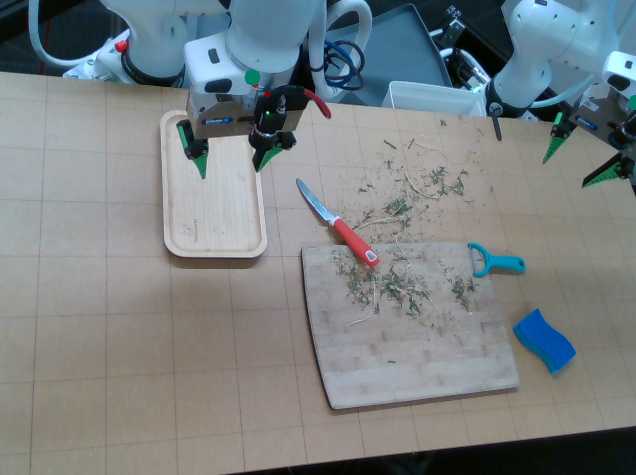}
\caption{ClearKitchenCounter}
\end{subfigure}
\begin{subfigure}[t]{.22\linewidth}
\includegraphics[width=\linewidth]{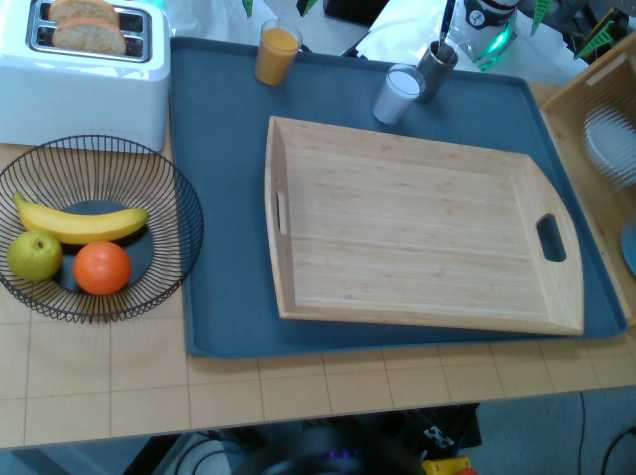}
\caption{SetUpBreakfastTable}
\end{subfigure}
\begin{subfigure}[t]{.22\linewidth}
\includegraphics[width=\linewidth]{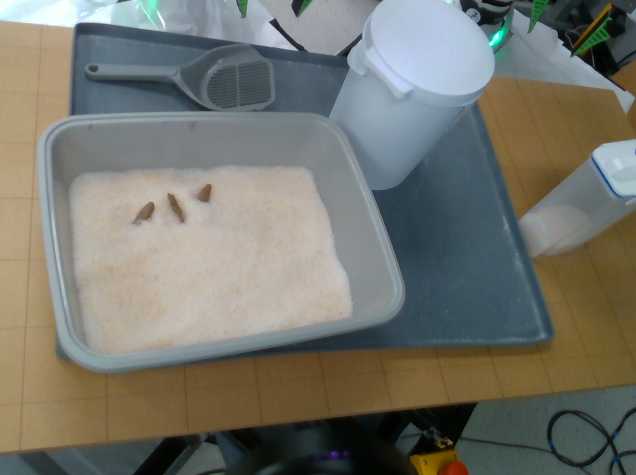}
\caption{CleanLitterBox}
\end{subfigure}
\begin{subfigure}[t]{.22\linewidth}
\includegraphics[width=\linewidth]{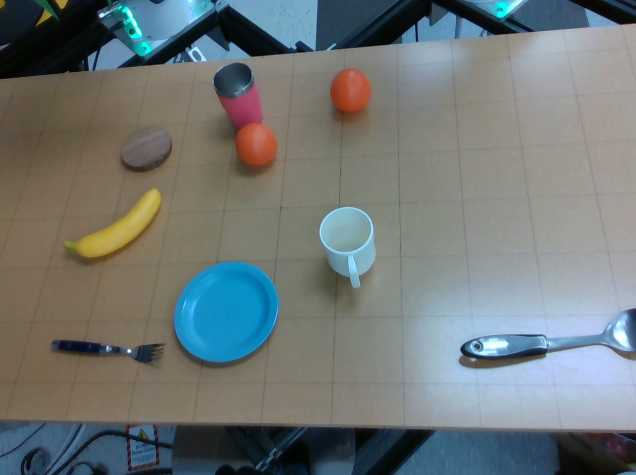}
\caption{PushCoasterToMug}
\end{subfigure}
\begin{subfigure}[t]{.22\linewidth}
\includegraphics[width=\linewidth]{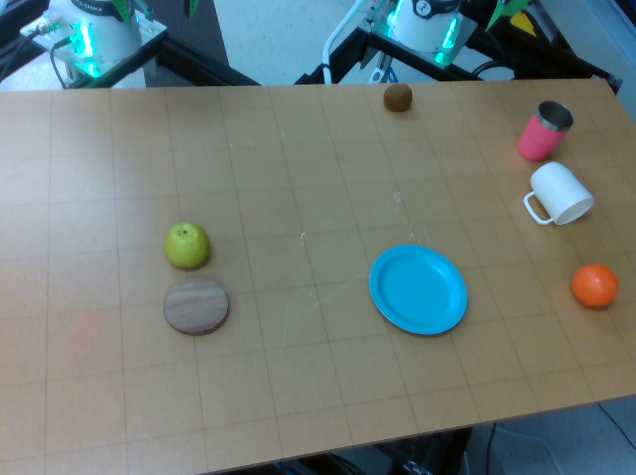}
\caption{PutKiwiInCenterOfTable}
\end{subfigure}
\begin{subfigure}[t]{.22\linewidth}
\includegraphics[width=\linewidth]{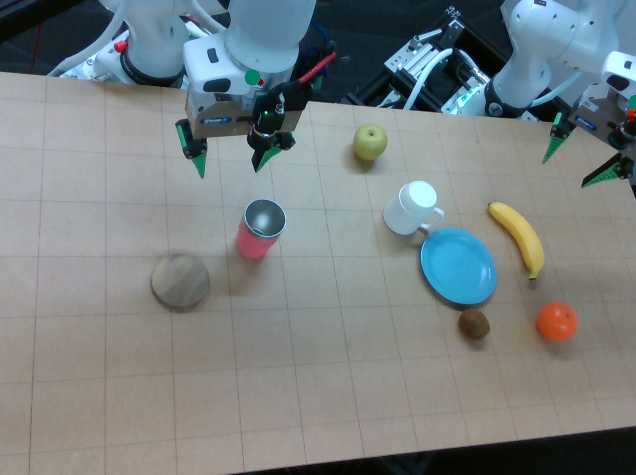}
\caption{TurnMugRightsideUp}
\end{subfigure}
\begin{subfigure}[t]{.22\linewidth}
\includegraphics[width=\linewidth]{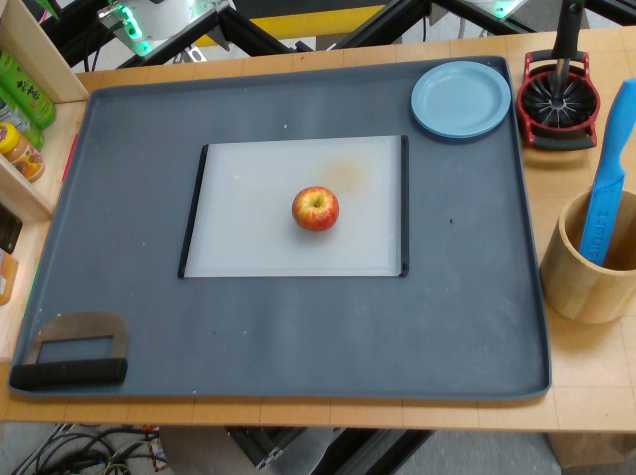}
\caption{CutAppleInSlices}
\end{subfigure}
\caption{\small\textbf{Initial conditions} for both ``seen" tasks (with corresponding quantitative results in Fig.~\ref{fig:seen_tasks_sim_and_real}) and ``unseen" tasks (with corresponding quantitative results in Fig.~\ref{fig:unseen_tasks_sim_and_real}) evaluated on hardware. The subfigure caption denotes the task for which the initial conditions are shown.}
\label{fig:hardware_task_ics}
\end{figure}

Fig.~\ref{fig:hardware_task_ics} shows one initial condition for each real-world task 
under nominal conditions. To illustrate distribution shift for real-world experiments, we provide sample initial conditions under nominal, station distribution shift, and object-centric distribution shift for \textit{PutKiwiInCenterOfTable} (Fig.~\ref{fig:hardware_ic_overlays_w_and_w_o_distribution_shift2}) and \textit{TurnMugRightsideUp} (Fig.~\ref{fig:hardware_ic_overlays_w_and_w_o_distribution_shift2_TurnMugRightsideUp}). 

\adaptivefigure{
\centering

\begin{subfigure}[t]{.655\linewidth}
\includegraphics[height=4.05cm]{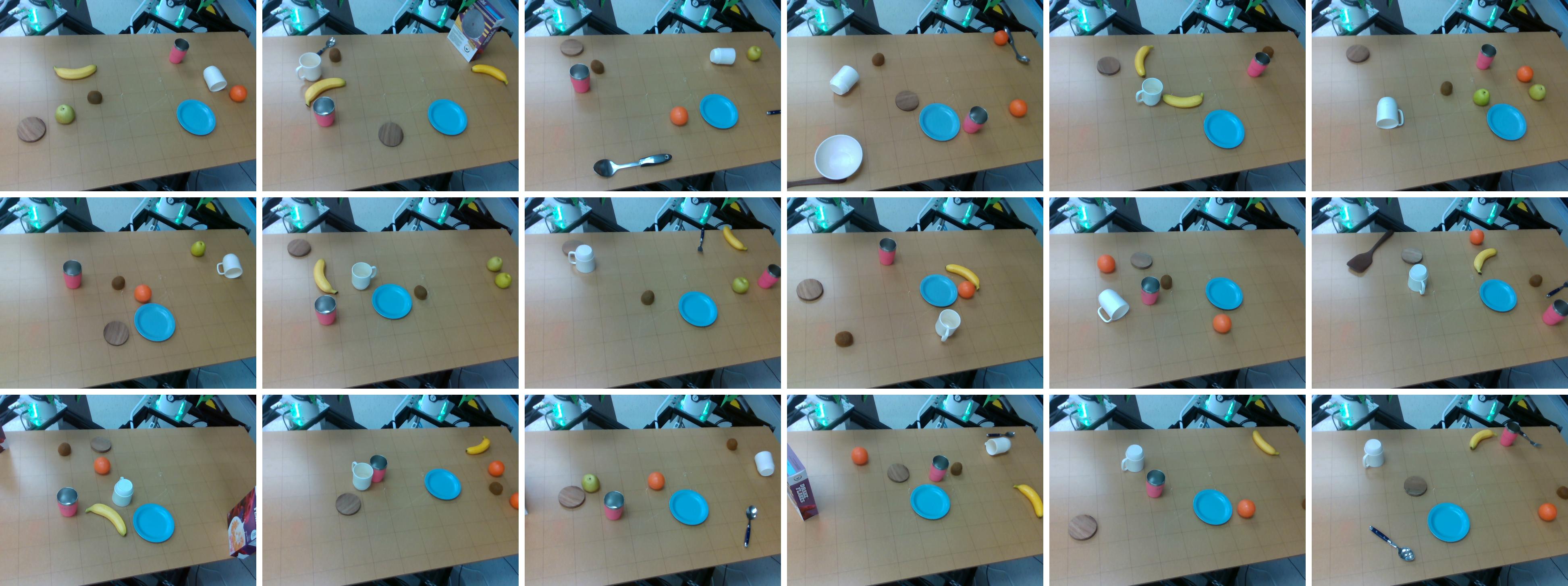}
\caption{Sample of initial conditions for Nominal (Real) on \textbf{salem} 
}
\label{subfig:hardware_ic_overlays_w_and_w_o_distribution_shift2_nominal}
\end{subfigure}
\hfill
\begin{subfigure}[t]{.33\linewidth}
\includegraphics[height=4.05cm]{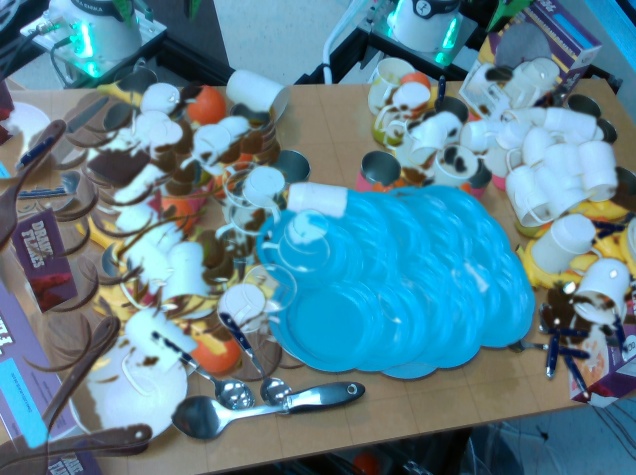}
\caption{All ICs overlaid to show distribution}
\label{subfig:hardware_ic_overlays_w_and_w_o_distribution_shift2_nominal_all}
\end{subfigure}

\vspace{0.50em}

\begin{subfigure}[t]{.655\linewidth}
\includegraphics[height=4.05cm]{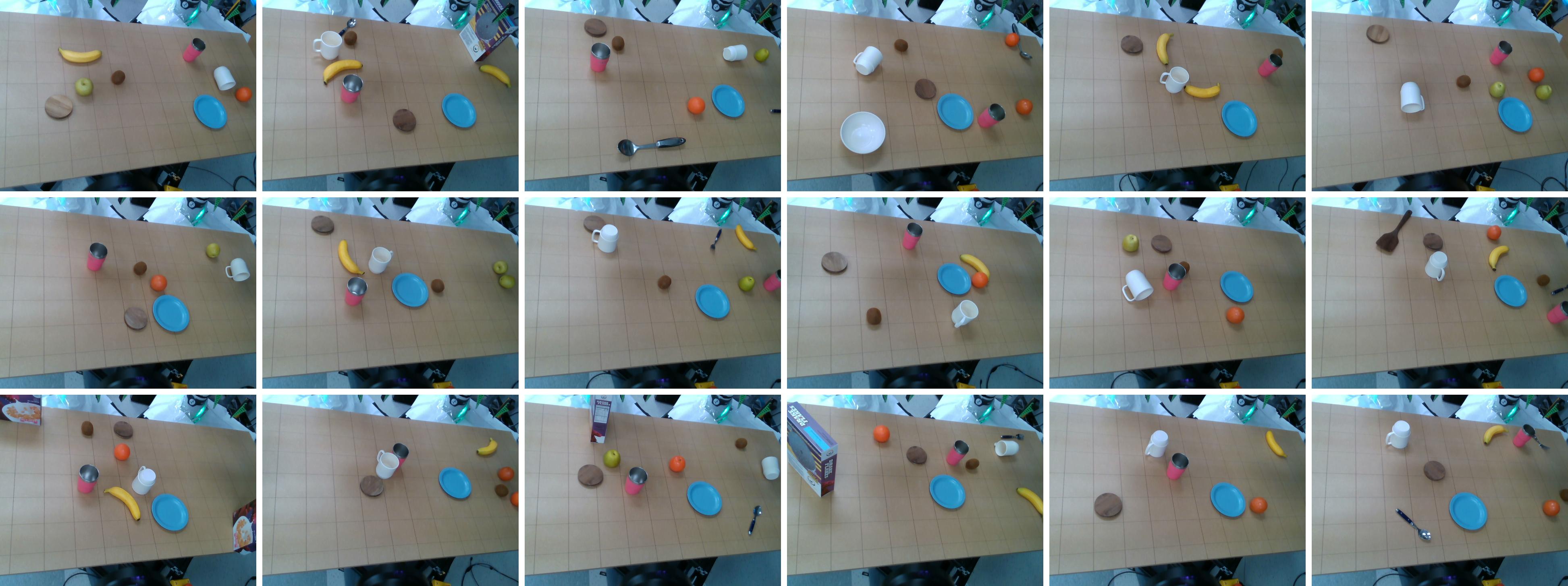}
\caption{Sample of initial conditions for Station Distribution Shift (Real) on \textbf{milton} 
}
\label{subfig:hardware_ic_overlays_w_and_w_o_distribution_shift2_novel_station}
\end{subfigure}
\hfill
\begin{subfigure}[t]{.33\linewidth}
\includegraphics[height=4.05cm]{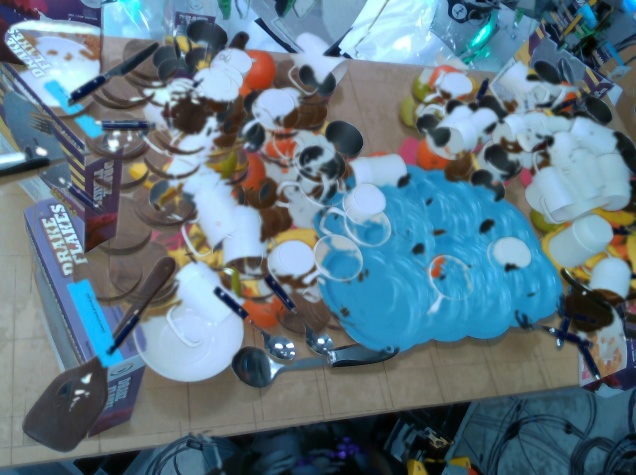}
\caption{All ICs overlaid to show distribution}
\label{subfig:hardware_ic_overlays_w_and_w_o_distribution_shift2_novel_station_all}
\end{subfigure}

\vspace{0.50em}

\begin{subfigure}[t]{.655\linewidth}
\includegraphics[height=4.05cm]{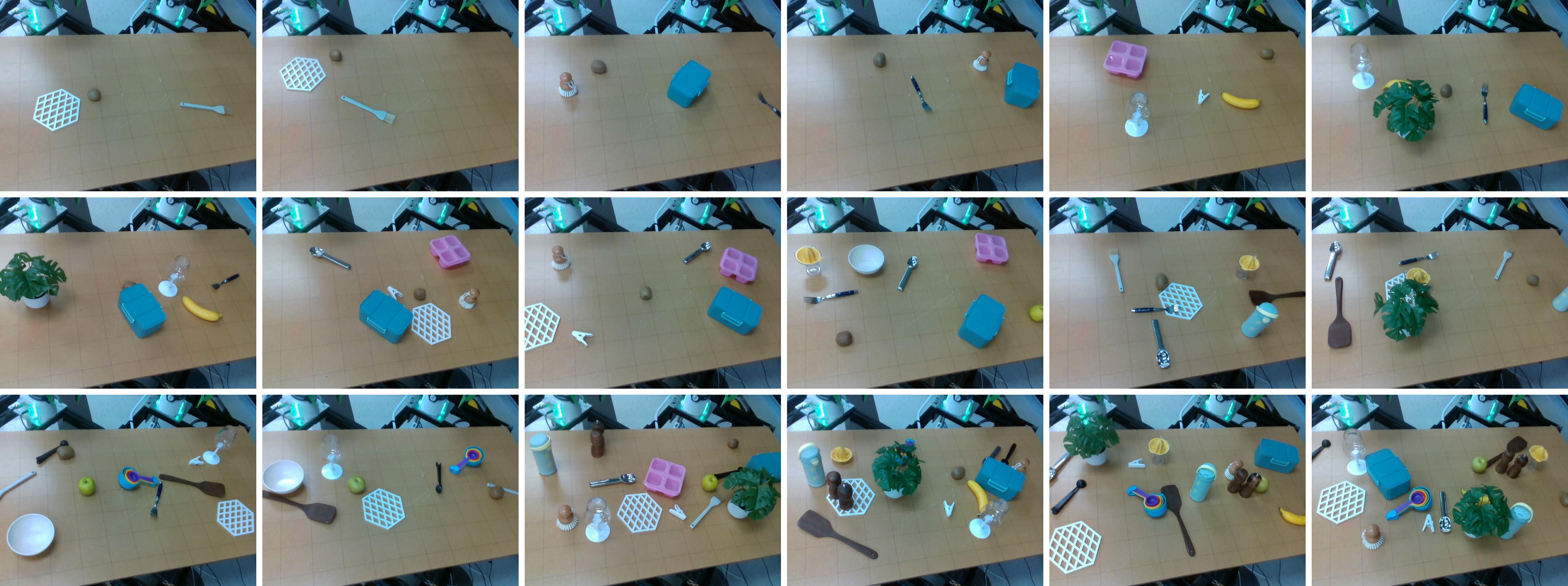}
\caption{Sample of initial conditions for Object-Centric Distribution Shift (Real) on \textbf{salem} 
}
\label{subfig:hardware_ic_overlays_w_and_w_o_distribution_shift2_dist_shift}
\end{subfigure}
\hfill
\begin{subfigure}[t]{.33\linewidth}
\includegraphics[height=4.05cm]{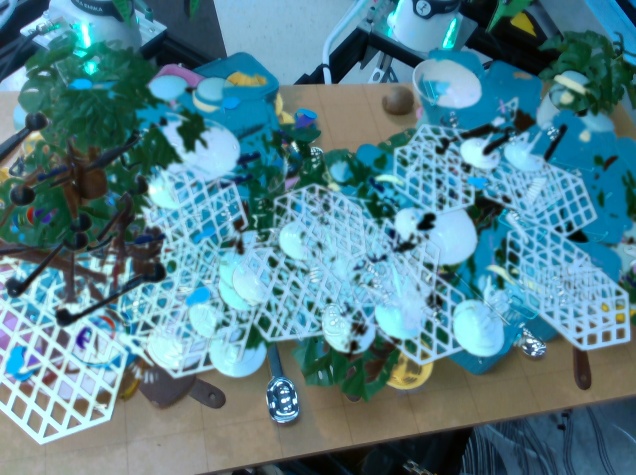}
\caption{All ICs overlaid to show distribution}
\label{subfig:hardware_ic_overlays_w_and_w_o_distribution_shift2_dist_shift_all}
\end{subfigure}

\caption{\small\textbf{Samples of initial conditions}  for nominal distribution (a), station distribution shift (c) and object-centric distribution shift (e) 
for the \textit{PutKiwiInCenterOfTable} task. We present overlays of all initial conditions for each condition (b,d,f). 
}
\label{fig:hardware_ic_overlays_w_and_w_o_distribution_shift2}
}

\adaptivefigure{
\centering

\begin{subfigure}[t]{.655\linewidth}
\includegraphics[height=4.05cm]{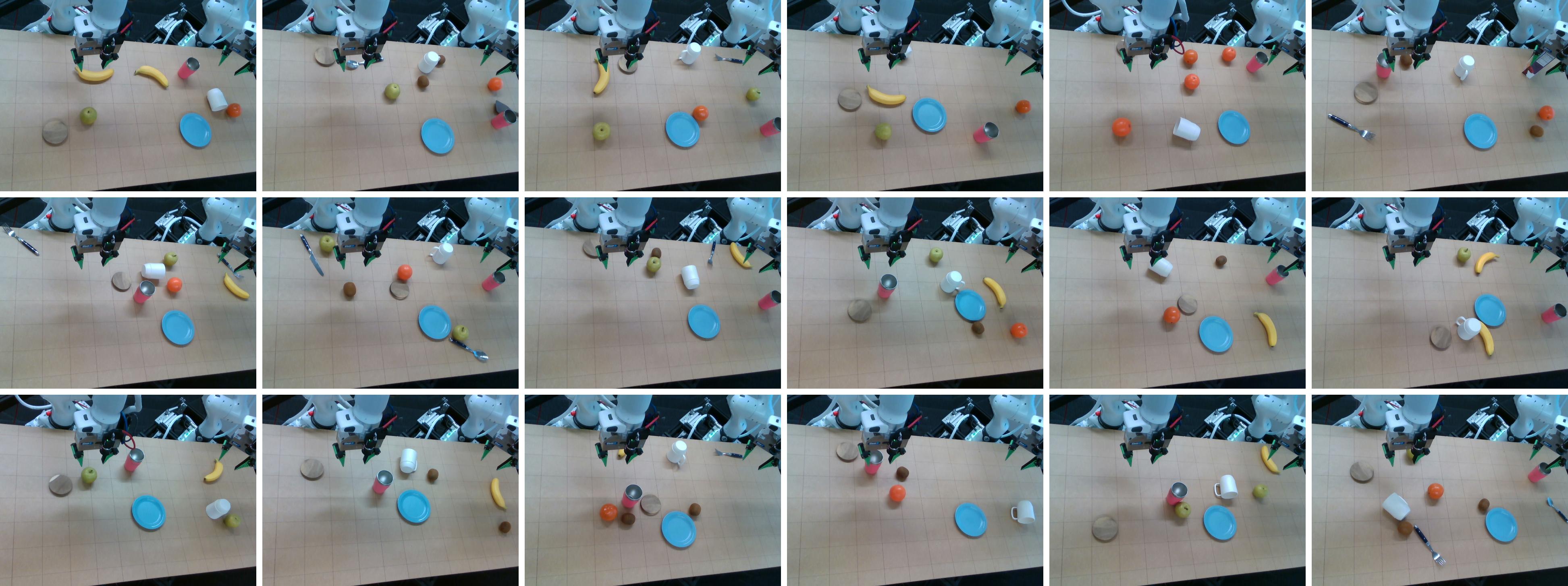}
\caption{Sample of initial conditions for Nominal (Real) \textbf{davis} 
}
\label{subfig:hardware_ic_overlays_w_and_w_o_distribution_shift2_TurnMugRightsideUp_nominal}
\end{subfigure}
\hfill
\begin{subfigure}[t]{.33\linewidth}
\includegraphics[height=4.05cm]{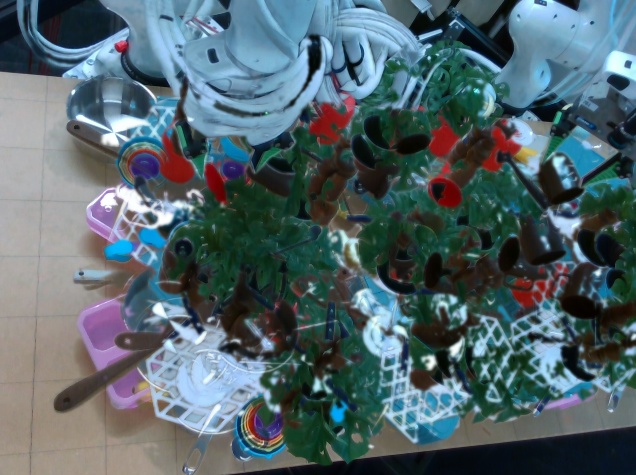}
\caption{All ICs overlaid to show distribution}
\label{subfig:hardware_ic_overlays_w_and_w_o_distribution_shift2_TurnMugRightsideUp_nominal_all}
\end{subfigure}

\vspace{0.50em}

\begin{subfigure}[t]{.655\linewidth}
\includegraphics[height=4.05cm]{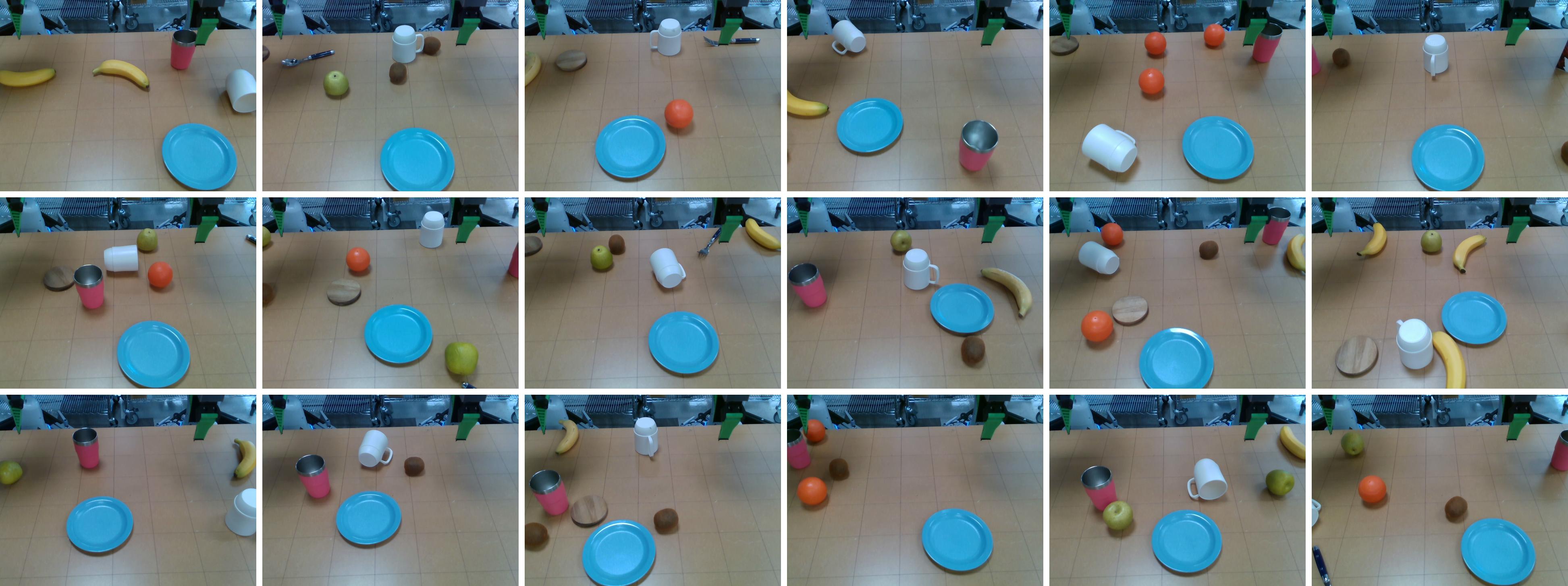}
\caption{Sample of initial conditions for Station Distribution Shift (Real) on \textbf{maverick} 
}
\label{subfig:hardware_ic_overlays_w_and_w_o_distribution_shift2_TurnMugRightsideUp_novel_station}
\end{subfigure}
\hfill
\begin{subfigure}[t]{.33\linewidth}
\includegraphics[height=4.05cm]{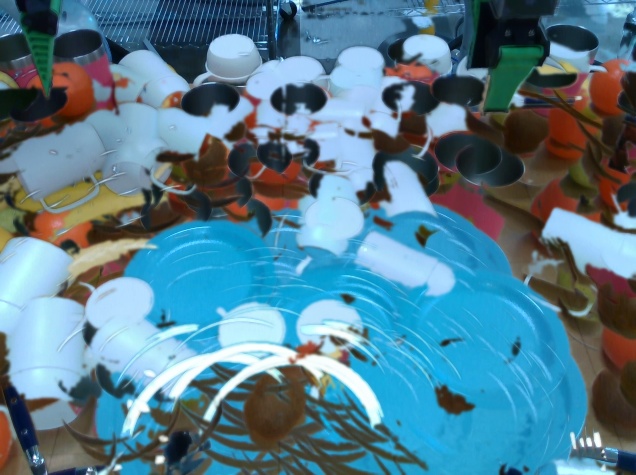}
\caption{All ICs overlaid to show distribution}
\label{subfig:hardware_ic_overlays_w_and_w_o_distribution_shift2_TurnMugRightsideUp_novel_station_all}
\end{subfigure}

\vspace{0.50em}

\begin{subfigure}[t]{.655\linewidth}
\includegraphics[height=4.05cm]{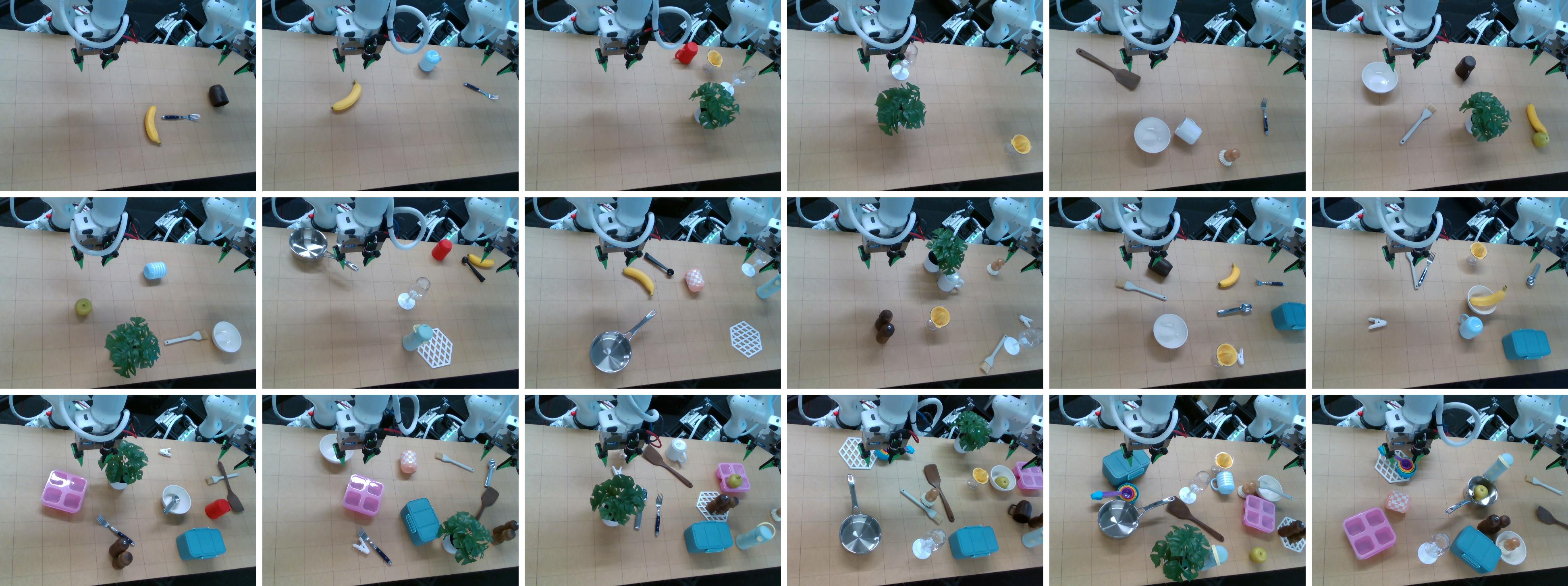}
\caption{Sample of initial conditions for Object-Centric Distribution Shift (Real) on \textbf{davis} 
}
\label{subfig:hardware_ic_overlays_w_and_w_o_distribution_shift2_TurnMugRightsideUp_dist_shift}
\end{subfigure}
\hfill
\begin{subfigure}[t]{.33\linewidth}
\includegraphics[height=4.05cm]{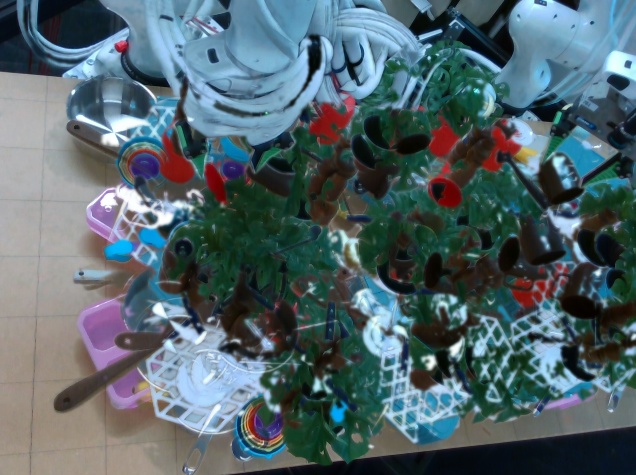}
\caption{All ICs overlaid to show distribution}
\label{subfig:hardware_ic_overlays_w_and_w_o_distribution_shift2_TurnMugRightsideUp_dist_shift_all}
\end{subfigure}

\caption{\small\textbf{Samples of initial conditions}  for nominal distribution (a), station distribution shift (c) and object-centric distribution shift (e) 
for the \textit{TurnMugRightsideUp} task. We present overlays of all initial conditions for each condition (b,d,f).
}
\label{fig:hardware_ic_overlays_w_and_w_o_distribution_shift2_TurnMugRightsideUp}
}

Fig.~\ref{fig:novel_object_overview} shows an overview of the object variation for the real-world experiments, with the objects used under \textit{nominal conditions} shown in the top row and the objects used under \textit{object-centric distribution shift} in the bottom row. Note that for the objects used under distribution shift we vary color, texture as well as shape. The distribution-shift setting is particularly challenging for the tasks involving mugs and coasters, while the tasks involving kiwis have relatively less variation in the manipulands. For the task involving the kiwi, a kiwi was selected randomly from a bin for each IC test bundle.

\adaptivefigure{
    \centering
    \includegraphics[width=\textwidth]{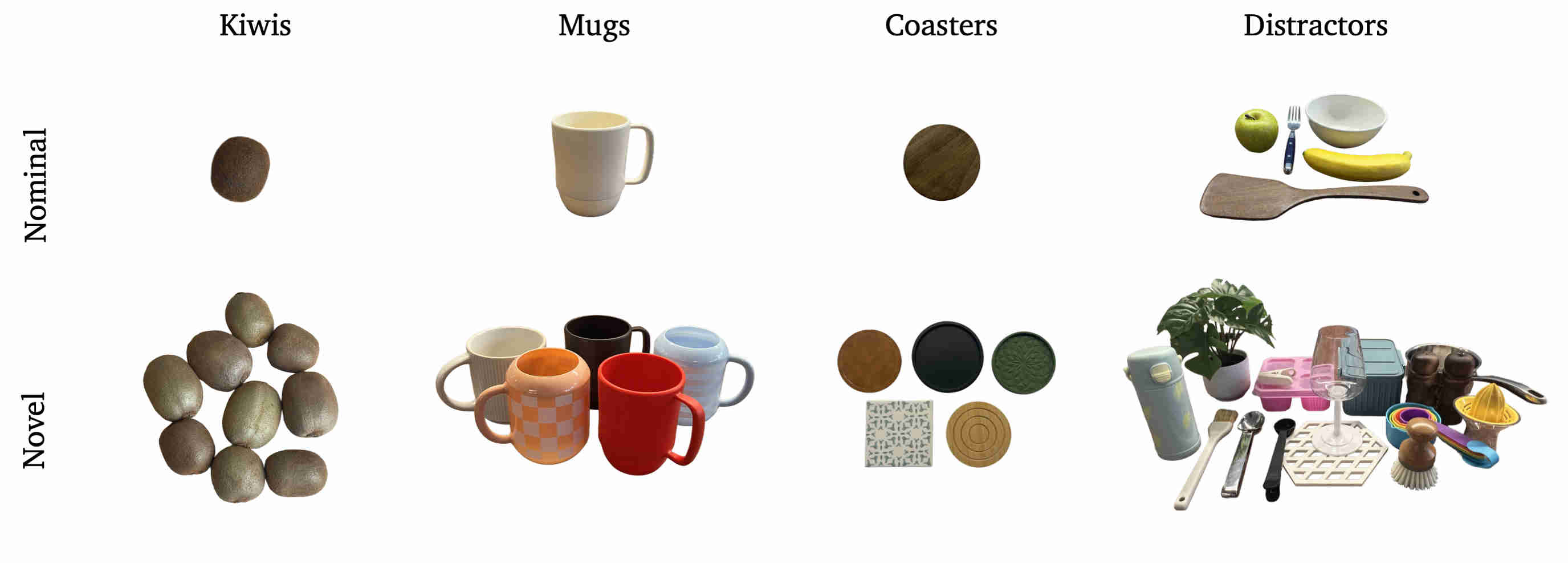}
    \caption{\small Qualitative overview of object variation for real-world \textbf{Distribution Shift} experiments. (top) Objects seen during pretraining (bottom). Novel objects used as manipulands or distractors in the \textbf{Distribution Shift} setting, which vary in shape and color. Clutter is sampled from the combined set of Nominal and Novel distractors at varying levels. Objects are not pictured to scale and only a representative set of kiwis pictured; a new kiwi was randomly chosen from a larger set for each new IC evaluated.}
    \label{fig:novel_object_overview}
}

\subsection{Rubrics for real-world tasks}
\label{subsec:supp_rubrics_hardware}

We present the rubrics used to calculate task completion for the ``unseen" real-world tasks. Each question corresponds to a milestone towards task completion and can only be answered with a Yes or No answer. Note that the rubrics are scored manually by human operators. For the corresponding quantitative results, refer to Section~\ref{subsec:LBMs_unseen_tasks}. 

\vspace{1em}

{
\noindent
\scriptsize
\begin{tabular}{p{0.95\linewidth}}
\textbf{BimanualClearKitchenCounter} \\
\hline
Robot picked up and placed all tools in the tray receptacle \\
Robot picked up the sponge \\
Robot held the cutting board in place \\
Robot cleaned the cutting board with the sponge \\
Robot moved the cutting board aside to make room for cleaning \\
Robot swept the waste into the trash bin \\
Robot dropped the sponge and returned to the home position \\
\end{tabular}
}

\vspace{1em}

{
\noindent
\scriptsize
\begin{tabular}{p{0.95\linewidth}}
\textbf{BimanualSetUpBreakfastTable} \\
\hline
Robot opened the cabinet door \\
Robot picked the bowl up and placed it on the tray \\
Robot placed the milk cup on the tray \\
Robot placed the juice cup on the tray \\
Are the milk and juice cups upright? \\
Robot placed the plate on the tray \\
Robot put the apple on the tray \\
Robot put the banana on the tray \\
Robot picked up at least one toast from the toaster \\
Robot put the toast on the plate \\
Robot poured the cereal into the bowl \\
Robot put the spoon on the tray \\
\end{tabular}
}

\vspace{1em}

{
\noindent
\scriptsize
\begin{tabular}{l}
\textbf{BimanualBikeRotorInstall} \\
\hline
Robot picked up rotor \\
Robot handed over the rotor to other arm \\
Robot placed the rotor on the bike wheel\\
Robot seated the rotor onto the bike wheel \\
Robot picked up the lockring \\
Robot placed the lockring over the rotor \\
Robot handed over the tool to other arm \\
Robot placed the tool onto lockring \\
Robot positioned the tool into lockring grooves \\
Robot tightened the lockring at least one revolution \\
Robot engaged the threads of the lockring \\
Robot took off the tool from the lockring \\
Robot puts the tool back onto the table \\
\end{tabular}
}

\vspace{1em}

{
\noindent
\scriptsize
\begin{tabular}{l}
\textbf{CutAppleInSlices} \\
\hline
Robot grasped the apple \\
Robot grabed the corer and aligned it with apple center \\
Robot cored the apple successfully \\
Robot unsheathed the knife and aligned it with the center of the apple \\
Robot sliced the apple into two halves \\
Robot fliped one half of apple \\
Robot fliped the other half of apple \\
Robot sliced one half of the apple in at least 3 slices \\
Robot picked up the cloth \\
Robot wiped the knife \\
Robot placed the cloth on top of the shelf \\
Robot picked up the knife sheath \\
Robot sheathed the knife \\
Robot placed the knife into the utensil crock \\
\end{tabular}
}

\vspace{1em}

{
\noindent
\scriptsize
\begin{tabular}{l}
\textbf{CleanLitterBox} \\
\hline
Robot grasped the cat litter scoop \\
Robot scooped up the cat excrement \\
Robot used the other arm to open the trash can \\
Robot poured the cat excrement into the trash can \\
Robot repeated scooping until all excrement is removed from the litter box \\
Robot closed the trash can \\
Robot placed the scoop back onto the table \\
\end{tabular}
}

%% file: sections/13_appendix_additional_results.tex

\section{``Seen" tasks under nominal conditions analysis: Breakfast Scenario}
\label{subsec:supplemental_breakfast_sceario}

We provide additional insight into the results of evaluating policies on ``seen" tasks during training (Sec.~\ref{subsec:LBMs_seen_tasks} and Fig.~\ref{fig:seen_tasks_sim_and_real}), specifically in the \textit{Breakfast (B)} scenario.

 We note that task success rate correlates with number of demonstrations, i.e., tasks \textit{PutBananaOnSaucer} and \textit{PutKiwiInCenterOfTable} have only 49 demonstrations and low success rate (i.e., single-task success rate of 19.5\% and 13.5\%, respectively), while \textit{PlaceCupByCoaster} (196 demonstrations) and \textit{TurnCupUpsideDown} (490 demonstrations) have significantly higher success rates (i.e., single-task success rate of 50\% and 68.5\%). By inspecting many rollouts, we identified two common failure modes: LBMs performing wrong tasks, and policies idling at the beginning of rollouts. To further introspect the performance of our policies on these tasks, we present 1) qualitative results showing the terminal frame of rollouts for each policy (i.e., the last image recorded during a rollout)-- this indicates how well a task was executed, and whether there was any confusion with respect to the language command provided; and 2) a quantitative measure of when the first motion was initiated by the policy-- this indicates whether the policy started executing the task with a delay, leading to timeouts. 

\subsection{PutKiwiInCenterOfTable and PutBananaOnSaucer}
\label{subsec:supp_qual_analysis_put_kiwi_banana} 
Figures~\ref{fig:supp_qual_analysis_put_kiwi} and~\ref{fig:supp_qual_analysis_put_banana} show the  final frames of the policy rollouts, and Figure~\ref{fig:supp_qual_analysis_seen_single_skill_pause_histogram} presents an analysis of the time when the robot's first motion occurred during rollouts. When inspecting single-task policies' rollouts, the robot always performs the correct task but often misses the target manipuland or is unable to place it correctly. Pretrained and finetuned LBM have two common main failure modes:
1) the robot is unable to move away from its starting pose, as seen in Figure~\ref{fig:supp_qual_analysis_seen_single_skill_pause_histogram}; 2) the robot performs a different task, as seen in  Figures~\ref{fig:supp_qual_analysis_put_kiwi} and~\ref{fig:supp_qual_analysis_put_banana}. In both cases, most rollouts end in failure. The finetuned LBM performs better than the pretrained LBM by staying static less often, and by performing the correct task more often.

\subsection{TurnCupUpsideDown}
\label{subsec:supp_qual_analysis_put_turncup}
\begin{figure}
\begin{subfigure}[b]{\linewidth}
\includegraphics[width=\linewidth]{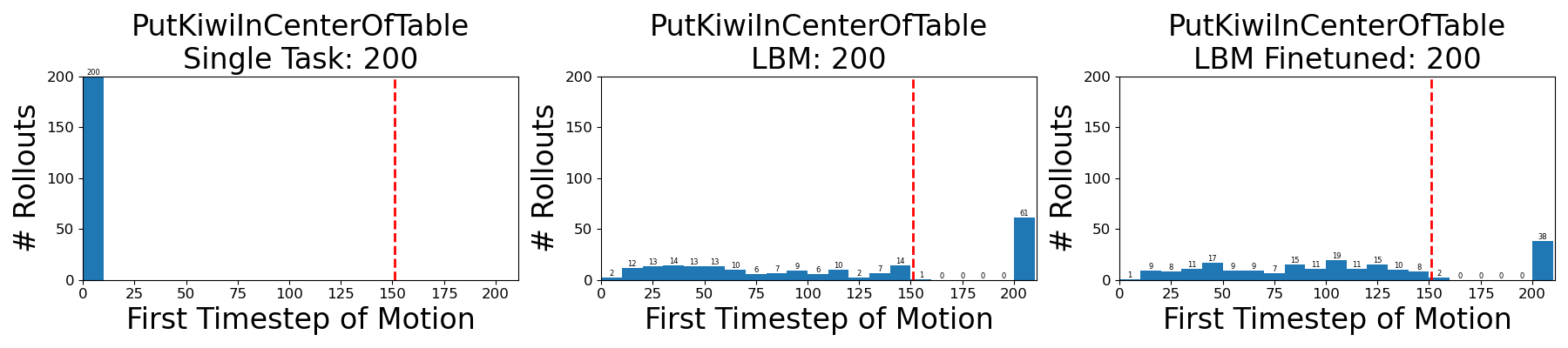}
\caption{\textit{PutKiwiInCenterOfTable} (Simulation - nominal)}
\end{subfigure}
\begin{subfigure}[b]{\linewidth}
\includegraphics[width=\linewidth]{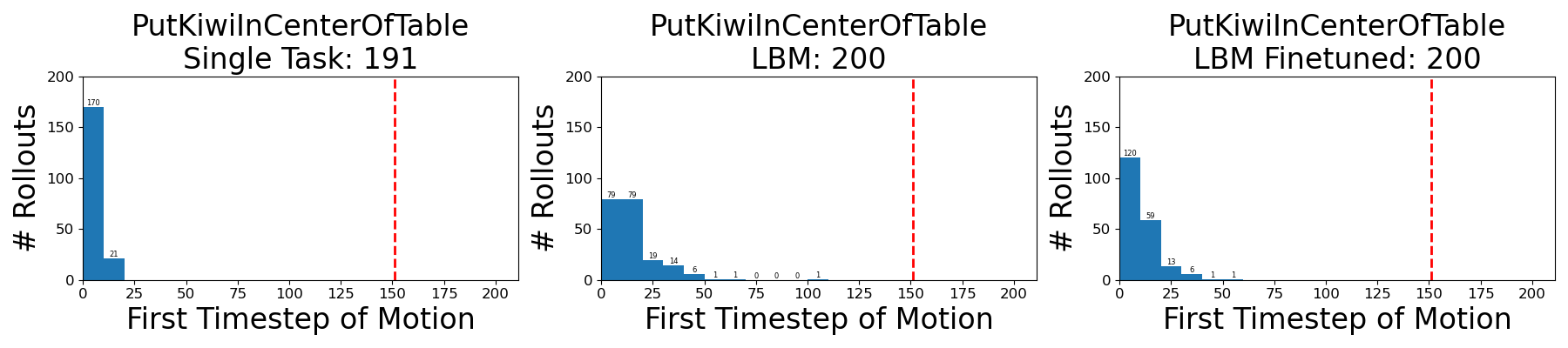}
\caption{\textit{PutKiwiInCenterOfTable} (Simulation - distribution shift)}
\end{subfigure}
\begin{subfigure}[b]{\linewidth}
\includegraphics[width=\linewidth]{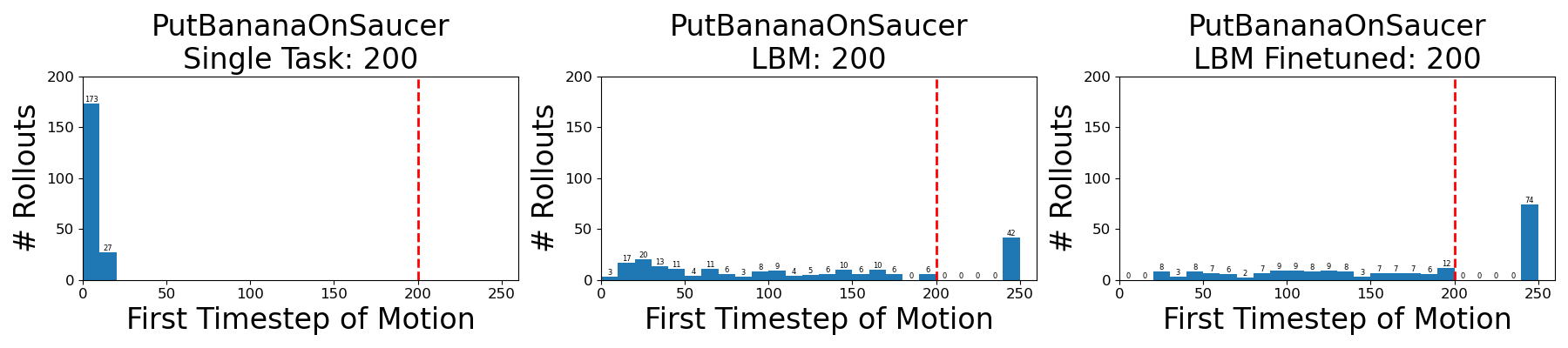}
\caption{\textit{PutBananaOnSaucer} (Simulation - nominal)}
\end{subfigure}
\begin{subfigure}[b]{\linewidth}
\includegraphics[width=\linewidth]{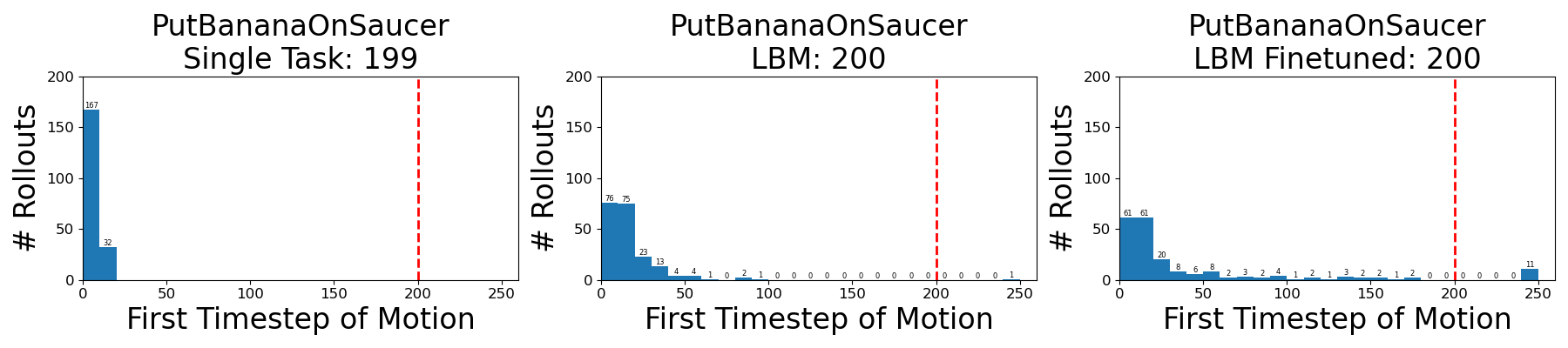}
\caption{\textit{PutBananaOnSaucer} (Simulation - distribution shift)}

\end{subfigure}
\begin{subfigure}[b]{\linewidth}
\includegraphics[width=\linewidth]{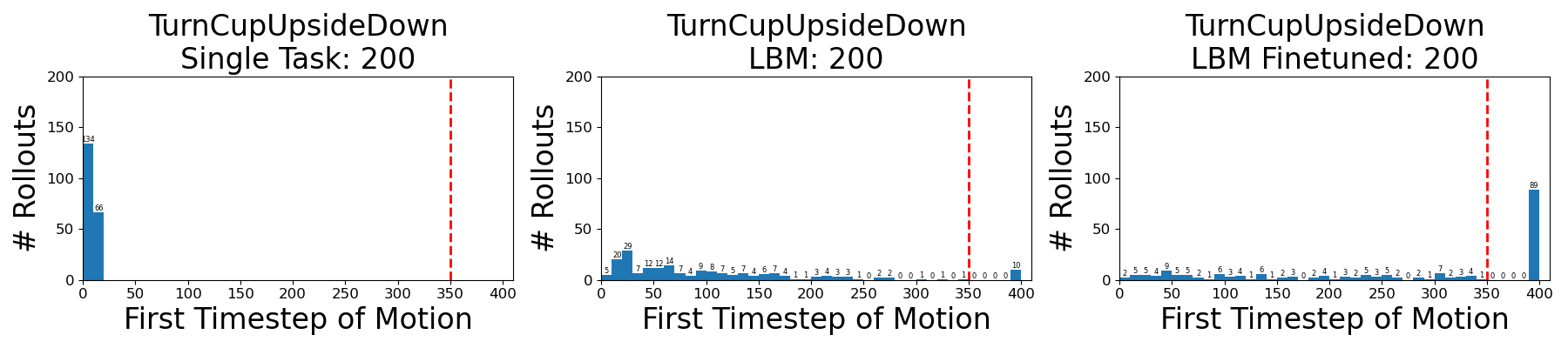}
\caption{\textit{TurnCupUpsideDown} (Simulation - nominal)}
\end{subfigure}
\begin{subfigure}[b]{\linewidth}
\includegraphics[width=\linewidth]{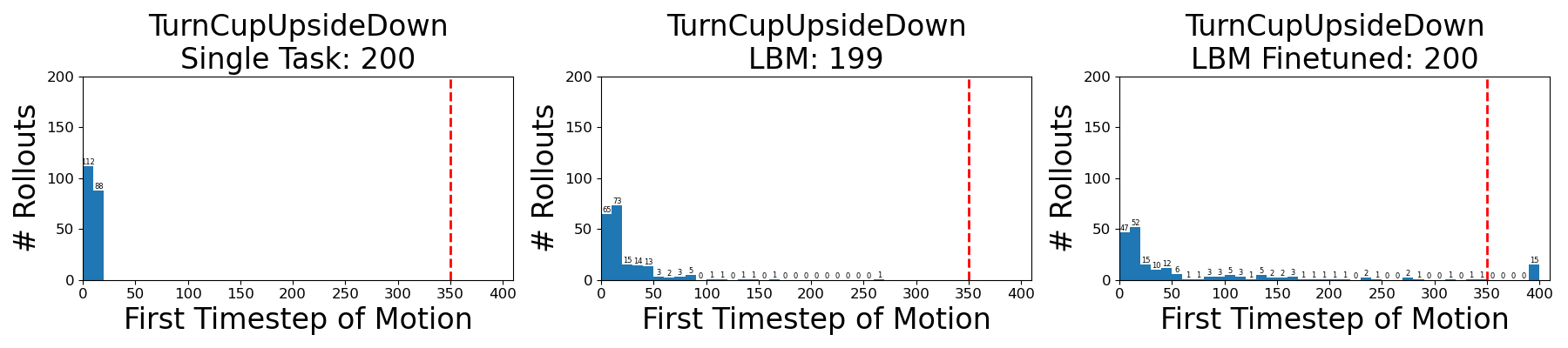}
\caption{\textit{TurnCupUpsideDown} (Simulation - distribution shift)}

\end{subfigure}

\begin{subfigure}[b]{\linewidth}
\includegraphics[width=\linewidth]{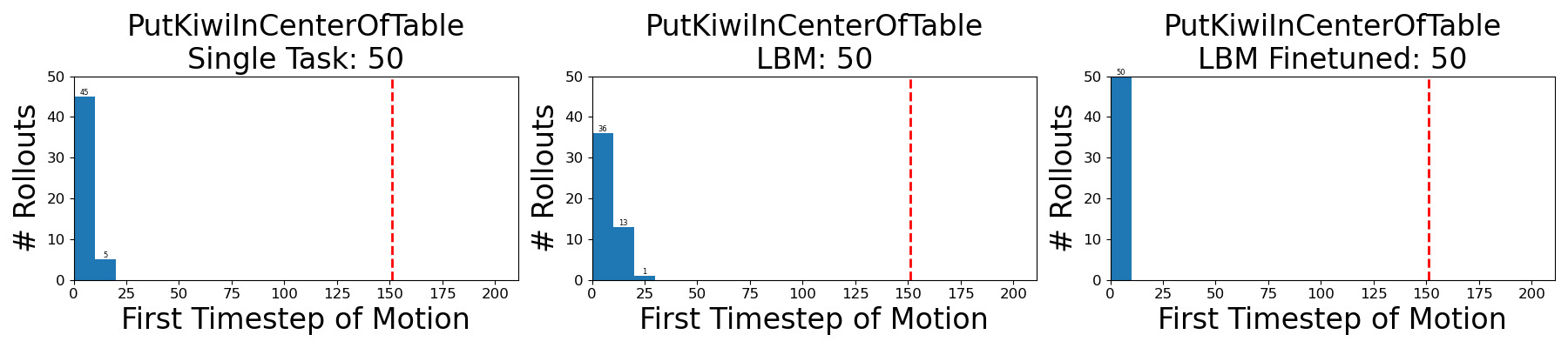}
\caption{\textit{PutKiwiInCenterOfTable} (Real-world - nominal)}
\end{subfigure}
\caption{\small\textbf{Histogram of robot's first time of motion} (defined in \ref{subsec:pretraining_dataset_filtering}). Red lines indicate timeout. First column is single-task, second column is pretrained LBM, and third is finetuned LBM. For the three simulation tasks shown here (subplots (a) to (f)), the pretrained and finetuned LBM take significantly longer than single-task to start performing any actions, and in many rollouts, the robot never moved. When LBM policies are tested under distribution shift, they tend to move much sooner than nominal situations. Additionally, in subplot (g) we show the real-world rollouts for the \textit{PutKiwiInCenterOfTable} task to highlight the differences in execution: in real-world, LBMs do not exhibit the same delays before initiating motion as in simulation.}
\label{fig:supp_qual_analysis_seen_single_skill_pause_histogram}
\end{figure}

For this task, the finetuned LBM performance is worse than the single-task or pretrained LBM performance. The finetuned LBM often fails to initiate any motion for this task, as seen in  Figure~\ref{fig:supp_qual_analysis_seen_single_skill_pause_histogram}. This behavior is further investigated in Sec.~\ref{sec:supp_pauses}. By inspecting the final frames for this task (Figure~\ref{fig:supp_qual_analysis_turn_cup}), we see that the pretrained LBM tends to execute the wrong task, while the single-task baseline sometimes fails when attempting to grasp the object.

\adaptivefigure{
\begin{subfigure}[b]{0.23\linewidth}
\includegraphics[width=\linewidth]{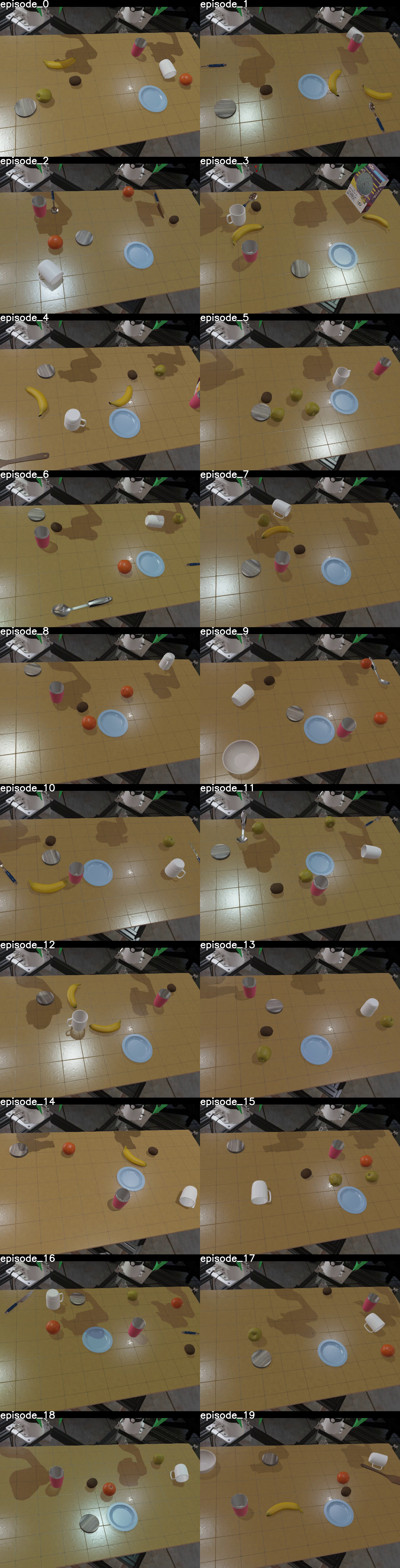}
\caption{Initial conditions}
\end{subfigure}
\begin{subfigure}[b]{0.23\linewidth}
\includegraphics[width=\linewidth]{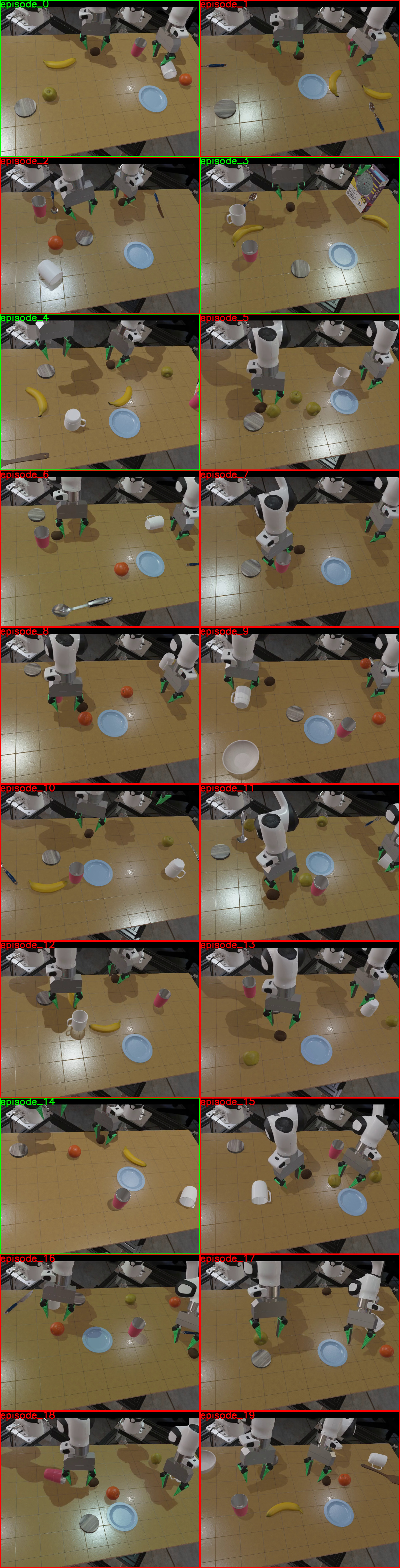}
\caption{Single Task}
\end{subfigure}
\begin{subfigure}[b]{0.23\linewidth}
\includegraphics[width=\linewidth]{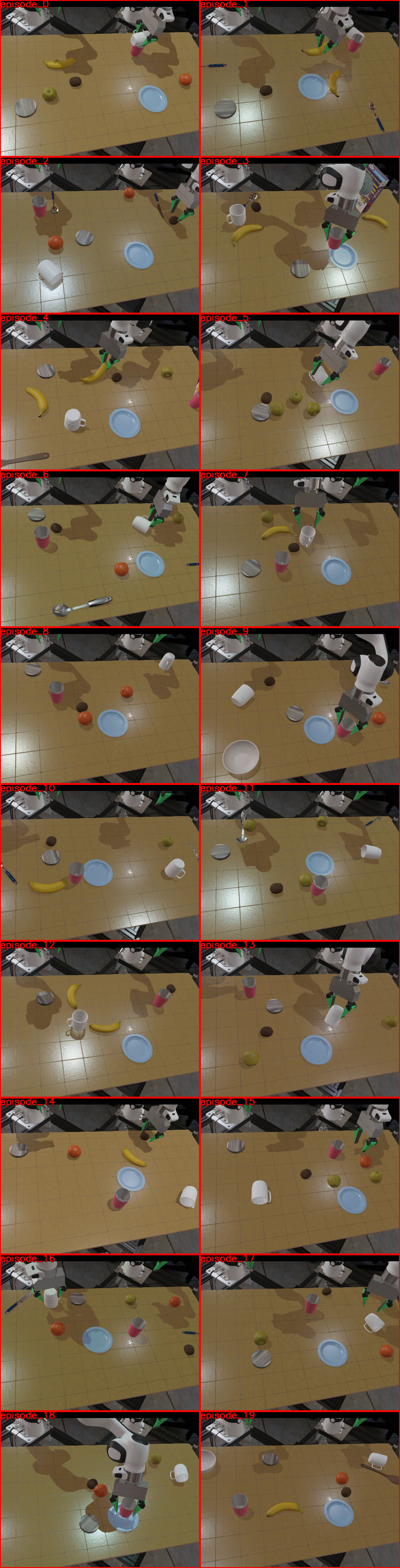}
\caption{Pretrained LBM}
\end{subfigure}
\begin{subfigure}[b]{0.23\linewidth}
\includegraphics[width=\linewidth]{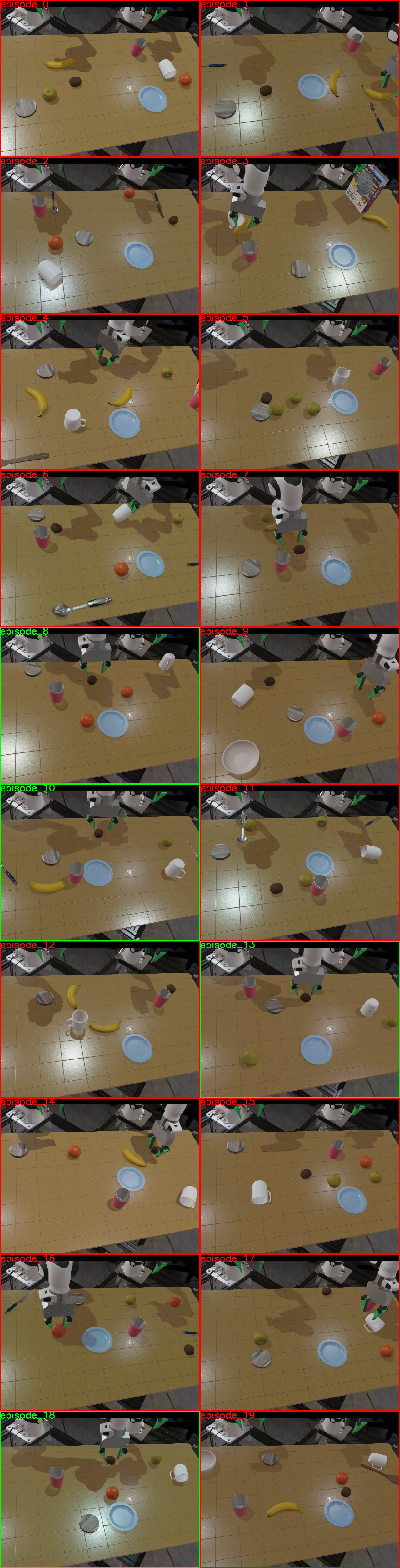}
\caption{Finetuned LBM}
\end{subfigure}
\caption{\small\textbf{Tiled terminal frames of rollouts from different policies performing \textit{PutKiwiInCenterOfTable} under nominal conditions.} The single-task policy often missed the kiwi while grasping. The pretrained LBM often manipulates the wrong object, and sometimes suffers from the inability to move. The finetuned LBM suffers heavily from the inability to move and sometimes performs the wrong task. The green outline around a rollout indicates success, while the red outline indicates failure.}
\label{fig:supp_qual_analysis_put_kiwi}
}

\adaptivefigure{
\begin{subfigure}[b]{0.23\linewidth}
\includegraphics[width=\linewidth]{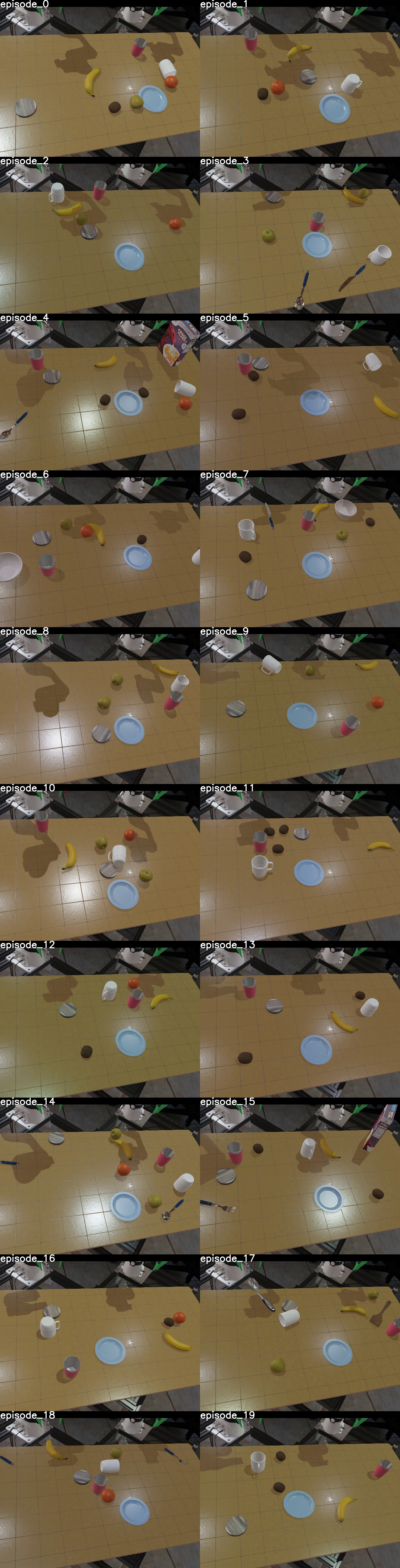}
\caption{Initial conditions}
\end{subfigure}
\begin{subfigure}[b]{0.23\linewidth}
\includegraphics[width=\linewidth]{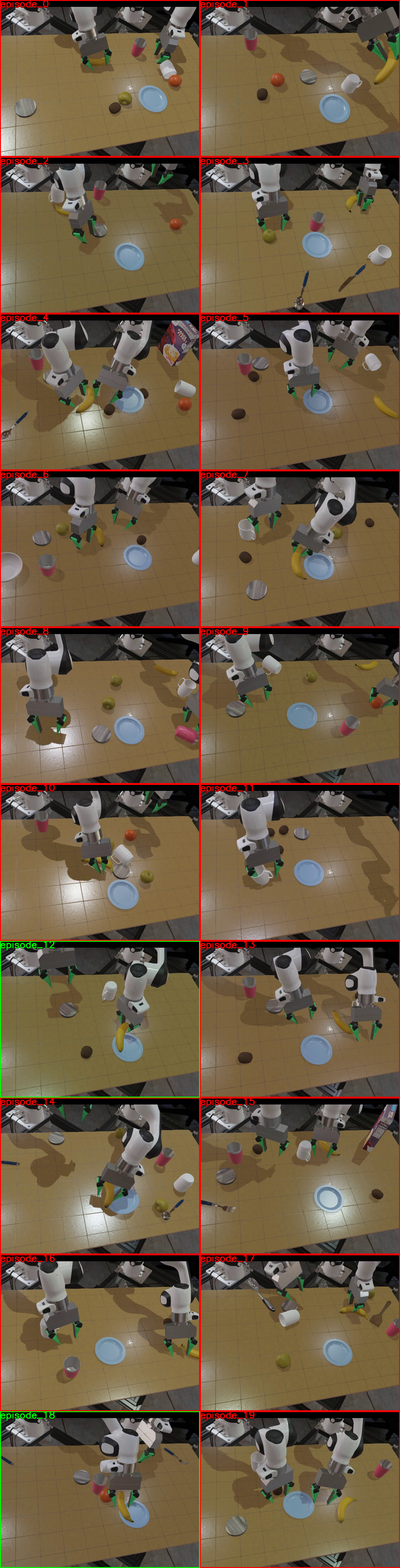}
\caption{Single Task}
\end{subfigure}
\begin{subfigure}[b]{0.23\linewidth}
\includegraphics[width=\linewidth]{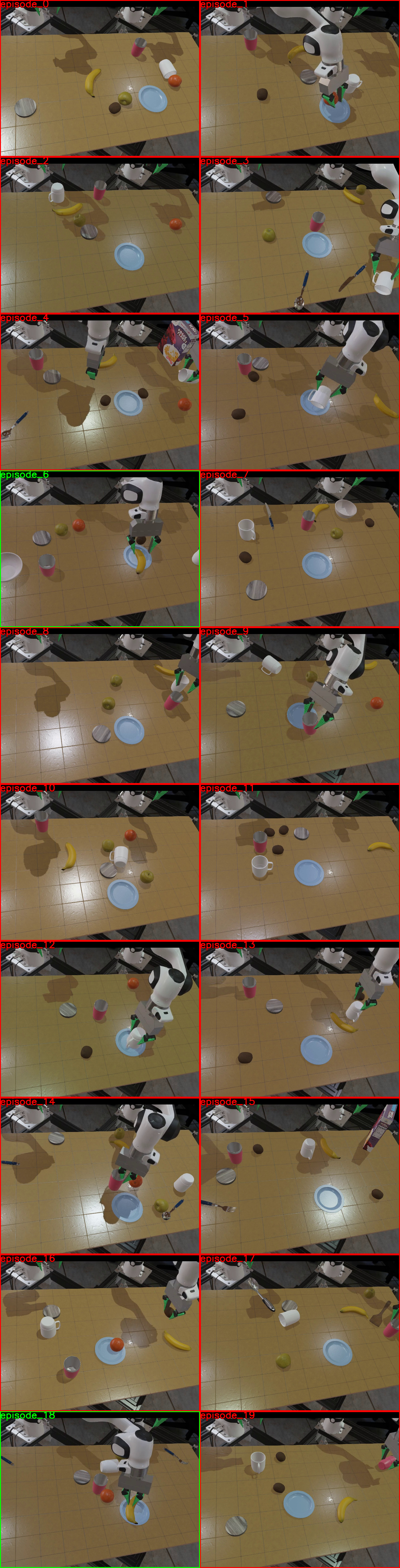}
\caption{Pretrained LBM}
\end{subfigure}
\begin{subfigure}[b]{0.23\linewidth}
\includegraphics[width=\linewidth]{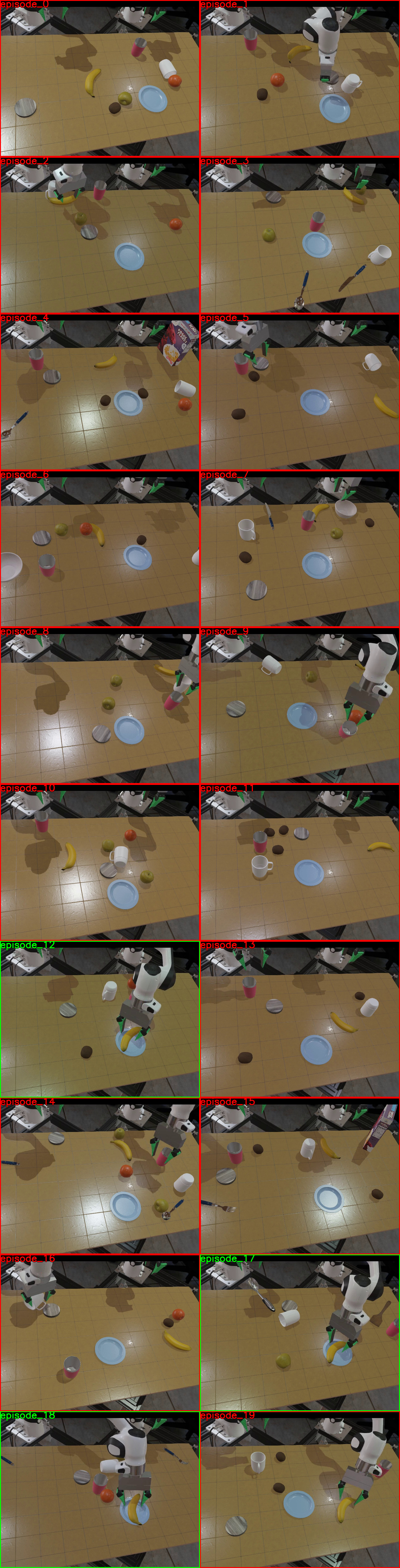}
\caption{Finetuned LBM}
\end{subfigure}
\caption{\small\textbf{Tiled terminal frames of rollouts from different policies performing \textit{PutBananaOnSaucer} under nominal conditions.} The single-task policy often fails to grasp the banana or to place it correctly afterwards. The pretrained LBM's main failure mode is manipulating other objects. The finetuned LBM suffers heavily from the inability to move. The green outline around a rollout indicates success, while the red outline indicates failure.}
\label{fig:supp_qual_analysis_put_banana}
}

\adaptivefigure{
\begin{subfigure}[b]{0.23\linewidth}
\includegraphics[width=\linewidth]{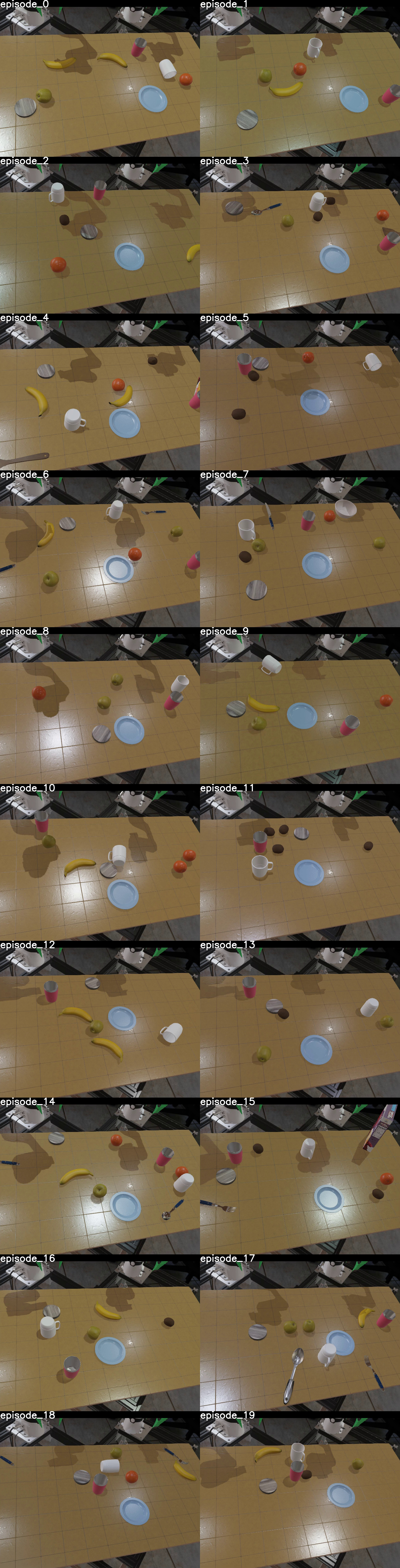}
\caption{Initial conditions}
\end{subfigure}
\begin{subfigure}[b]{0.23\linewidth}
\includegraphics[width=\linewidth]{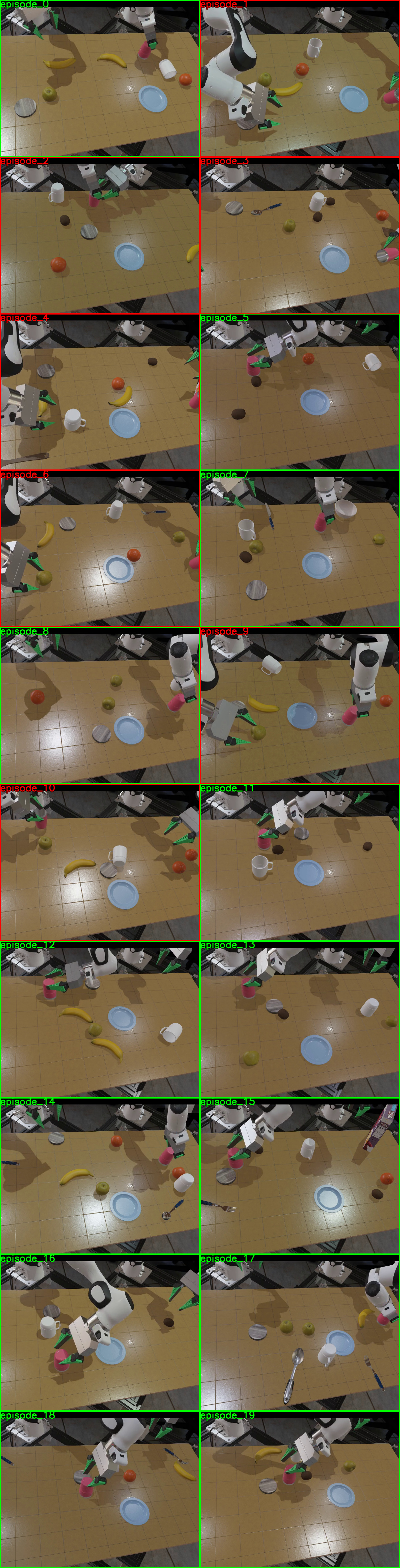}
\caption{Single Task}
\end{subfigure}
\begin{subfigure}[b]{0.23\linewidth}
\includegraphics[width=\linewidth]{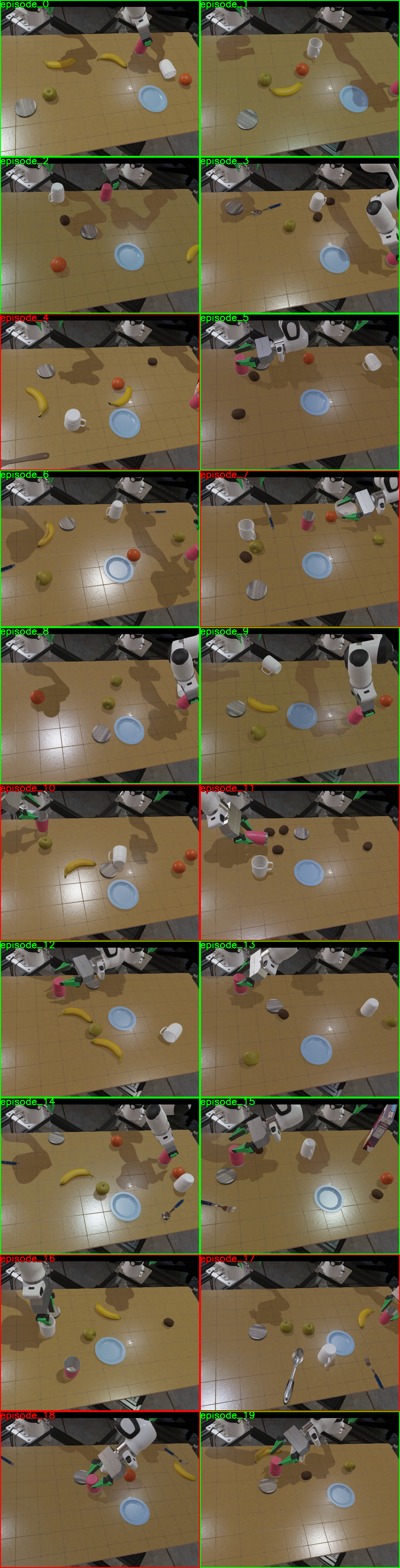}
\caption{Pretrained LBM}
\end{subfigure}
\begin{subfigure}[b]{0.23\linewidth}
\includegraphics[width=\linewidth]{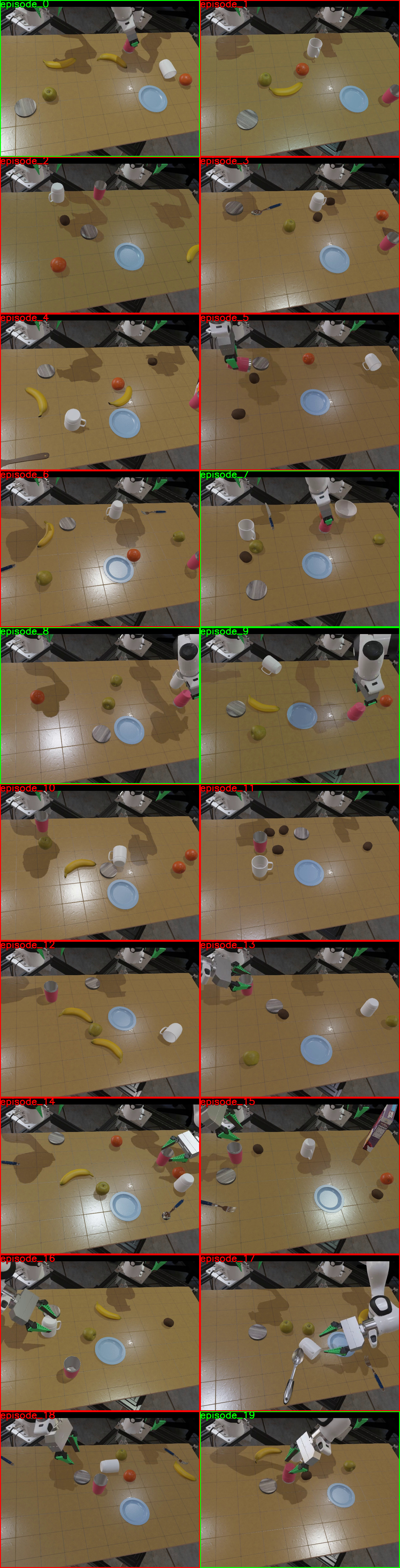}
\caption{Finetuned LBM}
\end{subfigure}
\caption{\small\textbf{Tiled terminal frames of rollouts from different policies performing \textit{TurnCupUpsideDown} under nominal conditions.} The single-task policy performs reasonably, and its main failure mode is when initiating grasps. The pretrained LBM sometimes performs the wrong tasks. The finetuned LBM suffers from the inability to move, further analyzed in Figure~\ref{fig:supp_qual_analysis_seen_single_skill_pause_histogram}. The green outline around a rollout indicates success, while the red outline indicates failure.}
\label{fig:supp_qual_analysis_turn_cup}
}

\subsection{Simulation and Real Comparison for PutKiwiInCenterOfTable, TurnMugRightsideUp}
\label{subsec:supp_qual_analysis_sim_and_real}
These skills are evaluated both in simulation and real-world, with large performance gaps as shown in Figure~\ref{fig:seen_tasks_sim_and_real}. One of the main reasons is that timeouts are enforced differently in simulation and real evaluations. In simulation, since evaluations are fully automated, timeouts are set with respect to the beginning of each rollout, and are enforced automatically. In real, timeouts (e.g., due to the robot not moving its end effectors) are determined at the discretion of the operators; see Section~\ref{sec:hardware_eval_details} for more details on our real-world evaluation protocol. This difference gives the real policies an advantage in the sense that they can make mistakes and recover. Moreover, this helps the pretrained and finetuned LBM more due to their tendency to perform an incorrect task and then eventually the correct one. 
After applying the same simulation timeouts to real rollouts (see results in Table~\ref{tab:supp_qual_analysis_sim_and_real_retimed_sr}), the success rates are closer to simulation for \textit{TurnMugRightsideUp}. Further inspecting the rollouts generated when evaluating  \textit{PutKiwiInCenterOfTable}, we found that pretrained and finetuned LBM policies in real have almost no visible pauses at the beginning of each rollout, which is in stark contrast to their simulation counterparts (shown in ~\ref{fig:supp_qual_analysis_seen_single_skill_pause_histogram}). Note that we use the same pretrained LBM  for both simulation and real-world evaluation. Finally, we also show side-by-side terminal frames for all policies in both simulation and real-world in Figures~\ref{fig:supp_qual_analysis_sim_and_real_kiwi} and~\ref{fig:supp_qual_analysis_sim_and_real_turnmug}.

\begin{table}[]
\scriptsize
{%
\begin{tabular}{|c|c|c|c|}
\hline
Task x Policy  &  Sim  &  Real  &  Real w.  \\
   &     &     &  Sim Timeout  \\
\hline
\textit{PutKiwiInCenterOfTable} x single task & 0.135 & 0.44 & 0.2 \\
\textit{PutKiwiInCenterOfTable} x LBM pretrained & 0.02 & 0.2 & 0.04 \\
\textit{PutKiwiInCenterOfTable} x LBM finetuned & 0.09 & 0.82 & 0.7 \\
\hline
\textit{TurnMugRightsideUp} x single task & 0.375 & 0.84 & 0.46 \\
\textit{TurnMugRightsideUp} x LBM pretrained & 0.485 & 0.56 & 0.28 \\
\textit{TurnMugRightsideUp} x LBM finetuned & 0.465 & 0.88 & 0.5 \\
\hline
\end{tabular}}
\caption{\small Success rates for real-world ``seen" tasks. Sim and Real columns are the same as reported in Fig. ~\ref{fig:seen_tasks_sim_and_real}. The last column corresponds to applying the simulation timeout to real rollouts. 
}
\label{tab:supp_qual_analysis_sim_and_real_retimed_sr}
\end{table}

\adaptivefigure{
\begin{subfigure}[b]{0.23\linewidth}
\includegraphics[width=\linewidth]{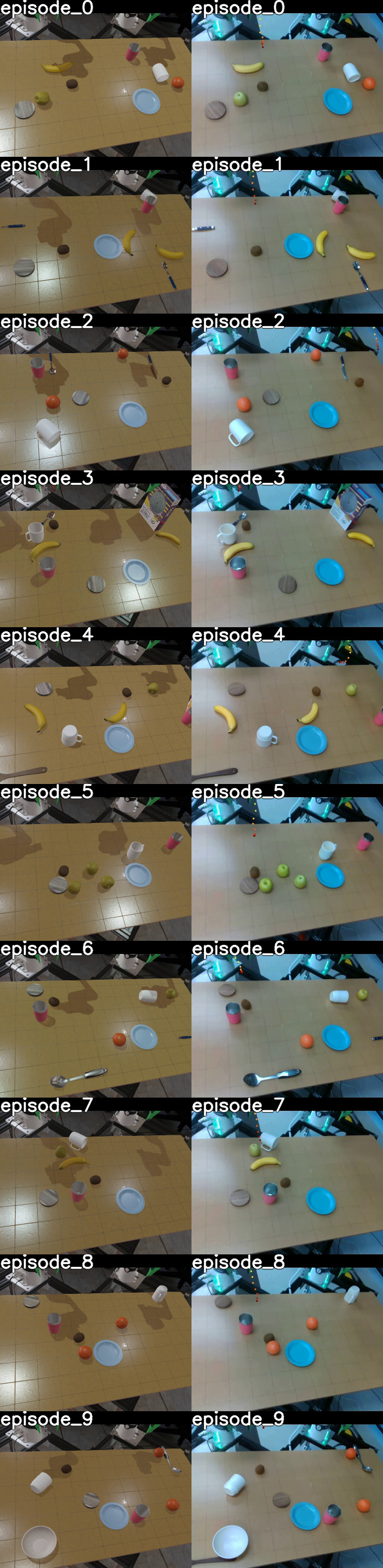}
\caption{Initial Conditions}
\end{subfigure}
\begin{subfigure}[b]{0.23\linewidth}
\includegraphics[width=\linewidth]{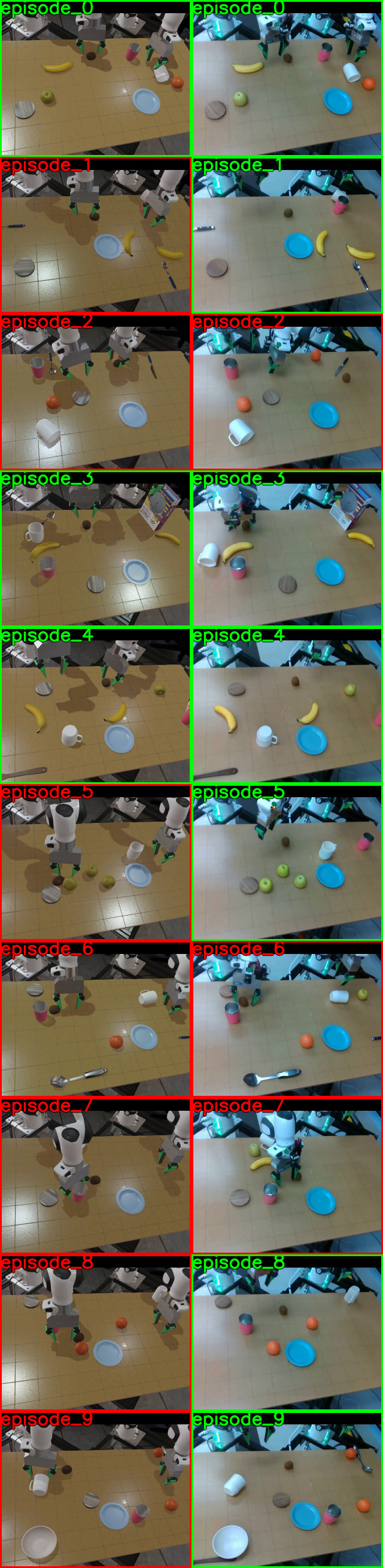}
\caption{Single Task}
\end{subfigure}
\begin{subfigure}[b]{0.23\linewidth}
\includegraphics[width=\linewidth]{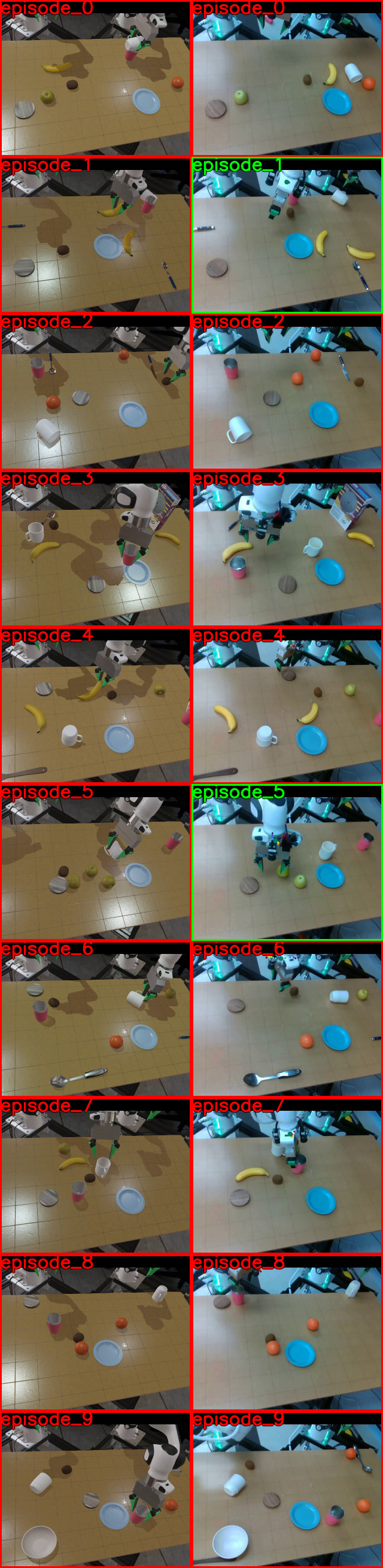}
\caption{Pretrained LBM}
\end{subfigure}
\begin{subfigure}[b]{0.23\linewidth}
\includegraphics[width=\linewidth]{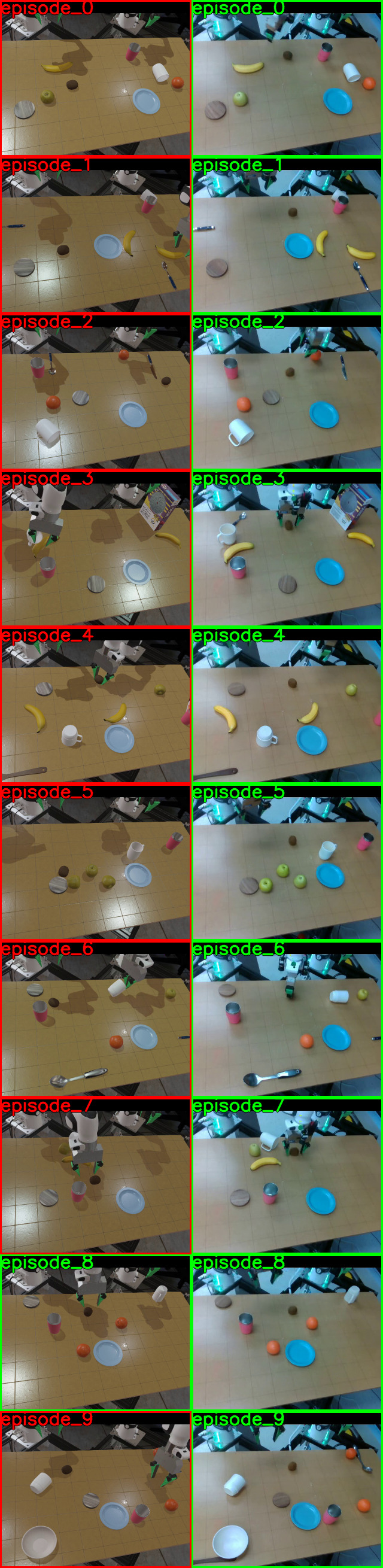}
\caption{Finetuned LBM}
\end{subfigure}
\caption{\small\textbf{Tiled terminal frames of rollouts from different policies performing \textit{PutKiwiInCenterOfTable} in simulation (left) and on hardware (right) under nominal conditions.} For single-task baselines, the policy in simulation is worse at grasping the kiwi compared to on hardware. The pretrained LBM in both simulation and real-world often performs the wrong task (interacting with the task irrelevant objects). The finetuned LBM in simulation suffers more from delays in initiating motion, and performs the wrong task more often than the real counterpart. The green outline around a rollout indicates success, while the red outline indicates failure.}
\label{fig:supp_qual_analysis_sim_and_real_kiwi}
}

\adaptivefigure{
\begin{subfigure}[b]{0.23\linewidth}
\includegraphics[width=\linewidth]{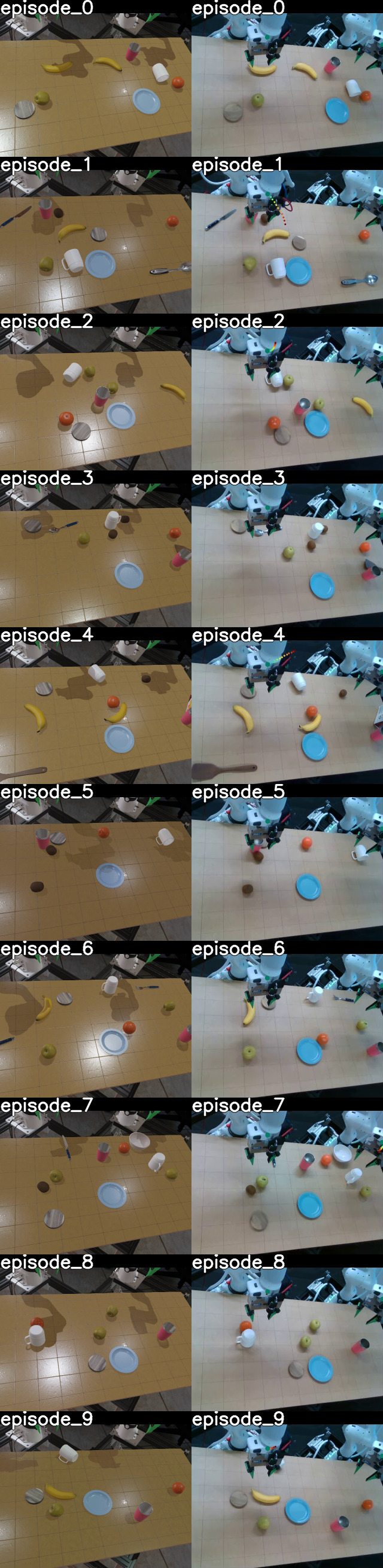}
\caption{Initial Conditions}
\end{subfigure}
\begin{subfigure}[b]{0.23\linewidth}
\includegraphics[width=\linewidth]{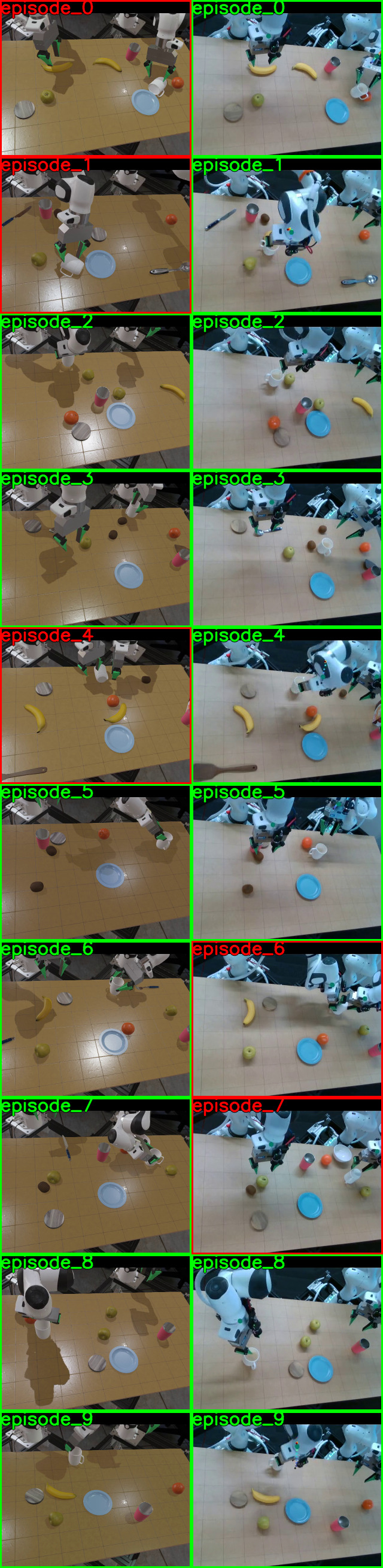}
\caption{Single Task}
\end{subfigure}
\begin{subfigure}[b]{0.23\linewidth}
\includegraphics[width=\linewidth]{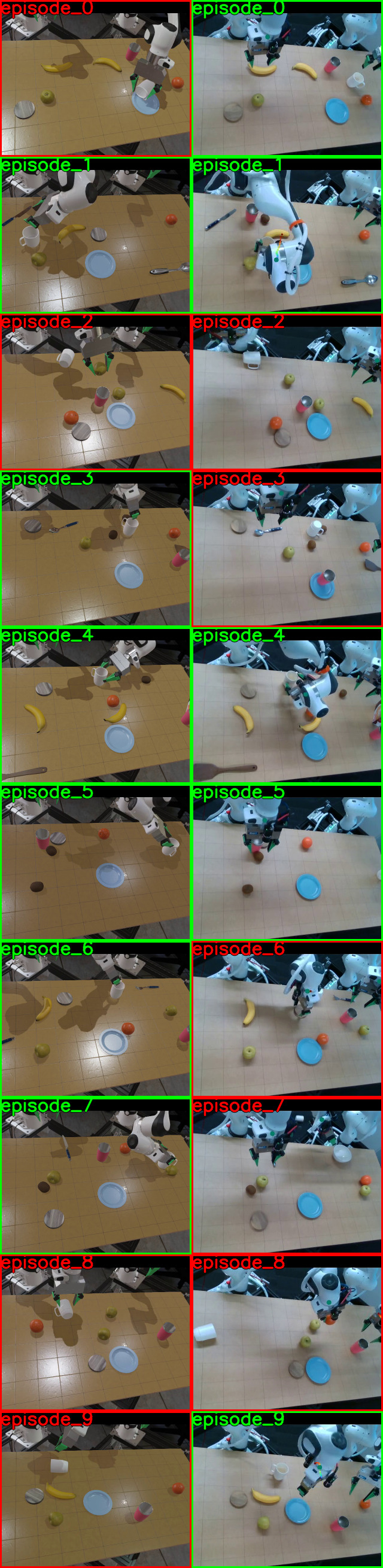}
\caption{Pretrained LBM}
\end{subfigure}
\begin{subfigure}[b]{0.23\linewidth}
\includegraphics[width=\linewidth]{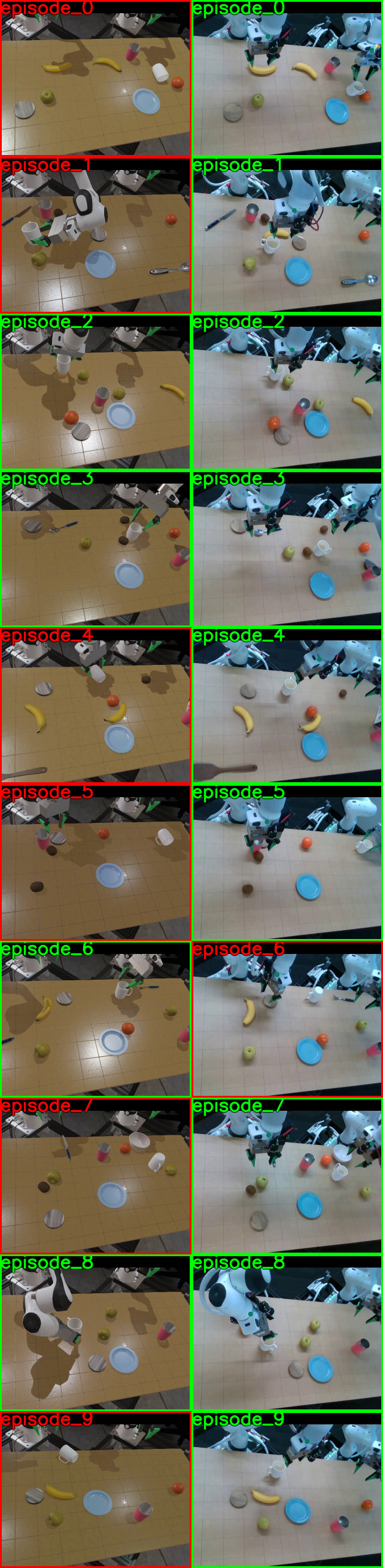}
\caption{Finetuned LBM}
\end{subfigure}
\caption{\small\textbf{Tiled terminal frames of rollouts from different policies performing \textit{TurnMugRightsideUp} in simulation (left) and on hardware (right) under nominal conditions.} The real-world single-task baseline much better than the single-task in simulation due to the more generous timeout protocol for real-world eval (see Section~\ref{subsec:supplemental_breakfast_sceario} for more details). The pretrained LBM performs roughly the same in simulation and on hardware. The finetuned LBM performs much worse in simulation due to delays in initiating motion resulting in timeouts. The green outline around a rollout indicates success, while the red outline indicates failure.}
\label{fig:supp_qual_analysis_sim_and_real_turnmug}
}

%% file: sections/21_appendix_violins.tex
\section{Bayesian analysis for task completion}
\label{sec:supp_bayesian}
\begin{figure}[htbp]
    \centering
    \includegraphics[width=\linewidth]{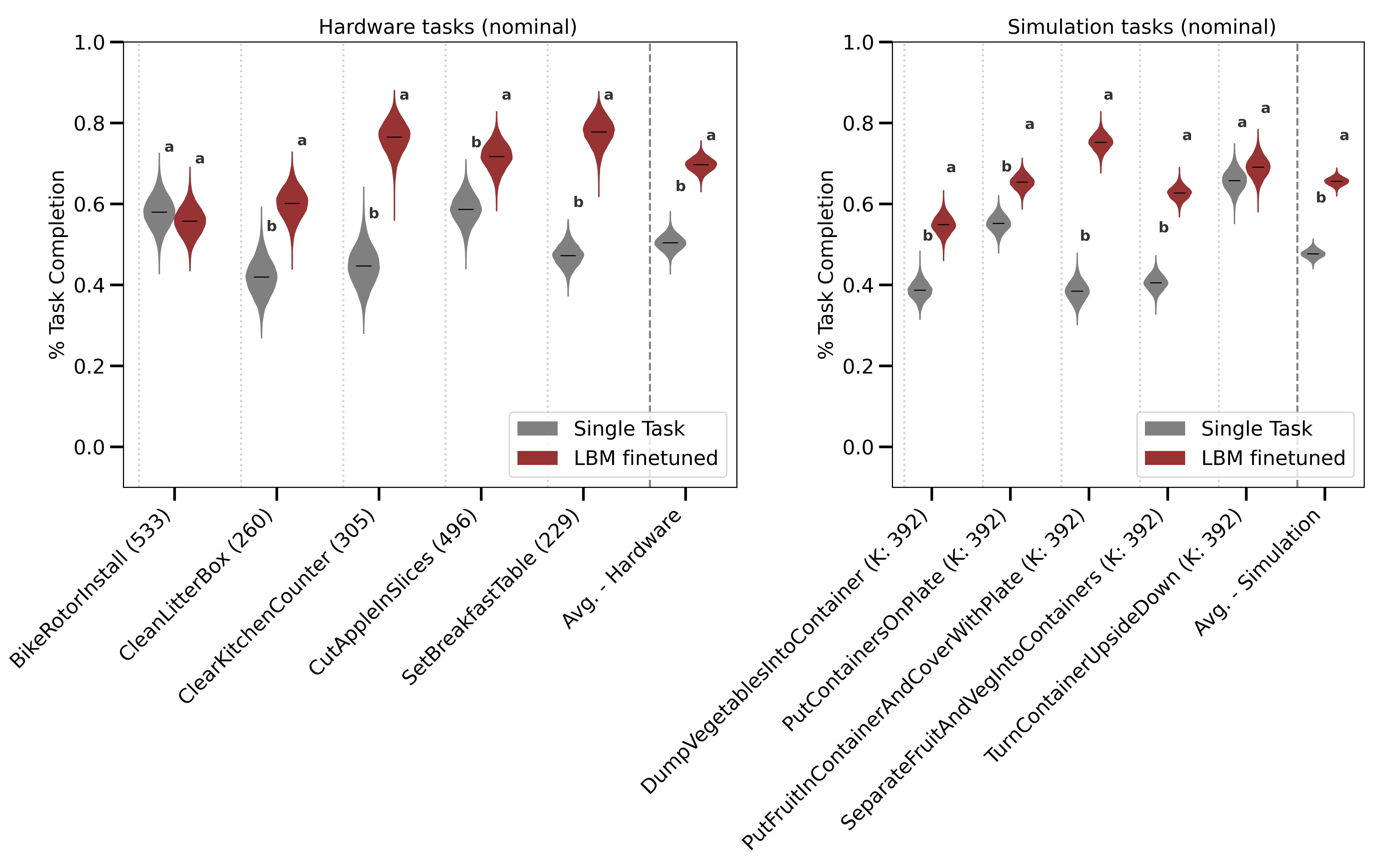}
    \caption{\small\textbf{LBM performance on ``unseen" tasks in hardware and in simulation evaluated under nominal conditions.} We compare the single-task baseline, with LBMs after finetuning. The violin plots represent the Bayesian posterior of the mean Task Completion under a uniform Dirichlet prior. We use statistical hypothesis tests over the mean TC for the CLD letters shown in these plot. These results complement the entire TC data distribution shown in Fig.~\ref{fig:unseen_tasks_sim_and_real}.}
    \label{fig:individual_task_progress_no_ds_violin}
\end{figure}

\begin{figure}[htbp]
    \centering
    \includegraphics[width=\linewidth]{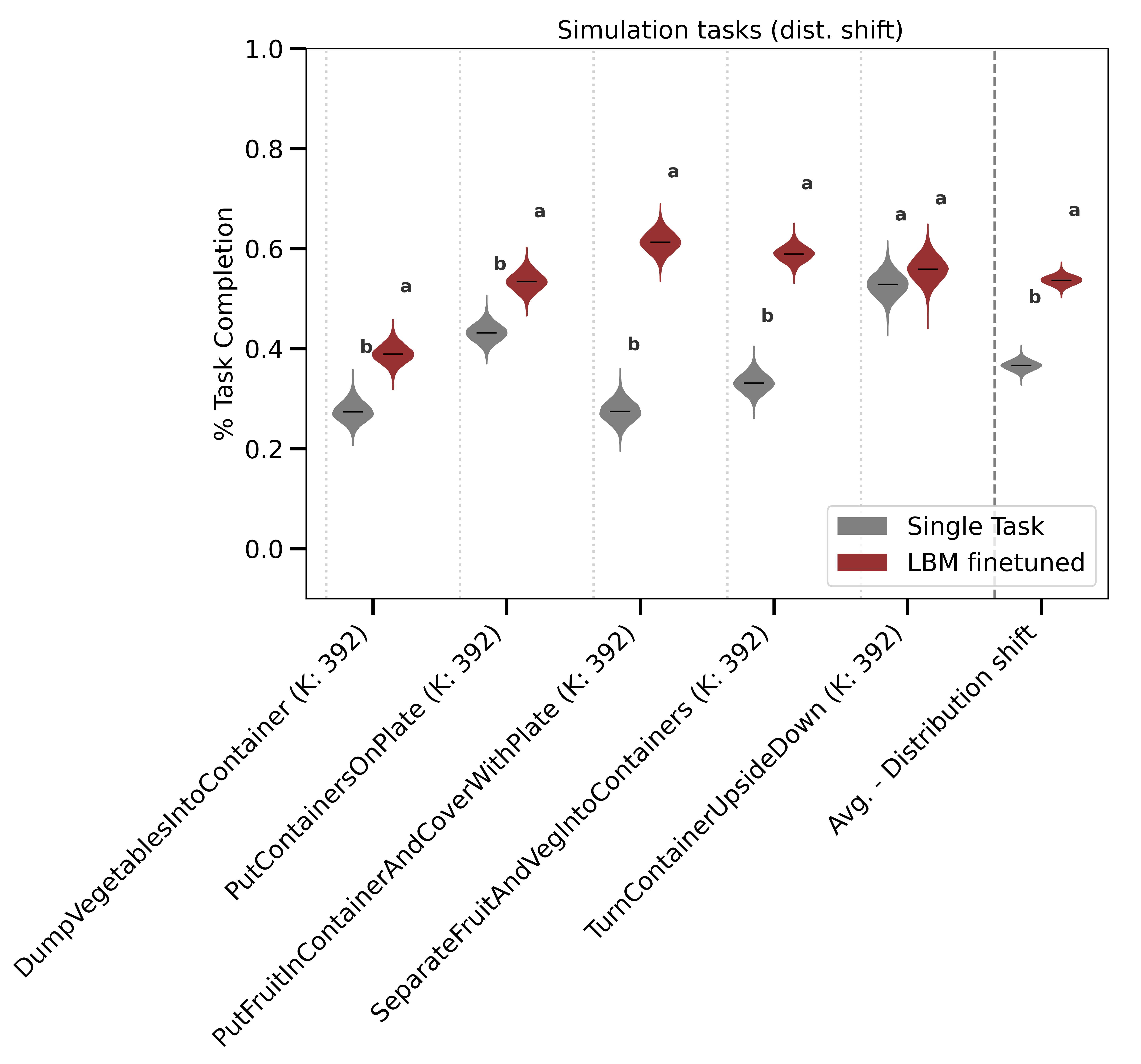}
    \caption{\small\textbf{LBM performance on ``unseen" tasks in simulation evaluated under distribution shift.} We compare the single-task baseline, with LBMs after finetuning. The violin plots represent the Bayesian posterior of the mean Task Completion under a uniform Dirichlet prior. We use statistical hypothesis tests over the mean TC for the CLD letters shown in these plot. These results complement the entire TC data distribution shown in Fig.~\ref{fig:unseen_tasks_sim_and_real_ds}.}
    \label{fig:individual_task_progress_with_ds_violin}
\end{figure}

In Sec.~\ref{subsec:LBMs_unseen_tasks}, Figures~\ref{fig:unseen_tasks_sim_and_real} and~\ref{fig:unseen_tasks_sim_and_real_ds}, for task completion, we depict the full data distribution, the mean and whether the policies are statistically distinct. Figures~\ref{fig:individual_task_progress_no_ds_violin} and~\ref{fig:individual_task_progress_with_ds_violin} correspond to the same data, only here we depict the uncertainty in the estimate of the true mean of the distribution, used to evaluate whether the policies are statistically distinct.

%% file: sections/18_appendix_pauses.tex
\section{Effects of dataset filtering}
\label{sec:supp_pauses}

In this section, we provide additional details regarding the effects of low-motion frames in the training data on policy performance in simulation, as discussed in Section~\ref{subsec:pretraining_dataset_filtering}. In particular, we experiment with filtering low-motion data at the beginning of each demonstration episode.  The filtering process is: for each demonstration, filter out all data points prior to the first timestep that the motion threshold is satisfied. This results in 3.3\% \textbf{TRI-Ramen-Real} and 11.7\% \textbf{TRI-Ramen-Sim} data filtered out. 

In the following we refer to LBMs trained with filtered data as \textit{filtered-pretrained} LBMs, and contrast them with \textit{unfiltered-pretrained} LBMs. As mentioned in Section~\ref{subsec:pretraining_dataset_filtering}, we examine the effects of filtering only on the pretraining phase, and finetune LBMs and train single-task baselines using the filtered data.

We observe that the \textit{filtered-pretrained} LBM  very quickly commits to a task, whereas the \textit{unfiltered-pretrained} LBM would often take a long time to initiate any motion. However, we encountered an unexpected effect where the \textit{filtered-pretrained} LBM would commit to a different task more often than before, i.e., the policy started suffering from language steerability issues. This significantly reduced performance of the pretrained LBM as shown in Fig.~\ref{fig:pauses_no_ds}, where single-task is statistically better than \textit{unfiltered-pretrained} LBM in 3/16 tasks and 6/16 tasks for \textit{filtered-pretrained} LBM, and finetuned LBM is statistically better than \textit{unfiltered-pretrained} LBM in 5/16 tasks and 9/16 tasks for \textit{filtered-pretrained} LBM.

\begin{figure}
\begin{subfigure}[b]{0.49\linewidth}
\includegraphics[width=\linewidth]{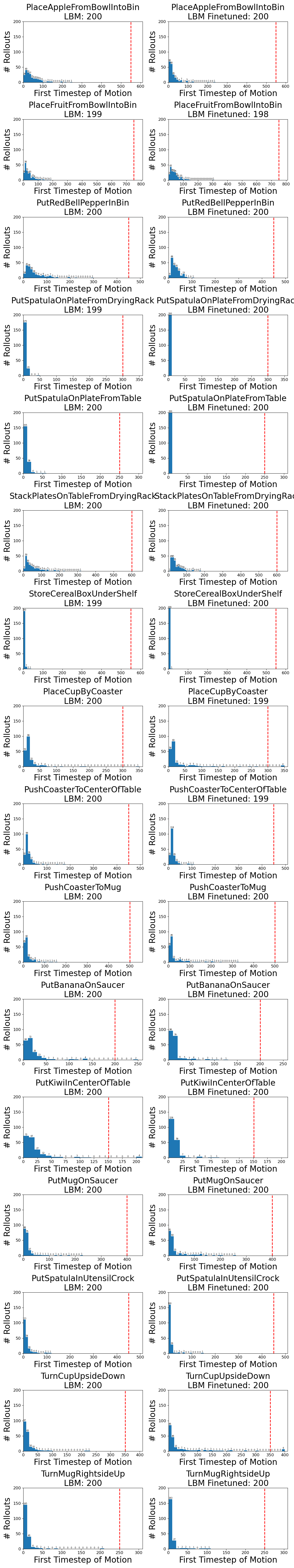}
\caption{\scriptsize Low-motion data filtered for all policies} 
\end{subfigure}
\begin{subfigure}[b]{0.49\linewidth}
\includegraphics[width=\linewidth]{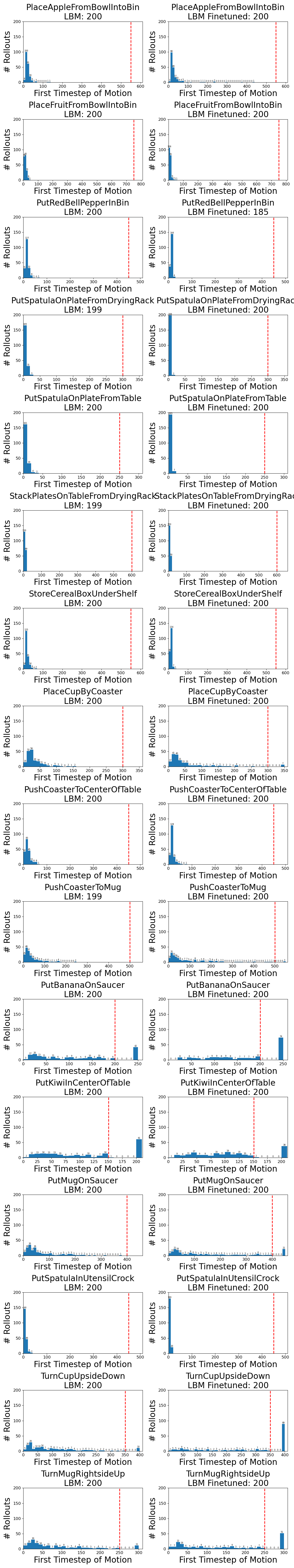}

\caption{\scriptsize Low-motion data filtered only for finetuned LBM and single-task}
\end{subfigure}
\caption{\scriptsize\textbf{Histogram of robot's first time of motion} (defined in Section~\ref{subsec:pretraining_dataset_filtering}) for 16 simulation ``seen" tasks under nominal conditions. Red lines indicate timeout. Each row corresponds to one task. Each subplot has two columns that corresponds to finetuned and pretrained LBMs respectively. Note that all finetuning are trained without low-motion data. After filtering low-motion data from pretraining data, both pretrained and subsequently finetuned LBMs initiate motions much faster.}
\label{fig:pauses_timeouts_new_vs_old_pretrain}
\end{figure}

We further introspected the effect of starting finetuning from either filtered- or unfiltered- pretrained LBMs. We observed that \textit{filtered-pretrained and finetuned} LBMs quickly initiate motions, whereas the \textit{unfiltered-pretrained and finetuned} LBMs can take excessively long to start any motion. Time-to-motion plots for LBMs are shown in Figure~\ref{fig:pauses_timeouts_new_vs_old_pretrain}. The overall task performance for the \textit{filtered-pretrained and finetuned} LBMs are similar to the \textit{unfiltered-pretrained and finetuned models}. But for specific tasks where the \textit{unfiltered-pretrained and finetuned} LBMs take excessively long to initiate any motion, the \textit{filtered-pretrained and finetuned} LBMs make significant improvements. We provide additional per-task details in Sections~\ref{subsec:supp_qual_analysis_put_kiwi_banana} and~\ref{subsec:supp_qual_analysis_put_turncup}.

Our hypothesis for why filtering low-motion data from the pretraining data caused more severe language steerability issues for our pretrained LBMs  is that low-motion data ends up  increasing the importance of the language conditioning during training. These data all concentrate at the beginning of each rollout, where scenes are maximally visually ambiguous (by benchmark design) and the policy needs to learn to pay close attention to language to perform the correct tasks. See Figure~\ref{fig:scenario4_ICs_nominal_vs_dist_shift} for an example where the starting conditions are the same for five different tasks, and the robot must learn to disambiguate based on language. Another interesting observation is that we have not observed pretrained or finetuned LBMs having difficulties initiating any motions on hardware regardless of filtering. This is likely due to the percentage of low-motion data being much smaller in hardware than in simulation.

\adaptivefigure{
\centering

\hfill
\begin{subfigure}[t]{\linewidth}
\includegraphics[height=8.05cm]{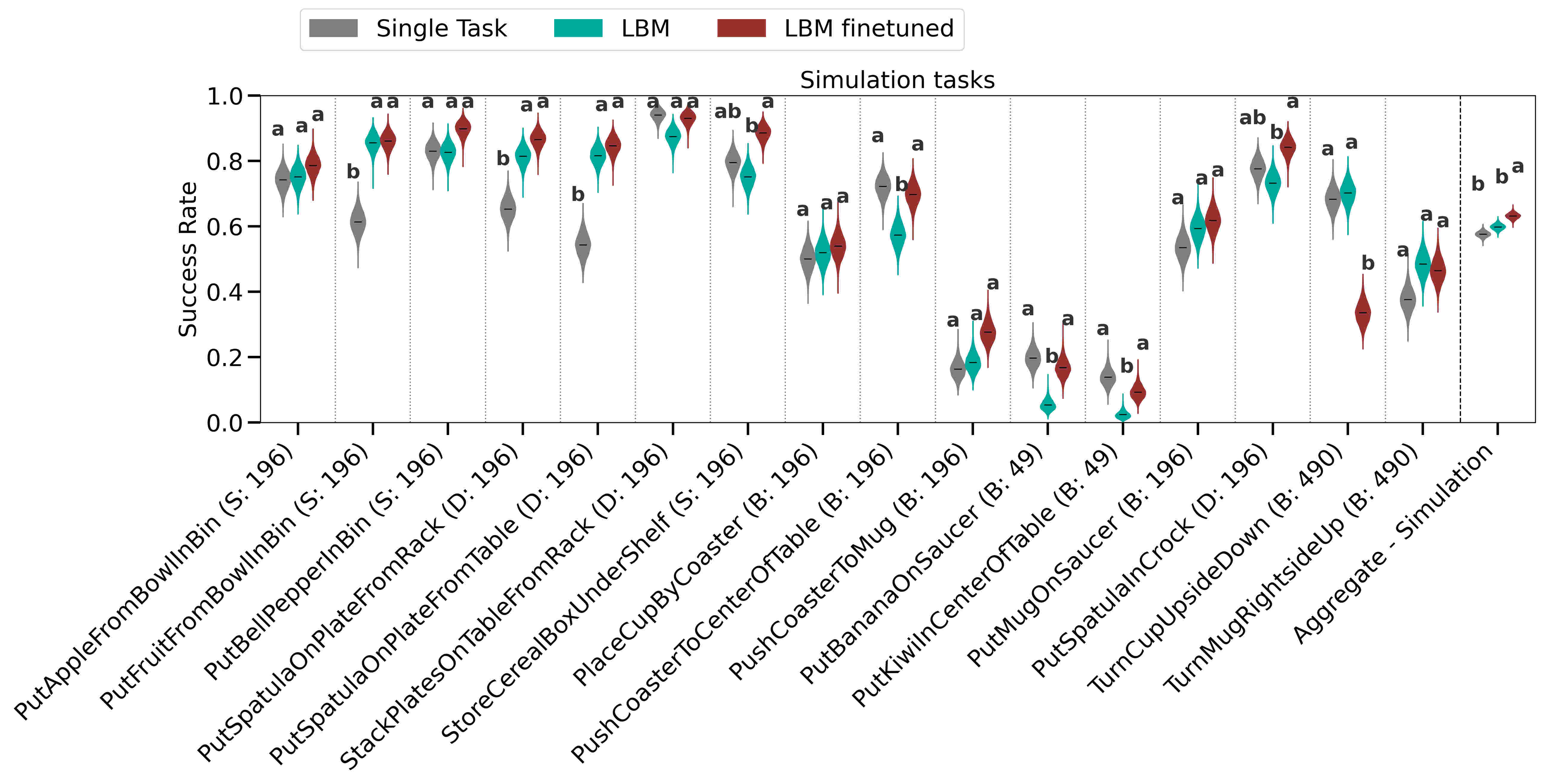}
\caption{LBM Pretrained using unfiltered dataset, while single-task and finetuned LBM use the filtered dataset. }
\label{subfig:pauses_pt_filtered_no_ds}
\end{subfigure}

\vspace{0.50em}

\begin{subfigure}[t]{\linewidth}
\includegraphics[height=8.05cm]{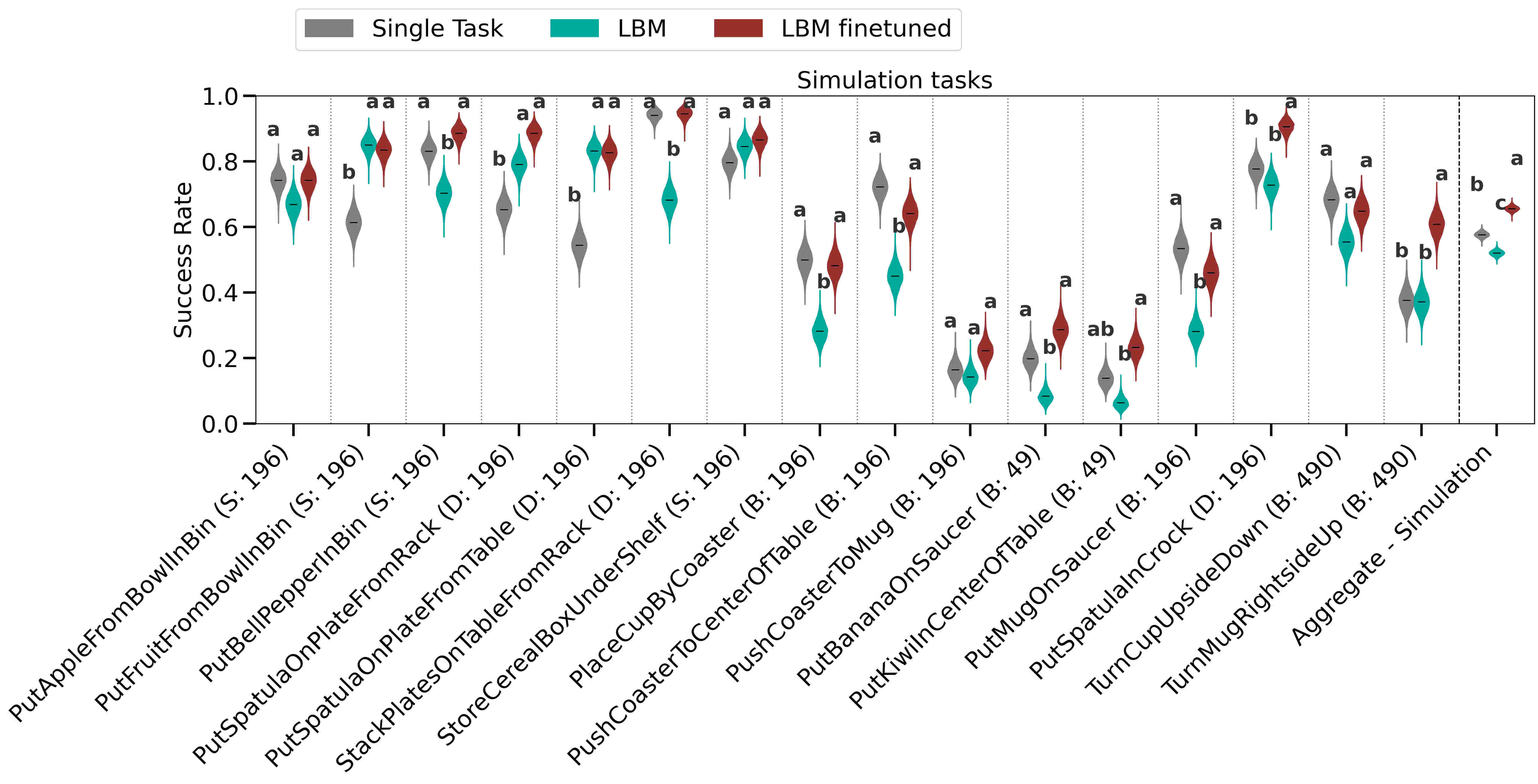}
\caption{Single-task, pretrained LBM and finetuned LBM all use the filtered dataset.}
\label{subfig:pauses_all_filtered_no_ds}
\end{subfigure}
\hfill
\caption{\small Analyzing the effect of low-motion data in the pretraining dataset. Finetuning and single-task baseline training were done using the filtered dataset. The evaluations are done under nominal conditions on simulation tasks which are seen during training.}
\label{fig:pauses_no_ds}
}

%% file: sections/14_appendix_dataset_details.tex
\section{Additional pretraining dataset details}
\label{sec:supp_oxe_dataset_details}

Table~\ref{tab:supp_batch_balance_weights} shows detailed information regarding the batch balancing weights used to create the LBM pretraining dataset. This complements Sec.~\ref{subsec:data}. 

\begin{table}
\centering
\scriptsize
\begin{tabular}{|l|r|}
\hline
Data source name  & Weights  \\
\hline
lbm\_real & 0.5 \\
lbm\_sim & 0.25 \\
lbm\_umi & 0.05 \\
\hline
bc\_z & 0.03 \\
berkeley\_autolab\_ur5 & 0.007 \\
bridge\_data\_v2 & 0.02 \\
droid & 0.05 \\
fractal20220817\_data & 0.03 \\
furniture\_bench\_dataset\_converted\_externally\_to\_rlds & 0.02 \\
jaco\_play & 0.005 \\
language\_table & 0.015 \\
nyu\_franka\_play\_dataset\_converted\_externally\_to\_rlds & 0.05 \\
stanford\_hydra\_dataset\_converted\_externally\_to\_rlds & 0.01 \\
utokyo\_xarm\_pick\_and\_place\_converted\_externally\_to\_rlds & 0.003 \\
viola & 0.005 \\
\hline
Total weight & 1.045 \\
\hline

\end{tabular}
\caption{\small Unnormalized batch balance weights used to train LBM. Weights are normalized to sum to one during training. }
\label{tab:supp_batch_balance_weights}
\end{table}

%% file: sections/23_appendix_stage5_bugfix.tex
\section{Data normalization experiment }
\label{sec:supp_stage5_bug_comparison}

As described in Section~\ref{subsec:normalization}, we discovered that some of the datagrams in the training data were not normalized correctly. We ran an additional evaluation on ``seen" simulation tasks, both in nominal conditions and under distribution shift, to assess the impact of the normalization error. Figure~\ref{fig:stage5_bug_fix_comparison} show how the pretrained LBM with the corrected parameters performs with respect to the rest of the policies. We can see that for these tasks, the error did not significantly affect the performance of the pretrained LBM under nominal conditions; however, for distribution shift, the pretrained LBM without the error outperforms the pretrained LBM with the error on 4/16 tasks and in the aggregate.

\adaptivefigure{
\centering

\hfill
\begin{subfigure}[t]{\linewidth}
\includegraphics[height=8.05cm]{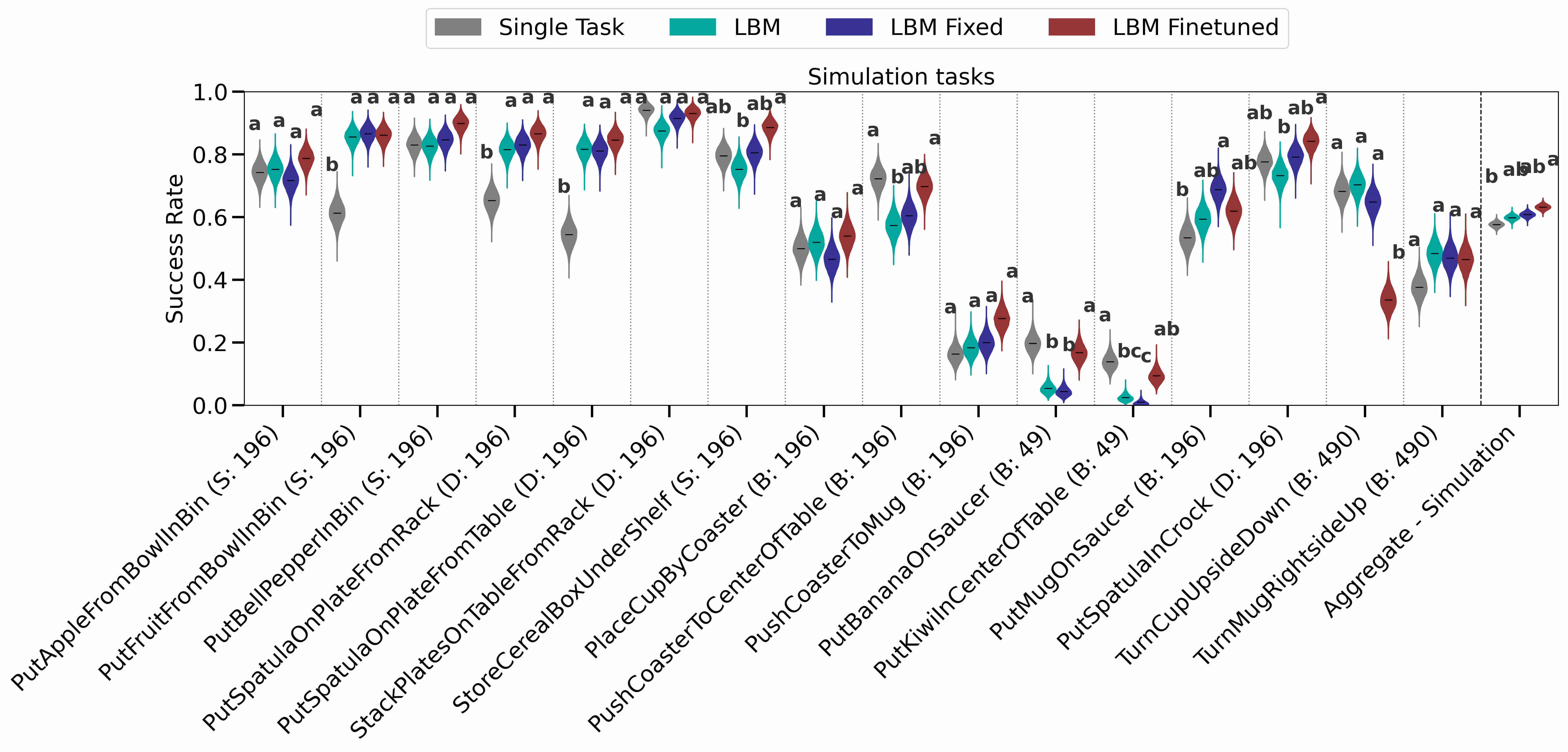}
\caption{LBM vs LBM Fixed (corrected normalization parameters) evaluated on ``seen" tasks under nominal conditions.}
\label{subfig:stage5_bugfix_comp_no_ds}
\end{subfigure}

\vspace{0.50em}

\begin{subfigure}[t]{\linewidth}
\includegraphics[height=8.05cm]{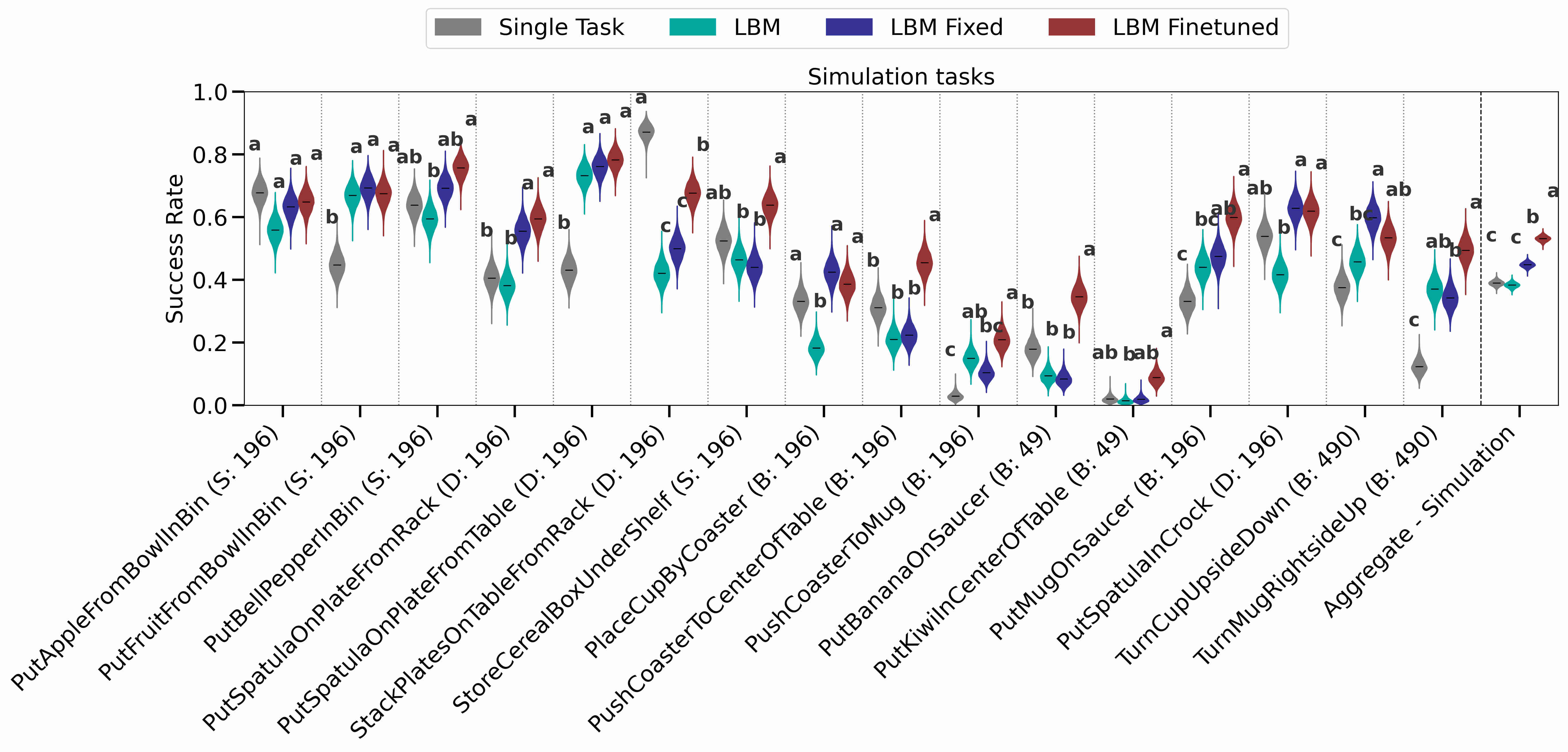}
\caption{LBM vs LBM  Fixed (corrected normalization parameters) evaluated on ``seen" tasks under distribution-shift conditions.}
\label{subfig:stage5_bugfix_comp_ds}
\end{subfigure}
\hfill
\caption{\small Comparing pretrained LBMs with and without the normalizer error on ``seen" tasks in simulation. The top row is for nominal conditions and the bottom row for distribution shift. This figure adds the corrected pretrained LBM to the results in Figure~\ref{fig:seen_tasks_sim_and_real} for easier comparison. 
}
\label{fig:stage5_bug_fix_comparison}
}

%% file: lbm_tpami_format.bbl
\begin{thebibliography}{10}
\providecommand{\url}[1]{#1}
\csname url@samestyle\endcsname
\providecommand{\newblock}{\relax}
\providecommand{\bibinfo}[2]{#2}
\providecommand{\BIBentrySTDinterwordspacing}{\spaceskip=0pt\relax}
\providecommand{\BIBentryALTinterwordstretchfactor}{4}
\providecommand{\BIBentryALTinterwordspacing}{\spaceskip=\fontdimen2\font plus
\BIBentryALTinterwordstretchfactor\fontdimen3\font minus \fontdimen4\font\relax}
\providecommand{\BIBforeignlanguage}[2]{{%
\expandafter\ifx\csname l@#1\endcsname\relax
\typeout{** WARNING: IEEEtran.bst: No hyphenation pattern has been}%
\typeout{** loaded for the language `#1'. Using the pattern for}%
\typeout{** the default language instead.}%
\else
\language=\csname l@#1\endcsname
\fi
#2}}
\providecommand{\BIBdecl}{\relax}
\BIBdecl

\bibitem{chi2024diffusionpolicy}
C.~Chi, Z.~Xu, S.~Feng, E.~Cousineau, Y.~Du, B.~Burchfiel, R.~Tedrake, and S.~Song, ``Diffusion policy: Visuomotor policy learning via action diffusion,'' \emph{The International Journal of Robotics Research}, 2024.

\bibitem{zhao2023learningfinegrainedbimanualmanipulation}
\BIBentryALTinterwordspacing
T.~Z. Zhao, V.~Kumar, S.~Levine, and C.~Finn, ``Learning fine-grained bimanual manipulation with low-cost hardware,'' in \emph{Robotics: Science and Systems XIX}.\hskip 1em plus 0.5em minus 0.4em\relax Robotics: Science and Systems Foundation, 2023. [Online]. Available: \url{https://roboticsproceedings.org/rss19/p078.pdf}
\BIBentrySTDinterwordspacing

\bibitem{zhao2024aloha}
T.~Z. Zhao, J.~Tompson, D.~Driess, P.~Florence, K.~Ghasemipour, C.~Finn, and A.~Wahid, ``Aloha unleashed: A simple recipe for robot dexterity,'' in \emph{8th Annual Conference on Robot Learning}, 2024.

\bibitem{octomodelteam2024octoopensourcegeneralistrobot}
\BIBentryALTinterwordspacing
O.~M. Team, D.~Ghosh, H.~Walke, K.~Pertsch, K.~Black, O.~Mees, S.~Dasari, J.~Hejna, T.~Kreiman, C.~Xu, J.~Luo, Y.~L. Tan, L.~Y. Chen, P.~Sanketi, Q.~Vuong, T.~Xiao, D.~Sadigh, C.~Finn, and S.~Levine, ``Octo: An open-source generalist robot policy,'' 2024. [Online]. Available: \url{https://arxiv.org/abs/2405.12213}
\BIBentrySTDinterwordspacing

\bibitem{black2024pi0visionlanguageactionflowmodel}
\BIBentryALTinterwordspacing
K.~Black, N.~Brown, D.~Driess, A.~Esmail, M.~Equi, C.~Finn, N.~Fusai, L.~Groom, K.~Hausman, B.~Ichter, S.~Jakubczak, T.~Jones, L.~Ke, S.~Levine, A.~Li-Bell, M.~Mothukuri, S.~Nair, K.~Pertsch, L.~X. Shi, J.~Tanner, Q.~Vuong, A.~Walling, H.~Wang, and U.~Zhilinsky, ``$\pi_0$: A vision-language-action flow model for general robot control,'' 2024. [Online]. Available: \url{https://arxiv.org/abs/2410.24164}
\BIBentrySTDinterwordspacing

\bibitem{nvidia2025gr00tn1openfoundation}
\BIBentryALTinterwordspacing
NVIDIA, :, J.~Bjorck, F.~Castañeda, N.~Cherniadev, X.~Da, R.~Ding, L.~J. Fan, Y.~Fang, D.~Fox, F.~Hu, S.~Huang, J.~Jang, Z.~Jiang, J.~Kautz, K.~Kundalia, L.~Lao, Z.~Li, Z.~Lin, K.~Lin, G.~Liu, E.~Llontop, L.~Magne, A.~Mandlekar, A.~Narayan, S.~Nasiriany, S.~Reed, Y.~L. Tan, G.~Wang, Z.~Wang, J.~Wang, Q.~Wang, J.~Xiang, Y.~Xie, Y.~Xu, Z.~Xu, S.~Ye, Z.~Yu, A.~Zhang, H.~Zhang, Y.~Zhao, R.~Zheng, and Y.~Zhu, ``Gr00t n1: An open foundation model for generalist humanoid robots,'' 2025. [Online]. Available: \url{https://arxiv.org/abs/2503.14734}
\BIBentrySTDinterwordspacing

\bibitem{geminiroboticsteam2025geminiroboticsbringingai}
\BIBentryALTinterwordspacing
G.~R. Team, S.~Abeyruwan, J.~Ainslie, J.-B. Alayrac, M.~G. Arenas, T.~Armstrong, A.~Balakrishna, R.~Baruch, M.~Bauza, M.~Blokzijl, S.~Bohez, K.~Bousmalis, A.~Brohan, T.~Buschmann, A.~Byravan, S.~Cabi, K.~Caluwaerts, F.~Casarini, O.~Chang, J.~E. Chen, X.~Chen, H.-T.~L. Chiang, K.~Choromanski, D.~D'Ambrosio, S.~Dasari, T.~Davchev, C.~Devin, N.~D. Palo, T.~Ding, A.~Dostmohamed, D.~Driess, Y.~Du, D.~Dwibedi, M.~Elabd, C.~Fantacci, C.~Fong, E.~Frey, C.~Fu, M.~Giustina, K.~Gopalakrishnan, L.~Graesser, L.~Hasenclever, N.~Heess, B.~Hernaez, A.~Herzog, R.~A. Hofer, J.~Humplik, A.~Iscen, M.~G. Jacob, D.~Jain, R.~Julian, D.~Kalashnikov, M.~E. Karagozler, S.~Karp, C.~Kew, J.~Kirkland, S.~Kirmani, Y.~Kuang, T.~Lampe, A.~Laurens, I.~Leal, A.~X. Lee, T.-W.~E. Lee, J.~Liang, Y.~Lin, S.~Maddineni, A.~Majumdar, A.~H. Michaely, R.~Moreno, M.~Neunert, F.~Nori, C.~Parada, E.~Parisotto, P.~Pastor, A.~Pooley, K.~Rao, K.~Reymann, D.~Sadigh, S.~Saliceti, P.~Sanketi, P.~Sermanet, D.~Shah, M.~Sharma, K.~Shea, C.~Shu, V.~Sindhwani,
  S.~Singh, R.~Soricut, J.~T. Springenberg, R.~Sterneck, R.~Surdulescu, J.~Tan, J.~Tompson, V.~Vanhoucke, J.~Varley, G.~Vesom, G.~Vezzani, O.~Vinyals, A.~Wahid, S.~Welker, P.~Wohlhart, F.~Xia, T.~Xiao, A.~Xie, J.~Xie, P.~Xu, S.~Xu, Y.~Xu, Z.~Xu, Y.~Yang, R.~Yao, S.~Yaroshenko, W.~Yu, W.~Yuan, J.~Zhang, T.~Zhang, A.~Zhou, and Y.~Zhou, ``Gemini robotics: Bringing ai into the physical world,'' 2025. [Online]. Available: \url{https://arxiv.org/abs/2503.20020}
\BIBentrySTDinterwordspacing

\bibitem{wang2024scaling}
L.~Wang, X.~Chen, J.~Zhao, and K.~He, ``Scaling proprioceptive-visual learning with heterogeneous pre-trained transformers,'' \emph{Advances in neural information processing systems}, vol.~37, pp. 124\,420--124\,450, 2024.

\bibitem{liu_rdt-1b_2025}
\BIBentryALTinterwordspacing
S.~Liu, L.~Wu, B.~Li, H.~Tan, H.~Chen, Z.~Wang, K.~Xu, H.~Su, and J.~Zhu, ``{RDT}-{1B}: a {Diffusion} {Foundation} {Model} for {Bimanual} {Manipulation},'' Mar. 2025, arXiv:2410.07864 [cs]. [Online]. Available: \url{http://arxiv.org/abs/2410.07864}
\BIBentrySTDinterwordspacing

\bibitem{kirillov2023segment}
A.~Kirillov, E.~Mintun, N.~Ravi, H.~Mao, C.~Rolland, L.~Gustafson, T.~Xiao, S.~Whitehead, A.~C. Berg, W.-Y. Lo \emph{et~al.}, ``Segment anything,'' in \emph{2023 IEEE/CVF International Conference on Computer Vision (ICCV)}.\hskip 1em plus 0.5em minus 0.4em\relax IEEE, 2023, pp. 3992--4003.

\bibitem{radford2021learningtransferablevisualmodels}
A.~Radford, J.~W. Kim, C.~Hallacy, A.~Ramesh, G.~Goh, S.~Agarwal, G.~Sastry, A.~Askell, P.~Mishkin, J.~Clark \emph{et~al.}, ``Learning transferable visual models from natural language supervision,'' in \emph{International conference on machine learning}.\hskip 1em plus 0.5em minus 0.4em\relax PmLR, 2021, pp. 8748--8763.

\bibitem{zhai2023sigmoidlosslanguageimage}
X.~Zhai, B.~Mustafa, A.~Kolesnikov, and L.~Beyer, ``Sigmoid loss for language image pre-training,'' in \emph{Proceedings of the IEEE/CVF international conference on computer vision}, 2023, pp. 11\,975--11\,986.

\bibitem{oquab2024dinov2learningrobustvisual}
M.~Oquab, T.~Darcet, T.~Moutakanni, H.~Vo, M.~Szafraniec, V.~Khalidov, P.~Fernandez, D.~Haziza, F.~Massa, A.~El-Nouby \emph{et~al.}, ``Dinov2: Learning robust visual features without supervision,'' \emph{Transactions on Machine Learning Research Journal}, pp. 1--31, 2024.

\bibitem{openai2024gpt4technicalreport}
\BIBentryALTinterwordspacing
OpenAI, J.~Achiam, S.~Adler, S.~Agarwal, L.~Ahmad, I.~Akkaya, F.~L. Aleman, D.~Almeida, J.~Altenschmidt, S.~Altman, S.~Anadkat, R.~Avila, I.~Babuschkin, S.~Balaji, V.~Balcom, P.~Baltescu, H.~Bao, M.~Bavarian, J.~Belgum, I.~Bello, J.~Berdine, G.~Bernadett-Shapiro, C.~Berner, L.~Bogdonoff, O.~Boiko, M.~Boyd, A.-L. Brakman, G.~Brockman, T.~Brooks, M.~Brundage, K.~Button, T.~Cai, R.~Campbell, A.~Cann, B.~Carey, C.~Carlson, R.~Carmichael, B.~Chan, C.~Chang, F.~Chantzis, D.~Chen, S.~Chen, R.~Chen, J.~Chen, M.~Chen, B.~Chess, C.~Cho, C.~Chu, H.~W. Chung, D.~Cummings, J.~Currier, Y.~Dai, C.~Decareaux, T.~Degry, N.~Deutsch, D.~Deville, A.~Dhar, D.~Dohan, S.~Dowling, S.~Dunning, A.~Ecoffet, A.~Eleti, T.~Eloundou, D.~Farhi, L.~Fedus, N.~Felix, S.~P. Fishman, J.~Forte, I.~Fulford, L.~Gao, E.~Georges, C.~Gibson, V.~Goel, T.~Gogineni, G.~Goh, R.~Gontijo-Lopes, J.~Gordon, M.~Grafstein, S.~Gray, R.~Greene, J.~Gross, S.~S. Gu, Y.~Guo, C.~Hallacy, J.~Han, J.~Harris, Y.~He, M.~Heaton, J.~Heidecke, C.~Hesse, A.~Hickey,
  W.~Hickey, P.~Hoeschele, B.~Houghton, K.~Hsu, S.~Hu, X.~Hu, J.~Huizinga, S.~Jain, S.~Jain, J.~Jang, A.~Jiang, R.~Jiang, H.~Jin, D.~Jin, S.~Jomoto, B.~Jonn, H.~Jun, T.~Kaftan, Łukasz Kaiser, A.~Kamali, I.~Kanitscheider, N.~S. Keskar, T.~Khan, L.~Kilpatrick, J.~W. Kim, C.~Kim, Y.~Kim, J.~H. Kirchner, J.~Kiros, M.~Knight, D.~Kokotajlo, Łukasz Kondraciuk, A.~Kondrich, A.~Konstantinidis, K.~Kosic, G.~Krueger, V.~Kuo, M.~Lampe, I.~Lan, T.~Lee, J.~Leike, J.~Leung, D.~Levy, C.~M. Li, R.~Lim, M.~Lin, S.~Lin, M.~Litwin, T.~Lopez, R.~Lowe, P.~Lue, A.~Makanju, K.~Malfacini, S.~Manning, T.~Markov, Y.~Markovski, B.~Martin, K.~Mayer, A.~Mayne, B.~McGrew, S.~M. McKinney, C.~McLeavey, P.~McMillan, J.~McNeil, D.~Medina, A.~Mehta, J.~Menick, L.~Metz, A.~Mishchenko, P.~Mishkin, V.~Monaco, E.~Morikawa, D.~Mossing, T.~Mu, M.~Murati, O.~Murk, D.~Mély, A.~Nair, R.~Nakano, R.~Nayak, A.~Neelakantan, R.~Ngo, H.~Noh, L.~Ouyang, C.~O'Keefe, J.~Pachocki, A.~Paino, J.~Palermo, A.~Pantuliano, G.~Parascandolo, J.~Parish, E.~Parparita,
  A.~Passos, M.~Pavlov, A.~Peng, A.~Perelman, F.~de~Avila Belbute~Peres, M.~Petrov, H.~P. de~Oliveira~Pinto, Michael, Pokorny, M.~Pokrass, V.~H. Pong, T.~Powell, A.~Power, B.~Power, E.~Proehl, R.~Puri, A.~Radford, J.~Rae, A.~Ramesh, C.~Raymond, F.~Real, K.~Rimbach, C.~Ross, B.~Rotsted, H.~Roussez, N.~Ryder, M.~Saltarelli, T.~Sanders, S.~Santurkar, G.~Sastry, H.~Schmidt, D.~Schnurr, J.~Schulman, D.~Selsam, K.~Sheppard, T.~Sherbakov, J.~Shieh, S.~Shoker, P.~Shyam, S.~Sidor, E.~Sigler, M.~Simens, J.~Sitkin, K.~Slama, I.~Sohl, B.~Sokolowsky, Y.~Song, N.~Staudacher, F.~P. Such, N.~Summers, I.~Sutskever, J.~Tang, N.~Tezak, M.~B. Thompson, P.~Tillet, A.~Tootoonchian, E.~Tseng, P.~Tuggle, N.~Turley, J.~Tworek, J.~F.~C. Uribe, A.~Vallone, A.~Vijayvergiya, C.~Voss, C.~Wainwright, J.~J. Wang, A.~Wang, B.~Wang, J.~Ward, J.~Wei, C.~Weinmann, A.~Welihinda, P.~Welinder, J.~Weng, L.~Weng, M.~Wiethoff, D.~Willner, C.~Winter, S.~Wolrich, H.~Wong, L.~Workman, S.~Wu, J.~Wu, M.~Wu, K.~Xiao, T.~Xu, S.~Yoo, K.~Yu, Q.~Yuan,
  W.~Zaremba, R.~Zellers, C.~Zhang, M.~Zhang, S.~Zhao, T.~Zheng, J.~Zhuang, W.~Zhuk, and B.~Zoph, ``Gpt-4 technical report,'' 2024. [Online]. Available: \url{https://arxiv.org/abs/2303.08774}
\BIBentrySTDinterwordspacing

\bibitem{touvron2023llama1}
H.~Touvron, T.~Lavril, G.~Izacard, X.~Martinet, M.-A. Lachaux, T.~Lacroix, B.~Rozi{\`e}re, N.~Goyal, E.~Hambro, F.~Azhar \emph{et~al.}, ``Llama: Open and efficient foundation language models,'' \emph{arXiv preprint arXiv:2302.13971}, 2023.

\bibitem{khazatsky2024droid}
A.~Khazatsky, K.~Pertsch, S.~Nair, A.~Balakrishna, S.~Dasari, S.~Karamcheti, S.~Nasiriany, M.~K. Srirama, L.~Y. Chen, K.~Ellis, P.~D. Fagan, J.~Hejna, M.~Itkina, M.~Lepert, Y.~J. Ma, P.~T. Miller, J.~Wu, S.~Belkhale, S.~Dass, H.~Ha, A.~Jain, A.~Lee, Y.~Lee, M.~Memmel, S.~Park, I.~Radosavovic, K.~Wang, A.~Zhan, K.~Black, C.~Chi, K.~B. Hatch, S.~Lin, J.~Lu, J.~Mercat, A.~Rehman, P.~R. Sanketi, A.~Sharma, C.~Simpson, Q.~Vuong, H.~R. Walke, B.~Wulfe, T.~Xiao, J.~H. Yang, A.~Yavary, T.~Z. Zhao, C.~Agia, R.~Baijal, M.~G. Castro, D.~Chen, Q.~Chen, T.~Chung, J.~Drake, E.~P. Foster, J.~Gao, V.~Guizilini, D.~A. Herrera, M.~Heo, K.~Hsu, J.~Hu, M.~Z. Irshad, D.~Jackson, C.~Le, Y.~Li, K.~Lin, R.~Lin, Z.~Ma, A.~Maddukuri, S.~Mirchandani, D.~Morton, T.~Nguyen, A.~O'Neill, R.~Scalise, D.~Seale, V.~Son, S.~Tian, E.~Tran, A.~E. Wang, Y.~Wu, A.~Xie, J.~Yang, P.~Yin, Y.~Zhang, O.~Bastani, G.~Berseth, J.~Bohg, K.~Goldberg, A.~Gupta, A.~Gupta, D.~Jayaraman, J.~J. Lim, J.~Malik, R.~Martín-Martín, S.~Ramamoorthy, D.~Sadigh,
  S.~Song, J.~Wu, M.~C. Yip, Y.~Zhu, T.~Kollar, S.~Levine, and C.~Finn, ``Droid: A large-scale in-the-wild robot manipulation dataset,'' 2024.

\bibitem{embodimentcollaboration2024openxembodimentroboticlearning}
\BIBentryALTinterwordspacing
E.~Collaboration, A.~O'Neill, A.~Rehman, A.~Gupta, A.~Maddukuri, A.~Gupta, A.~Padalkar, A.~Lee, A.~Pooley, A.~Gupta, A.~Mandlekar, A.~Jain, A.~Tung, A.~Bewley, A.~Herzog, A.~Irpan, A.~Khazatsky, A.~Rai, A.~Gupta, A.~Wang, A.~Kolobov, A.~Singh, A.~Garg, A.~Kembhavi, A.~Xie, A.~Brohan, A.~Raffin, A.~Sharma, A.~Yavary, A.~Jain, A.~Balakrishna, A.~Wahid, B.~Burgess-Limerick, B.~Kim, B.~Schölkopf, B.~Wulfe, B.~Ichter, C.~Lu, C.~Xu, C.~Le, C.~Finn, C.~Wang, C.~Xu, C.~Chi, C.~Huang, C.~Chan, C.~Agia, C.~Pan, C.~Fu, C.~Devin, D.~Xu, D.~Morton, D.~Driess, D.~Chen, D.~Pathak, D.~Shah, D.~Büchler, D.~Jayaraman, D.~Kalashnikov, D.~Sadigh, E.~Johns, E.~Foster, F.~Liu, F.~Ceola, F.~Xia, F.~Zhao, F.~V. Frujeri, F.~Stulp, G.~Zhou, G.~S. Sukhatme, G.~Salhotra, G.~Yan, G.~Feng, G.~Schiavi, G.~Berseth, G.~Kahn, G.~Yang, G.~Wang, H.~Su, H.-S. Fang, H.~Shi, H.~Bao, H.~B. Amor, H.~I. Christensen, H.~Furuta, H.~Bharadhwaj, H.~Walke, H.~Fang, H.~Ha, I.~Mordatch, I.~Radosavovic, I.~Leal, J.~Liang, J.~Abou-Chakra, J.~Kim, J.~Drake,
  J.~Peters, J.~Schneider, J.~Hsu, J.~Vakil, J.~Bohg, J.~Bingham, J.~Wu, J.~Gao, J.~Hu, J.~Wu, J.~Wu, J.~Sun, J.~Luo, J.~Gu, J.~Tan, J.~Oh, J.~Wu, J.~Lu, J.~Yang, J.~Malik, J.~Silvério, J.~Hejna, J.~Booher, J.~Tompson, J.~Yang, J.~Salvador, J.~J. Lim, J.~Han, K.~Wang, K.~Rao, K.~Pertsch, K.~Hausman, K.~Go, K.~Gopalakrishnan, K.~Goldberg, K.~Byrne, K.~Oslund, K.~Kawaharazuka, K.~Black, K.~Lin, K.~Zhang, K.~Ehsani, K.~Lekkala, K.~Ellis, K.~Rana, K.~Srinivasan, K.~Fang, K.~P. Singh, K.-H. Zeng, K.~Hatch, K.~Hsu, L.~Itti, L.~Y. Chen, L.~Pinto, L.~Fei-Fei, L.~Tan, L.~J. Fan, L.~Ott, L.~Lee, L.~Weihs, M.~Chen, M.~Lepert, M.~Memmel, M.~Tomizuka, M.~Itkina, M.~G. Castro, M.~Spero, M.~Du, M.~Ahn, M.~C. Yip, M.~Zhang, M.~Ding, M.~Heo, M.~K. Srirama, M.~Sharma, M.~J. Kim, N.~Kanazawa, N.~Hansen, N.~Heess, N.~J. Joshi, N.~Suenderhauf, N.~Liu, N.~D. Palo, N.~M.~M. Shafiullah, O.~Mees, O.~Kroemer, O.~Bastani, P.~R. Sanketi, P.~T. Miller, P.~Yin, P.~Wohlhart, P.~Xu, P.~D. Fagan, P.~Mitrano, P.~Sermanet, P.~Abbeel,
  P.~Sundaresan, Q.~Chen, Q.~Vuong, R.~Rafailov, R.~Tian, R.~Doshi, R.~Mart'in-Mart'in, R.~Baijal, R.~Scalise, R.~Hendrix, R.~Lin, R.~Qian, R.~Zhang, R.~Mendonca, R.~Shah, R.~Hoque, R.~Julian, S.~Bustamante, S.~Kirmani, S.~Levine, S.~Lin, S.~Moore, S.~Bahl, S.~Dass, S.~Sonawani, S.~Tulsiani, S.~Song, S.~Xu, S.~Haldar, S.~Karamcheti, S.~Adebola, S.~Guist, S.~Nasiriany, S.~Schaal, S.~Welker, S.~Tian, S.~Ramamoorthy, S.~Dasari, S.~Belkhale, S.~Park, S.~Nair, S.~Mirchandani, T.~Osa, T.~Gupta, T.~Harada, T.~Matsushima, T.~Xiao, T.~Kollar, T.~Yu, T.~Ding, T.~Davchev, T.~Z. Zhao, T.~Armstrong, T.~Darrell, T.~Chung, V.~Jain, V.~Kumar, V.~Vanhoucke, W.~Zhan, W.~Zhou, W.~Burgard, X.~Chen, X.~Chen, X.~Wang, X.~Zhu, X.~Geng, X.~Liu, X.~Liangwei, X.~Li, Y.~Pang, Y.~Lu, Y.~J. Ma, Y.~Kim, Y.~Chebotar, Y.~Zhou, Y.~Zhu, Y.~Wu, Y.~Xu, Y.~Wang, Y.~Bisk, Y.~Dou, Y.~Cho, Y.~Lee, Y.~Cui, Y.~Cao, Y.-H. Wu, Y.~Tang, Y.~Zhu, Y.~Zhang, Y.~Jiang, Y.~Li, Y.~Li, Y.~Iwasawa, Y.~Matsuo, Z.~Ma, Z.~Xu, Z.~J. Cui, Z.~Zhang, Z.~Fu, and Z.~Lin,
  ``Open x-embodiment: Robotic learning datasets and rt-x models,'' 2024. [Online]. Available: \url{https://arxiv.org/abs/2310.08864}
\BIBentrySTDinterwordspacing

\bibitem{agibot-world}
T.~AgiBot-World, ``\BIBforeignlanguage{en}{{AgiBot} {World} {Colosseo}: {Large}-scale {Manipulation} {Platform} for {Scalable} and {Intelligent} {Embodied} {Systems}}.''

\bibitem{kim2024openvlaopensourcevisionlanguageactionmodel}
M.~J. Kim, K.~Pertsch, S.~Karamcheti, T.~Xiao, A.~Balakrishna, S.~Nair, R.~Rafailov, E.~Foster, G.~Lam, P.~Sanketi, Q.~Vuong, T.~Kollar, B.~Burchfiel, R.~Tedrake, D.~Sadigh, S.~Levine, P.~Liang, and C.~Finn, ``Openvla: An open-source vision-language-action model,'' in \emph{8th Annual Conference on Robot Learning}.

\bibitem{brohan2023rt2visionlanguageactionmodelstransfer}
\BIBentryALTinterwordspacing
A.~Brohan, N.~Brown, J.~Carbajal, Y.~Chebotar, X.~Chen, K.~Choromanski, T.~Ding, D.~Driess, A.~Dubey, C.~Finn, P.~Florence, C.~Fu, M.~G. Arenas, K.~Gopalakrishnan, K.~Han, K.~Hausman, A.~Herzog, J.~Hsu, B.~Ichter, A.~Irpan, N.~Joshi, R.~Julian, D.~Kalashnikov, Y.~Kuang, I.~Leal, L.~Lee, T.-W.~E. Lee, S.~Levine, Y.~Lu, H.~Michalewski, I.~Mordatch, K.~Pertsch, K.~Rao, K.~Reymann, M.~Ryoo, G.~Salazar, P.~Sanketi, P.~Sermanet, J.~Singh, A.~Singh, R.~Soricut, H.~Tran, V.~Vanhoucke, Q.~Vuong, A.~Wahid, S.~Welker, P.~Wohlhart, J.~Wu, F.~Xia, T.~Xiao, P.~Xu, S.~Xu, T.~Yu, and B.~Zitkovich, ``Rt-2: Vision-language-action models transfer web knowledge to robotic control,'' 2023. [Online]. Available: \url{https://arxiv.org/abs/2307.15818}
\BIBentrySTDinterwordspacing

\bibitem{intelligence2025pi05visionlanguageactionmodelopenworld}
\BIBentryALTinterwordspacing
P.~Intelligence, K.~Black, N.~Brown, J.~Darpinian, K.~Dhabalia, D.~Driess, A.~Esmail, M.~Equi, C.~Finn, N.~Fusai, M.~Y. Galliker, D.~Ghosh, L.~Groom, K.~Hausman, B.~Ichter, S.~Jakubczak, T.~Jones, L.~Ke, D.~LeBlanc, S.~Levine, A.~Li-Bell, M.~Mothukuri, S.~Nair, K.~Pertsch, A.~Z. Ren, L.~X. Shi, L.~Smith, J.~T. Springenberg, K.~Stachowicz, J.~Tanner, Q.~Vuong, H.~Walke, A.~Walling, H.~Wang, L.~Yu, and U.~Zhilinsky, ``$\pi_{0.5}$: a vision-language-action model with open-world generalization,'' 2025. [Online]. Available: \url{https://arxiv.org/abs/2504.16054}
\BIBentrySTDinterwordspacing

\bibitem{yang2025magmafoundationmodelmultimodal}
J.~Yang, R.~Tan, Q.~Wu, R.~Zheng, B.~Peng, Y.~Liang, Y.~Gu, M.~Cai, S.~Ye, J.~Jang \emph{et~al.}, ``Magma: A foundation model for multimodal ai agents,'' in \emph{Proceedings of the Computer Vision and Pattern Recognition Conference}, 2025, pp. 14\,203--14\,214.

\bibitem{reed2022generalistagent}
S.~Reed, K.~Zolna, E.~Parisotto, S.~G. Colmenarejo, A.~Novikov, G.~Barth-maron, M.~Gim{\'e}nez, Y.~Sulsky, J.~Kay, J.~T. Springenberg \emph{et~al.}, ``A generalist agent,'' \emph{Transactions on Machine Learning Research}.

\bibitem{driess2023palmeembodiedmultimodallanguage}
\BIBentryALTinterwordspacing
D.~Driess, F.~Xia, M.~S.~M. Sajjadi, C.~Lynch, A.~Chowdhery, B.~Ichter, A.~Wahid, J.~Tompson, Q.~Vuong, T.~Yu, W.~Huang, Y.~Chebotar, P.~Sermanet, D.~Duckworth, S.~Levine, V.~Vanhoucke, K.~Hausman, M.~Toussaint, K.~Greff, A.~Zeng, I.~Mordatch, and P.~Florence, ``Palm-e: An embodied multimodal language model,'' 2023. [Online]. Available: \url{https://arxiv.org/abs/2303.03378}
\BIBentrySTDinterwordspacing

\bibitem{zawalski2025roboticcontrolembodiedchainofthought}
\BIBentryALTinterwordspacing
M.~Zawalski, W.~Chen, K.~Pertsch, O.~Mees, C.~Finn, and S.~Levine, ``Robotic control via embodied chain-of-thought reasoning,'' 2025. [Online]. Available: \url{https://arxiv.org/abs/2407.08693}
\BIBentrySTDinterwordspacing

\bibitem{bommasani2022opportunitiesrisksfoundationmodels}
\BIBentryALTinterwordspacing
R.~Bommasani, D.~A. Hudson, E.~Adeli, R.~Altman, S.~Arora, S.~von Arx, M.~S. Bernstein, J.~Bohg, A.~Bosselut, E.~Brunskill, E.~Brynjolfsson, S.~Buch, D.~Card, R.~Castellon, N.~Chatterji, A.~Chen, K.~Creel, J.~Q. Davis, D.~Demszky, C.~Donahue, M.~Doumbouya, E.~Durmus, S.~Ermon, J.~Etchemendy, K.~Ethayarajh, L.~Fei-Fei, C.~Finn, T.~Gale, L.~Gillespie, K.~Goel, N.~Goodman, S.~Grossman, N.~Guha, T.~Hashimoto, P.~Henderson, J.~Hewitt, D.~E. Ho, J.~Hong, K.~Hsu, J.~Huang, T.~Icard, S.~Jain, D.~Jurafsky, P.~Kalluri, S.~Karamcheti, G.~Keeling, F.~Khani, O.~Khattab, P.~W. Koh, M.~Krass, R.~Krishna, R.~Kuditipudi, A.~Kumar, F.~Ladhak, M.~Lee, T.~Lee, J.~Leskovec, I.~Levent, X.~L. Li, X.~Li, T.~Ma, A.~Malik, C.~D. Manning, S.~Mirchandani, E.~Mitchell, Z.~Munyikwa, S.~Nair, A.~Narayan, D.~Narayanan, B.~Newman, A.~Nie, J.~C. Niebles, H.~Nilforoshan, J.~Nyarko, G.~Ogut, L.~Orr, I.~Papadimitriou, J.~S. Park, C.~Piech, E.~Portelance, C.~Potts, A.~Raghunathan, R.~Reich, H.~Ren, F.~Rong, Y.~Roohani, C.~Ruiz, J.~Ryan, C.~Ré,
  D.~Sadigh, S.~Sagawa, K.~Santhanam, A.~Shih, K.~Srinivasan, A.~Tamkin, R.~Taori, A.~W. Thomas, F.~Tramèr, R.~E. Wang, W.~Wang, B.~Wu, J.~Wu, Y.~Wu, S.~M. Xie, M.~Yasunaga, J.~You, M.~Zaharia, M.~Zhang, T.~Zhang, X.~Zhang, Y.~Zhang, L.~Zheng, K.~Zhou, and P.~Liang, ``On the opportunities and risks of foundation models,'' 2022. [Online]. Available: \url{https://arxiv.org/abs/2108.07258}
\BIBentrySTDinterwordspacing

\bibitem{gemini_robotics_team_gemini_2025}
\BIBentryALTinterwordspacing
G.~R. Team, ``Gemini {Robotics}: {Bringing} {AI} into the {Physical} {World},'' Tech. Rep., Mar. 2025. [Online]. Available: \url{https://deepmind.google/discover/blog/gemini-robotics-brings-ai-into-the-physical-world/}
\BIBentrySTDinterwordspacing

\bibitem{brohan2023rt1roboticstransformerrealworld}
\BIBentryALTinterwordspacing
A.~Brohan, N.~Brown, J.~Carbajal, Y.~Chebotar, J.~Dabis, C.~Finn, K.~Gopalakrishnan, K.~Hausman, A.~Herzog, J.~Hsu, J.~Ibarz, B.~Ichter, A.~Irpan, T.~Jackson, S.~Jesmonth, N.~J. Joshi, R.~Julian, D.~Kalashnikov, Y.~Kuang, I.~Leal, K.-H. Lee, S.~Levine, Y.~Lu, U.~Malla, D.~Manjunath, I.~Mordatch, O.~Nachum, C.~Parada, J.~Peralta, E.~Perez, K.~Pertsch, J.~Quiambao, K.~Rao, M.~Ryoo, G.~Salazar, P.~Sanketi, K.~Sayed, J.~Singh, S.~Sontakke, A.~Stone, C.~Tan, H.~Tran, V.~Vanhoucke, S.~Vega, Q.~Vuong, F.~Xia, T.~Xiao, P.~Xu, S.~Xu, T.~Yu, and B.~Zitkovich, ``Rt-1: Robotics transformer for real-world control at scale,'' 2023. [Online]. Available: \url{https://arxiv.org/abs/2212.06817}
\BIBentrySTDinterwordspacing

\bibitem{kim_openvla_2024}
\BIBentryALTinterwordspacing
M.~J. Kim, K.~Pertsch, S.~Karamcheti, T.~Xiao, A.~Balakrishna, S.~Nair, R.~Rafailov, E.~Foster, G.~Lam, P.~Sanketi, Q.~Vuong, T.~Kollar, B.~Burchfiel, R.~Tedrake, D.~Sadigh, S.~Levine, P.~Liang, and C.~Finn, ``{OpenVLA}: {An} {Open}-{Source} {Vision}-{Language}-{Action} {Model},'' Sep. 2024, arXiv:2406.09246 [cs]. [Online]. Available: \url{http://arxiv.org/abs/2406.09246}
\BIBentrySTDinterwordspacing

\bibitem{brown2020language}
T.~B. Brown, B.~Mann, N.~Ryder, M.~Subbiah, J.~Kaplan, P.~Dhariwal, A.~Neelakantan, P.~Shyam, G.~Sastry, A.~Askell \emph{et~al.}, ``Language models are few-shot learners,'' in \emph{Advances in Neural Information Processing Systems}, vol.~33, 2020, pp. 1877--1901.

\bibitem{gao2020pile}
L.~Gao, S.~Biderman, S.~Black, L.~Golding, T.~Hoppe, C.~Foster, J.~Phang, H.~He, A.~Thite, N.~Nabeshima \emph{et~al.}, ``The pile: An 800gb dataset of diverse text for language modeling,'' \emph{arXiv preprint arXiv:2101.00027}, 2020.

\bibitem{dodge2021documenting}
J.~Dodge, M.~Sap, A.~Marasovi{\'c}, W.~Agnew, G.~Ilharco, D.~Groeneveld, M.~Mitchell, and M.~Gardner, ``Documenting large webtext corpora: A case study on the colossal clean crawled corpus,'' \emph{arXiv preprint arXiv:2104.08758}, 2021.

\bibitem{schuhmann2022laion}
C.~Schuhmann, R.~Beaumont, R.~Vencu, C.~Gordon, R.~Wightman, M.~Cherti, T.~Coombes, A.~Katta, C.~Mullis, M.~Wortsman \emph{et~al.}, ``Laion-5b: An open large-scale dataset for training next generation image-text models,'' \emph{Advances in neural information processing systems}, vol.~35, pp. 25\,278--25\,294, 2022.

\bibitem{schuhmann2021laion}
C.~Schuhmann, R.~Vencu, R.~Beaumont, R.~Kaczmarczyk, C.~Mullis, A.~Katta, T.~Coombes, J.~Jitsev, and A.~Komatsuzaki, ``Laion-400m: Open dataset of clip-filtered 400 million image-text pairs,'' \emph{arXiv preprint arXiv:2111.02114}, 2021.

\bibitem{liu2023llava}
H.~Liu, C.~Li, Q.~Wu, and Y.~J. Lee, ``Visual instruction tuning,'' in \emph{NeurIPS}, 2023.

\bibitem{ebert2021bridgedataboostinggeneralization}
\BIBentryALTinterwordspacing
F.~Ebert, Y.~Yang, K.~Schmeckpeper, B.~Bucher, G.~Georgakis, K.~Daniilidis, C.~Finn, and S.~Levine, ``Bridge data: Boosting generalization of robotic skills with cross-domain datasets,'' 2021. [Online]. Available: \url{https://arxiv.org/abs/2109.13396}
\BIBentrySTDinterwordspacing

\bibitem{fang2023rh20t}
H.-S. Fang, H.~Fang, Z.~Tang, J.~Liu, J.~Wang, H.~Zhu, and C.~Lu, ``Rh20t: A robotic dataset for learning diverse skills in one-shot,'' in \emph{RSS 2023 Workshop on Learning for Task and Motion Planning}, 2023.

\bibitem{agibotworldcontributors2025agibotworldcolosseolargescale}
\BIBentryALTinterwordspacing
AgiBot-World-Contributors, Q.~Bu, J.~Cai, L.~Chen, X.~Cui, Y.~Ding, S.~Feng, S.~Gao, X.~He, X.~Huang, S.~Jiang, Y.~Jiang, C.~Jing, H.~Li, J.~Li, C.~Liu, Y.~Liu, Y.~Lu, J.~Luo, P.~Luo, Y.~Mu, Y.~Niu, Y.~Pan, J.~Pang, Y.~Qiao, G.~Ren, C.~Ruan, J.~Shan, Y.~Shen, C.~Shi, M.~Shi, M.~Shi, C.~Sima, J.~Song, H.~Wang, W.~Wang, D.~Wei, C.~Xie, G.~Xu, J.~Yan, C.~Yang, L.~Yang, S.~Yang, M.~Yao, J.~Zeng, C.~Zhang, Q.~Zhang, B.~Zhao, C.~Zhao, J.~Zhao, and J.~Zhu, ``Agibot world colosseo: A large-scale manipulation platform for scalable and intelligent embodied systems,'' 2025. [Online]. Available: \url{https://arxiv.org/abs/2503.06669}
\BIBentrySTDinterwordspacing

\bibitem{geng2025roboverse}
\BIBentryALTinterwordspacing
H.~Geng, F.~Wang, S.~Wei, Y.~Li, B.~Wang, B.~An, C.~T. Cheng, H.~Lou, P.~Li, Y.-J. Wang, Y.~Liang, D.~Goetting, C.~Xu, H.~Chen, Y.~Qian, Y.~Geng, J.~Mao, W.~Wan, M.~Zhang, J.~Lyu, S.~Zhao, J.~Zhang, J.~Zhang, C.~Zhao, H.~Lu, Y.~Ding, R.~Gong, Y.~Wang, Y.~Kuang, R.~Wu, B.~Jia, C.~Sferrazza, H.~Dong, S.~Huang, K.~Sreenath, Y.~Wang, J.~Malik, and P.~Abbeel, ``Roboverse: Towards a unified platform, dataset and benchmark for scalable and generalizable robot learning,'' April 2025. [Online]. Available: \url{https://github.com/RoboVerseOrg/RoboVerse}
\BIBentrySTDinterwordspacing

\bibitem{mittal2023orbit}
M.~Mittal, C.~Yu, Q.~Yu, J.~Liu, N.~Rudin, D.~Hoeller, J.~L. Yuan, R.~Singh, Y.~Guo, H.~Mazhar \emph{et~al.}, ``Orbit: A unified simulation framework for interactive robot learning environments,'' \emph{IEEE Robotics and Automation Letters}, vol.~8, no.~6, pp. 3740--3747, 2023.

\bibitem{tao2024maniskill3gpuparallelizedrobotics}
\BIBentryALTinterwordspacing
S.~Tao, F.~Xiang, A.~Shukla, Y.~Qin, X.~Hinrichsen, X.~Yuan, C.~Bao, X.~Lin, Y.~Liu, T.~kai Chan, Y.~Gao, X.~Li, T.~Mu, N.~Xiao, A.~Gurha, Z.~Huang, R.~Calandra, R.~Chen, S.~Luo, and H.~Su, ``Maniskill3: Gpu parallelized robotics simulation and rendering for generalizable embodied ai,'' 2024. [Online]. Available: \url{https://arxiv.org/abs/2410.00425}
\BIBentrySTDinterwordspacing

\bibitem{wang2023robogen}
Y.~Wang, Z.~Xian, F.~Chen, T.-H. Wang, Y.~Wang, K.~Fragkiadaki, Z.~Erickson, D.~Held, and C.~Gan, ``Robogen: Towards unleashing infinite data for automated robot learning via generative simulation,'' \emph{arXiv preprint arXiv:2311.01455}, 2023.

\bibitem{nasiriany2024robocasa}
S.~Nasiriany, A.~Maddukuri, L.~Zhang, A.~Parikh, A.~Lo, A.~Joshi, A.~Mandlekar, and Y.~Zhu, ``Robocasa: Large-scale simulation of everyday tasks for generalist robots,'' \emph{arXiv preprint arXiv:2406.02523}, 2024.

\bibitem{mandlekar2023mimicgen}
A.~Mandlekar, S.~Nasiriany, B.~Wen, I.~Akinola, Y.~Narang, L.~Fan, Y.~Zhu, and D.~Fox, ``Mimicgen: A data generation system for scalable robot learning using human demonstrations,'' \emph{arXiv preprint arXiv:2310.17596}, 2023.

\bibitem{james2020rlbench}
S.~James, Z.~Ma, D.~R. Arrojo, and A.~J. Davison, ``Rlbench: The robot learning benchmark \& learning environment,'' \emph{IEEE Robotics and Automation Letters}, vol.~5, no.~2, pp. 3019--3026, 2020.

\bibitem{wei2025empiricalanalysissimandrealcotraining}
\BIBentryALTinterwordspacing
A.~Wei, A.~Agarwal, B.~Chen, R.~Bosworth, N.~Pfaff, and R.~Tedrake, ``Empirical analysis of sim-and-real cotraining of diffusion policies for planar pushing from pixels,'' 2025. [Online]. Available: \url{https://arxiv.org/abs/2503.22634}
\BIBentrySTDinterwordspacing

\bibitem{maddukuri2025simandrealcotrainingsimplerecipe}
\BIBentryALTinterwordspacing
A.~Maddukuri, Z.~Jiang, L.~Y. Chen, S.~Nasiriany, Y.~Xie, Y.~Fang, W.~Huang, Z.~Wang, Z.~Xu, N.~Chernyadev, S.~Reed, K.~Goldberg, A.~Mandlekar, L.~Fan, and Y.~Zhu, ``Sim-and-real co-training: A simple recipe for vision-based robotic manipulation,'' 2025. [Online]. Available: \url{https://arxiv.org/abs/2503.24361}
\BIBentrySTDinterwordspacing

\bibitem{chi2024universalmanipulationinterfaceinthewild}
\BIBentryALTinterwordspacing
C.~Chi, Z.~Xu, C.~Pan, E.~Cousineau, B.~Burchfiel, S.~Feng, R.~Tedrake, and S.~Song, ``Universal manipulation interface: In-the-wild robot teaching without in-the-wild robots,'' 2024. [Online]. Available: \url{https://arxiv.org/abs/2402.10329}
\BIBentrySTDinterwordspacing

\bibitem{Seo_2025}
\BIBentryALTinterwordspacing
M.~Seo, H.~A. Park, S.~Yuan, Y.~Zhu, and L.~Sentis, ``Legato: Cross-embodiment imitation using a grasping tool,'' \emph{IEEE Robotics and Automation Letters}, vol.~10, no.~3, p. 2854–2861, Mar. 2025. [Online]. Available: \url{http://dx.doi.org/10.1109/LRA.2025.3535182}
\BIBentrySTDinterwordspacing

\bibitem{kareer2024egomimicscalingimitationlearning}
\BIBentryALTinterwordspacing
S.~Kareer, D.~Patel, R.~Punamiya, P.~Mathur, S.~Cheng, C.~Wang, J.~Hoffman, and D.~Xu, ``Egomimic: Scaling imitation learning via egocentric video,'' 2024. [Online]. Available: \url{https://arxiv.org/abs/2410.24221}
\BIBentrySTDinterwordspacing

\bibitem{fang2024airexolowcostexoskeletonslearning}
\BIBentryALTinterwordspacing
H.~Fang, H.-S. Fang, Y.~Wang, J.~Ren, J.~Chen, R.~Zhang, W.~Wang, and C.~Lu, ``Airexo: Low-cost exoskeletons for learning whole-arm manipulation in the wild,'' 2024. [Online]. Available: \url{https://arxiv.org/abs/2309.14975}
\BIBentrySTDinterwordspacing

\bibitem{kress2024robot}
H.~Kress-Gazit, K.~Hashimoto, N.~Kuppuswamy, P.~Shah, P.~Horgan, G.~Richardson, S.~Feng, and B.~Burchfiel, ``Robot learning as an empirical science: Best practices for policy evaluation,'' \emph{arXiv preprint arXiv:2409.09491}, 2024.

\bibitem{mu2021maniskillgeneralizablemanipulationskill}
\BIBentryALTinterwordspacing
T.~Mu, Z.~Ling, F.~Xiang, D.~Yang, X.~Li, S.~Tao, Z.~Huang, Z.~Jia, and H.~Su, ``Maniskill: Generalizable manipulation skill benchmark with large-scale demonstrations,'' 2021. [Online]. Available: \url{https://arxiv.org/abs/2107.14483}
\BIBentrySTDinterwordspacing

\bibitem{yu2021metaworldbenchmarkevaluationmultitask}
\BIBentryALTinterwordspacing
T.~Yu, D.~Quillen, Z.~He, R.~Julian, A.~Narayan, H.~Shively, A.~Bellathur, K.~Hausman, C.~Finn, and S.~Levine, ``Meta-world: A benchmark and evaluation for multi-task and meta reinforcement learning,'' 2021. [Online]. Available: \url{https://arxiv.org/abs/1910.10897}
\BIBentrySTDinterwordspacing

\bibitem{robosuite2020}
Y.~Zhu, J.~Wong, A.~Mandlekar, R.~Mart\'{i}n-Mart\'{i}n, A.~Joshi, S.~Nasiriany, Y.~Zhu, and K.~Lin, ``robosuite: A modular simulation framework and benchmark for robot learning,'' in \emph{arXiv preprint arXiv:2009.12293}, 2020.

\bibitem{srivastava2021behaviorbenchmarkeverydayhousehold}
\BIBentryALTinterwordspacing
S.~Srivastava, C.~Li, M.~Lingelbach, R.~Martín-Martín, F.~Xia, K.~Vainio, Z.~Lian, C.~Gokmen, S.~Buch, C.~K. Liu, S.~Savarese, H.~Gweon, J.~Wu, and L.~Fei-Fei, ``Behavior: Benchmark for everyday household activities in virtual, interactive, and ecological environments,'' 2021. [Online]. Available: \url{https://arxiv.org/abs/2108.03332}
\BIBentrySTDinterwordspacing

\bibitem{deitke2020robothoropensimulationtorealembodied}
\BIBentryALTinterwordspacing
M.~Deitke, W.~Han, A.~Herrasti, A.~Kembhavi, E.~Kolve, R.~Mottaghi, J.~Salvador, D.~Schwenk, E.~VanderBilt, M.~Wallingford, L.~Weihs, M.~Yatskar, and A.~Farhadi, ``Robothor: An open simulation-to-real embodied ai platform,'' 2020. [Online]. Available: \url{https://arxiv.org/abs/2004.06799}
\BIBentrySTDinterwordspacing

\bibitem{Kadian_2020}
\BIBentryALTinterwordspacing
A.~Kadian, J.~Truong, A.~Gokaslan, A.~Clegg, E.~Wijmans, S.~Lee, M.~Savva, S.~Chernova, and D.~Batra, ``Sim2real predictivity: Does evaluation in simulation predict real-world performance?'' \emph{IEEE Robotics and Automation Letters}, vol.~5, no.~4, p. 6670–6677, Oct. 2020. [Online]. Available: \url{http://dx.doi.org/10.1109/LRA.2020.3013848}
\BIBentrySTDinterwordspacing

\bibitem{zhang2019vrgogglesrobotsrealtosimdomain}
\BIBentryALTinterwordspacing
J.~Zhang, L.~Tai, P.~Yun, Y.~Xiong, M.~Liu, J.~Boedecker, and W.~Burgard, ``Vr-goggles for robots: Real-to-sim domain adaptation for visual control,'' 2019. [Online]. Available: \url{https://arxiv.org/abs/1802.00265}
\BIBentrySTDinterwordspacing

\bibitem{li24simpler}
X.~Li, K.~Hsu, J.~Gu, K.~Pertsch, O.~Mees, H.~R. Walke, C.~Fu, I.~Lunawat, I.~Sieh, S.~Kirmani, S.~Levine, J.~Wu, C.~Finn, H.~Su, Q.~Vuong, and T.~Xiao, ``Evaluating real-world robot manipulation policies in simulation,'' \emph{arXiv preprint arXiv:2405.05941}, 2024.

\bibitem{memmel2024asidactiveexplorationidentification}
\BIBentryALTinterwordspacing
M.~Memmel, A.~Wagenmaker, C.~Zhu, P.~Yin, D.~Fox, and A.~Gupta, ``Asid: Active exploration for system identification in robotic manipulation,'' 2024. [Online]. Available: \url{https://arxiv.org/abs/2404.12308}
\BIBentrySTDinterwordspacing

\bibitem{pfaff2025_scalable_real2sim}
\BIBentryALTinterwordspacing
N.~Pfaff, E.~Fu, J.~Binagia, P.~Isola, and R.~Tedrake, ``Scalable real2sim: Physics-aware asset generation via robotic pick-and-place setups,'' 2025. [Online]. Available: \url{https://arxiv.org/abs/2503.00370}
\BIBentrySTDinterwordspacing

\bibitem{dasari2022rb2roboticmanipulationbenchmarking}
\BIBentryALTinterwordspacing
S.~Dasari, J.~Wang, J.~Hong, S.~Bahl, Y.~Lin, A.~Wang, A.~Thankaraj, K.~Chahal, B.~Calli, S.~Gupta, D.~Held, L.~Pinto, D.~Pathak, V.~Kumar, and A.~Gupta, ``Rb2: Robotic manipulation benchmarking with a twist,'' 2022. [Online]. Available: \url{https://arxiv.org/abs/2203.08098}
\BIBentrySTDinterwordspacing

\bibitem{morgan2019benchmarking}
A.~S. Morgan, K.~Hang, W.~G. Bircher, F.~M. Alladkani, A.~Gandhi, B.~Calli, and A.~M. Dollar, ``Benchmarking cluttered robot pick-and-place manipulation with the box and blocks test,'' \emph{IEEE Robotics and Automation Letters}, vol.~5, no.~2, pp. 454--461, 2019.

\bibitem{heo2023furniturebench}
M.~Heo, Y.~Lee, D.~Lee, and J.~J. Lim, ``Furniturebench: Reproducible real-world benchmark for long-horizon complex manipulation,'' \emph{The International Journal of Robotics Research}, p. 02783649241304789, 2023.

\bibitem{kimble2020benchmarking}
K.~Kimble, K.~Van~Wyk, J.~Falco, E.~Messina, Y.~Sun, M.~Shibata, W.~Uemura, and Y.~Yokokohji, ``Benchmarking protocols for evaluating small parts robotic assembly systems,'' \emph{IEEE robotics and automation letters}, vol.~5, no.~2, pp. 883--889, 2020.

\bibitem{khargonkar2024scenereplica}
N.~Khargonkar, S.~H. Allu, Y.~Lu, B.~Prabhakaran, Y.~Xiang \emph{et~al.}, ``Scenereplica: Benchmarking real-world robot manipulation by creating replicable scenes,'' in \emph{2024 IEEE International Conference on Robotics and Automation (ICRA)}.\hskip 1em plus 0.5em minus 0.4em\relax IEEE, 2024, pp. 8258--8264.

\bibitem{bekiroglu2019benchmarking}
Y.~Bekiroglu, N.~Marturi, M.~A. Roa, K.~J.~M. Adjigble, T.~Pardi, C.~Grimm, R.~Balasubramanian, K.~Hang, and R.~Stolkin, ``Benchmarking protocol for grasp planning algorithms,'' \emph{IEEE Robotics and Automation Letters}, vol.~5, no.~2, pp. 315--322, 2019.

\bibitem{bottarel2020graspa}
F.~Bottarel, G.~Vezzani, U.~Pattacini, and L.~Natale, ``Graspa 1.0: Graspa is a robot arm grasping performance benchmark,'' \emph{IEEE Robotics and Automation Letters}, vol.~5, no.~2, pp. 836--843, 2020.

\bibitem{chatzilygeroudis2020benchmark}
K.~Chatzilygeroudis, B.~Fichera, I.~Lauzana, F.~Bu, K.~Yao, F.~Khadivar, and A.~Billard, ``Benchmark for bimanual robotic manipulation of semi-deformable objects,'' \emph{IEEE Robotics and Automation Letters}, vol.~5, no.~2, pp. 2443--2450, 2020.

\bibitem{liu2021ocrtoc}
Z.~Liu, W.~Liu, Y.~Qin, F.~Xiang, M.~Gou, S.~Xin, M.~A. Roa, B.~Calli, H.~Su, Y.~Sun \emph{et~al.}, ``Ocrtoc: A cloud-based competition and benchmark for robotic grasping and manipulation,'' \emph{IEEE Robotics and Automation Letters}, vol.~7, no.~1, pp. 486--493, 2021.

\bibitem{bauer2022real}
S.~Bauer, M.~W{\"u}thrich, F.~Widmaier, A.~Buchholz, S.~Stark, A.~Goyal, T.~Steinbrenner, J.~Akpo, S.~Joshi, V.~Berenz \emph{et~al.}, ``Real robot challenge: A robotics competition in the cloud,'' in \emph{NeurIPS 2021 Competitions and Demonstrations Track}.\hskip 1em plus 0.5em minus 0.4em\relax PMLR, 2022, pp. 190--204.

\bibitem{zhou2023train}
G.~Zhou, V.~Dean, M.~K. Srirama, A.~Rajeswaran, J.~Pari, K.~Hatch, A.~Jain, T.~Yu, P.~Abbeel, L.~Pinto \emph{et~al.}, ``Train offline, test online: A real robot learning benchmark,'' in \emph{2023 IEEE International Conference on Robotics and Automation (ICRA)}.\hskip 1em plus 0.5em minus 0.4em\relax IEEE, 2023, pp. 9197--9203.

\bibitem{zhou2025autoevalautonomousevaluationgeneralist}
\BIBentryALTinterwordspacing
Z.~Zhou, P.~Atreya, Y.~L. Tan, K.~Pertsch, and S.~Levine, ``Autoeval: Autonomous evaluation of generalist robot manipulation policies in the real world,'' 2025. [Online]. Available: \url{https://arxiv.org/abs/2503.24278}
\BIBentrySTDinterwordspacing

\bibitem{snyder2025your}
D.~Snyder, A.~J. Hancock, A.~Badithela, E.~Dixon, P.~Miller, R.~A. Ambrus, A.~Majumdar, M.~Itkina, and H.~Nishimura, ``Is your imitation learning policy better than mine? policy comparison with near-optimal stopping,'' \emph{arXiv preprint arXiv:2503.10966}, 2025.

\bibitem{agarwal2021deep}
R.~Agarwal, M.~Schwarzer, P.~S. Castro, A.~C. Courville, and M.~Bellemare, ``Deep reinforcement learning at the edge of the statistical precipice,'' \emph{Advances in neural information processing systems}, vol.~34, pp. 29\,304--29\,320, 2021.

\bibitem{greenland2016statistical}
S.~Greenland, S.~J. Senn, K.~J. Rothman, J.~B. Carlin, C.~Poole, S.~N. Goodman, and D.~G. Altman, ``Statistical tests, p values, confidence intervals, and power: a guide to misinterpretations,'' \emph{European journal of epidemiology}, vol.~31, no.~4, pp. 337--350, 2016.

\bibitem{piepho2004algorithm}
H.-P. Piepho, ``An algorithm for a letter-based representation of all-pairwise comparisons,'' \emph{Journal of Computational and Graphical Statistics}, vol.~13, no.~2, pp. 456--466, 2004.

\bibitem{lai1988nearly}
T.~L. Lai, ``Nearly optimal sequential tests of composite hypotheses,'' \emph{The Annals of Statistics}, pp. 856--886, 1988.

\bibitem{welch1947generalization}
B.~L. Welch, ``The generalization of ‘student's’problem when several different population varlances are involved,'' \emph{Biometrika}, vol.~34, no. 1-2, pp. 28--35, 1947.

\bibitem{peebles2023scalablediffusionmodelstransformers}
\BIBentryALTinterwordspacing
W.~Peebles and S.~Xie, ``Scalable diffusion models with transformers,'' 2023. [Online]. Available: \url{https://arxiv.org/abs/2212.09748}
\BIBentrySTDinterwordspacing

\bibitem{song_denoising_2022}
\BIBentryALTinterwordspacing
J.~Song, C.~Meng, and S.~Ermon, ``Denoising {Diffusion} {Implicit} {Models},'' Oct. 2022, arXiv:2010.02502 [cs]. [Online]. Available: \url{http://arxiv.org/abs/2010.02502}
\BIBentrySTDinterwordspacing

\bibitem{ho_denoising_2020}
\BIBentryALTinterwordspacing
J.~Ho, A.~Jain, and P.~Abbeel, ``Denoising {Diffusion} {Probabilistic} {Models},'' Dec. 2020, arXiv:2006.11239 [cs, stat]. [Online]. Available: \url{http://arxiv.org/abs/2006.11239}
\BIBentrySTDinterwordspacing

\bibitem{black_0_nodate}
K.~Black, N.~Brown, D.~Driess, A.~Esmail, M.~Equi, C.~Finn, N.~Fusai, L.~Groom, K.~Hausman, B.~Ichter, S.~Jakubczak, T.~Jones, L.~Ke, S.~Levine, A.~Li-Bell, M.~Mothukuri, S.~Nair, K.~Pertsch, L.~X. Shi, J.~Tanner, Q.~Vuong, A.~Walling, H.~Wang, and U.~Zhilinsky, ``\BIBforeignlanguage{en}{$\pi$0: {A} {Vision}-{Language}-{Action} {Flow} {Model} for {General} {Robot} {Control}},'' Tech. Rep.

\bibitem{DBLP:journals/firai/CrooksVOMR16}
\BIBentryALTinterwordspacing
W.~Crooks, G.~Vukasin, M.~O'Sullivan, W.~C. Messner, and C.~Rogers, ``Fin ray{\textregistered} effect inspired soft robotic gripper: From the robosoft grand challenge toward optimization,'' \emph{Frontiers Robotics {AI}}, vol.~3, p.~70, 2016. [Online]. Available: \url{https://doi.org/10.3389/frobt.2016.00070}
\BIBentrySTDinterwordspacing

\bibitem{drake}
\BIBentryALTinterwordspacing
R.~Tedrake and the Drake Development~Team, ``Drake: Model-based design and verification for robotics,'' 2025. [Online]. Available: \url{https://drake.mit.edu}
\BIBentrySTDinterwordspacing

\end{thebibliography}
